\documentclass[lettersize,journal,twoside]{IEEEtran}
\usepackage{amsmath,amsfonts}
\usepackage{algorithm}
\usepackage[noend]{algpseudocode}
\usepackage{array}
\usepackage{textcomp}
\usepackage{stfloats}
\usepackage{url}
\usepackage{verbatim}
\usepackage{graphicx}
\usepackage{booktabs, multirow, makecell, subcaption}
\usepackage{cite}
\usepackage{hyperref}
\usepackage[table, svgnames, dvipsnames]{xcolor}
\usepackage{xcolor}
\usepackage{bm}
\usepackage[normalem]{ulem}

\usepackage{tabularx}

\begin{document}
\newcommand{\blue}{\textcolor{black}}
\newcommand{\red}{\textcolor{red}}

\title{GeFL: Model-Agnostic Federated Learning with Generative Models}

\author{Honggu~Kang\href{https://orcid.org/0000-0003-2138-8641}{\includegraphics[scale=0.5]{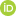}}, Seohyeon~Cha, and Joonhyuk~Kang\href{https://orcid.org/0000-0002-5508-3742}{\includegraphics[scale=0.5]{fig/ORCID-iD_icon_16x16.png}} ,~\IEEEmembership{Member,~IEEE}

\thanks{Manuscript received XXX, XX, 2024; revised XXX, XX, 2025; accepted XXX, XX, 2025.
Recommended for acceptance by Dr. X.
\textit{(Honggu Kang and Seohyeon Cha contributed equally to this work.)}}
\thanks{H.~Kang is with Samsung Electronics, Suwon 16677, South Korea. (e-mail: rednine13@gmail.com).}
\thanks{S.~Cha is with the University of Texas at Austin, TX 78712, USA (e-mail: seohyeon.cha@utexas.edu)}
\thanks{J. Kang is with the School of Electrical Engineering, Korea Advanced Institute of Science and Technology (KAIST), Daejeon 305-701, South Korea. (e-mail: jhkang@ee.kaist.ac.kr).}}

\markboth{IEEE Transactions on Mobile Computing,~Vol.~X, No.~X, X~2025}%
{Kang \MakeLowercase{\textit{et al.}}: GeFL: Model-Agnostic Federated Learning with Generative Models}


\maketitle

\begin{abstract}
\blue{Federated learning (FL) is a distributed training paradigm that enables collaborative learning across clients without sharing local data, thereby preserving privacy. However, the increasing scale and complexity of modern deep models often exceed the computational or memory capabilities of edge devices. Furthermore, clients may be constrained to use heterogeneous model architectures due to hardware variability (e.g., ASICs, FPGAs) or proprietary requirements that prevent the disclosure or modification of local model structures. These practical considerations motivate the need for model-heterogeneous FL, where clients participate using distinct model architectures.
In this work, we propose Generative Model-Aided Federated Learning (\textsc{GeFL}), a framework that enables cross-client knowledge sharing via a generative model trained in a federated manner. This generative model captures global data semantics and facilitates local training without requiring model homogeneity across clients.
While \textsc{GeFL} achieves strong performance, empirical analysis reveals limitations in scalability and potential privacy leakage due to generative sample memorization. To address these concerns, we propose \textsc{GeFL-F}, which utilizes feature-level generative modeling. This approach enhances scalability to large client populations and mitigates privacy risks.
Extensive experiments across image classification tasks demonstrate that both \textsc{GeFL} and \textsc{GeFL-F} offer competitive performance in heterogeneous settings. Code is available at \cite{Kang}.}
\end{abstract}


\section{Introduction}

\IEEEPARstart{D}{eep} learning has demonstrated remarkable success across various domains, primarily driven by the availability of abundant training data. However, with the proliferation of edge devices such as mobile and Internet-of-Things (IoT) devices, which generate substantial amounts of decentralized data, a significant challenge has emerged: effectively aggregating distributed information and collaboratively training neural networks without compromising privacy. Federated learning (FL) offers a promising solution, enabling collaborative model training across clients while preserving data privacy by not sharing raw data \cite{mcmahan2017communication}. This potential has been actively explored across numerous practical applications \cite{kang2023nefl, afonin2022towards}.

At the same time, two important trends pose additional challenges for FL systems: the rapid growth in model size \cite{geminiteam2023gemini, Brown2020Language, villalobos2022machine} and the increasing heterogeneity of edge devices \cite{Pfeiffer2023federated}. Devices differ widely in computational power, memory capacity, and communication capabilities \blue{In real-world deployments, devices differ significantly in computational resources, memory capacity, energy availability, and network bandwidth \cite{nguyen2021flsurvey}. These limitations make it impractical for many edge devices to train or store a large shared global model. Consequently, clients often require lightweight, personalized models tailored to their hardware capabilities and operational constraints.} 

\blue{Nevertheless, most existing FL algorithms assume the existence of \textit{a single shared global model} \cite{mcmahan2017communication, Li2020fedprox}, which is collaboratively trained by all clients. This assumption creates a substantial obstacle when clients possess heterogeneous models adapted to their own resources. Moreover, clients may use proprietary, hardware-specific, or pre-trained models that cannot easily be replaced or modified. Revealing model architectures may also be undesirable due to privacy, security, or competitive concerns. Therefore, a key challenge arises: \textit{how can clients with diverse, private model architectures collaboratively train models and share knowledge in a FL environment without sacrificing privacy or requiring model homogeneity}?}

While some prior works have explored multi-model FL \cite{horvath2021fjord,kang2023nefl,kim2023depthfl}, they often rely on submodels of a predefined global model, limiting applicability when client models differ fundamentally. Other approaches propose leveraging additional public datasets to align knowledge \cite{lin2020ensemble, huang2022learn, li2019fedmd, yi2023fedgh}; however, these methods depend heavily on the availability of curated public data, which may not be practical in real-world scenarios.

To address these challenges, we propose \textit{Generative model-aided Federated Learning (\textsc{GeFL})}, a novel framework that enables knowledge aggregation across clients with heterogeneous models. Instead of requiring model homogeneity or external public data, \textsc{GeFL} employs a federated generative model trained across local client data. This generative model can synthesize new training samples, allowing clients to enrich their own model training using shared global knowledge while maintaining architectural independence.
However, training a generative model in a federated setting presents its own challenges, particularly as the number of clients grows, leading to issues like mode collapse and degraded sample quality \cite{goodfellow2014generative, kingma2022autoencoding, sohl2015deep}. Furthermore, concerns remain regarding the potential leakage of sensitive information through generative models. To address these issues, we further propose {\textsc{GeFL-F}}, an extension of \textsc{GeFL} that employs feature-generative models — generative models that produce lower-dimensional feature representations rather than full data samples. \textsc{GeFL-F} improves privacy, scalability, and communication efficiency while maintaining the performance benefits offered by generative models.
In summary, our work addresses the dual challenge of enabling FL across heterogeneous target models while collaboratively training a shared federated generative model, without relying on strong assumptions such as global model homogeneity or the availability of public datasets.


The main contributions of this paper are as follows:
\begin{itemize}
\item We propose {\textsc{GeFL}}, a framework that enables federated learning with heterogeneous client models by leveraging federated generative models. \item We design {\textsc{GeFL-F}}, an improved framework using feature-generative models to enhance scalability, communication efficiency, and privacy. \item We provide extensive experiments demonstrating that \textsc{GeFL} and \textsc{GeFL-F} achieve significant performance gains compared to conventional FL and model-heterogeneous baselines under various system settings.
\end{itemize}
The remainder of this paper is organized as follows. Section \ref{sec:related_works} briefly reviews related works and we present our proposed method, \textsc{GeFL} in \ref{sec:gefl}. Then, we introduce \textsc{GeFL-F} in Section \ref{sec:geflf}. 
Section \ref{sec:insights} presents the insights obtained in experimental results and Section\ref{sec:conclusion} concludes the paper. Detailed experimental settings are presented in Appendix \cite{Kang}.

\begin{table}[t]
\caption{Summarization of main notations}
\centering
\resizebox{.48\textwidth}{!}{
{\renewcommand{\arraystretch}{1.15}
\begin{tabular}{cc}
\toprule
Symbol & Definition \\
\midrule
$\mathcal{C}$ & Client set in each round\\
\multirow{2}{*}{$T_{KA}$} & \multirowcell{2}{Number of FL communication rounds for\\ training generative model}\\\\
\multirow{2}{*}{$T_{TN}$} & \multirowcell{2}{Number of FL communication rounds for\\ training target networks}\\\\
\multirow{2}{*}{$T_{FE}$} & \multirowcell{2}{Number of FL communication rounds for\\ training feature extractor}\\\\
$T_g$ & Number of local epochs for training generative model\\
\multirow{2}{*}{$T_s$} & \multirowcell{2}{Number of local epochs for training target networks\\ by synthetic samples}\\\\
\multirow{2}{*}{$T_r$} & \multirowcell{2}{Number of local epochs for training target networks\\ by real samples}\\\\
\multirow{2}{*}{$T_w$} & \multirowcell{2}{Number of local epochs for warming up\\ common feature extractor}\\ \\
$\mathbf{w}_{g}$ & Global weights of generative model\\
$\mathbf{w}_{k}$ & Local weights of generative model of client $k$\\
$x, y$ & Data sample from a dataset\\
$x^{\text feat}, y$ & Feature sample from a feature geneative model\\
$B$ & Batch size\\
$\mathcal{D}_k$ & Local dataset of client $k$\\
$M$ & Number of heterogeneous models\\
$\bm{\theta}_{g,m}$ & Global weights of $m$-th target model \\
$\bm{\theta}_{k,m}$ & Local weights of $m$-th target model of client $k$ \\
$\bm{\theta}^{f}_{g}$ & Global weights of common feature extractor \\
$\bm{\theta}^{h}_{g,m}$ & Global weights of header of $m$-th target model \\
$F(\cdot)$ & Feature extractor model \\
$G(\cdot)$ & Federated generative model \\
$J_G(\cdot)$ & Loss function for training generative model\\
$J(\cdot)$ & Loss function for training target networks\\
$\alpha$ & Learning rate for training target networks \\
$\beta$ & Learning rate for training a generative model\\
\bottomrule
\end{tabular}}}
\vspace{-0.1in}
\end{table}

\section{Related Works}\label{sec:related_works}
\subsection{Federated learning}
FL trains models while data is distributed across multiple clients. Most approaches involve training a \textit{single} global model \cite{mcmahan2017communication, Li2020fedprox, karimireddy2020scaffold} and the foundational algorithm, FedAvg \cite{mcmahan2017communication} gathers global knowledge at the server by aggregating the model parameters from clients and averaging them. Another line of work integrates knowledge distillation (KD) \cite{hinton2015distilling} into FL to transfer knowledge from client models to the server model \cite{seo2020federated, lin2020ensemble, wu2022fedkd, li2023feddkd}. 
For example, FedKD \cite{wu2022fedkd} shares a small student model while keeping a local teacher model at each client to reduce communication costs. FedDKD \cite{li2023feddkd} addresses data heterogeneity across clients by averaging the knowledge of clients by KD. However, none of these works specifically aim to address model heterogeneity. 

\subsection{Model heterogeneous FL}
To address model heterogeneity in FL, several approaches have proposed scaling a global model into submodels in widthwise \cite{horvath2021fjord,diao2021heterofl}, in depthwise \cite{kim2023depthfl}, and both widthwise and depthwise \cite{kang2023nefl}. However, these methods scale the model into a subset of the global model, requiring all clients to share the same model architecture. Our proposed method allows for the training of heterogeneous models with different architectures across clients.

Another approach tackles model heterogeneous FL by leveraging the concept of ensembling various models. AvgKD \cite{afonin2022towards} requires a client to receive all other models from other clients and compute an averaged logit of all logits from local samples of these models. However, this can be computationally infeasible with a large number of clients and less effective when models across clients have significant architectural differences. In addition to the above techniques, some studies introduce architectural conditions such as training a common extractor \cite{liang2020think} or sharing additional model \cite{shen2020federated}. 

Recent studies have proposed to incorporate additional public data to enhance ensembling methods for addressing model heterogeneity \cite{lin2020ensemble, huang2022learn, li2019fedmd, yi2023fedgh}. For instance, FedDF \cite{lin2020ensemble} conducts ensemble distillation between the client and global models, utilizing synthetic data from a pre-trained generator or unlabeled public data. FCCL \cite{huang2022learn} incorporates unlabeled datasets by introducing a loss term defined on logits of unlabeled public data and the averaged logit across clients. FedMD \cite{li2019fedmd} leverages labeled public data for transfer learning through knowledge distillation. pFedHR \cite{wang2023towards} has the flexibility to leverage either labeled or unlabeled public data. However, these algorithms often require access to a well-curated public dataset in either clients or the server, which does not exist in practical for real-world training. Instead, our framework employs existing local data in each client.


 
\subsection{Generative models in FL}
Several works have delved into FL by incorporating generative models, such as VAE \cite{wen2022communication} and GAN \cite{jeong2023communicationefficient, wu2021fedcg, rasouli2020fedgan}. FedGAN \cite{rasouli2020fedgan} investigates the federated training of GANs, while other studies involve training a generator supervised by a well-trained model to address statistical heterogeneity when training a single model \cite{zhu2021datafree, zhang2022finetuning}. In previous studies, generative models could not be effectively trained when client models were heterogeneous, limiting their applicability. In contrast, our approach enables generative models to be trained collaboratively, even in the presence of heterogeneous client models.
Furthermore, existing methods often fail to significantly enhance generalization, as the generator tends to produce samples that only reinforce the existing capabilities of well-trained models. While FedCG \cite{wu2021fedcg} adopts a similar technique by employing a conditional GAN to generate intermediate features, it does not tackle model heterogeneity. FedCG relies on a shared classifier between clients, which inherently limits its ability to handle model heterogeneity. In contrast, \textsc{GeFL-F} shares the feature extractor, offering a more effective method for accommodating heterogeneous models.

\section{Generative Model-aided Model Heterogeneous FL\label{sec:gefl}}
\subsection{Framework}
\begin{figure}[!t]
    \centering
    \includegraphics[width=0.9\columnwidth]{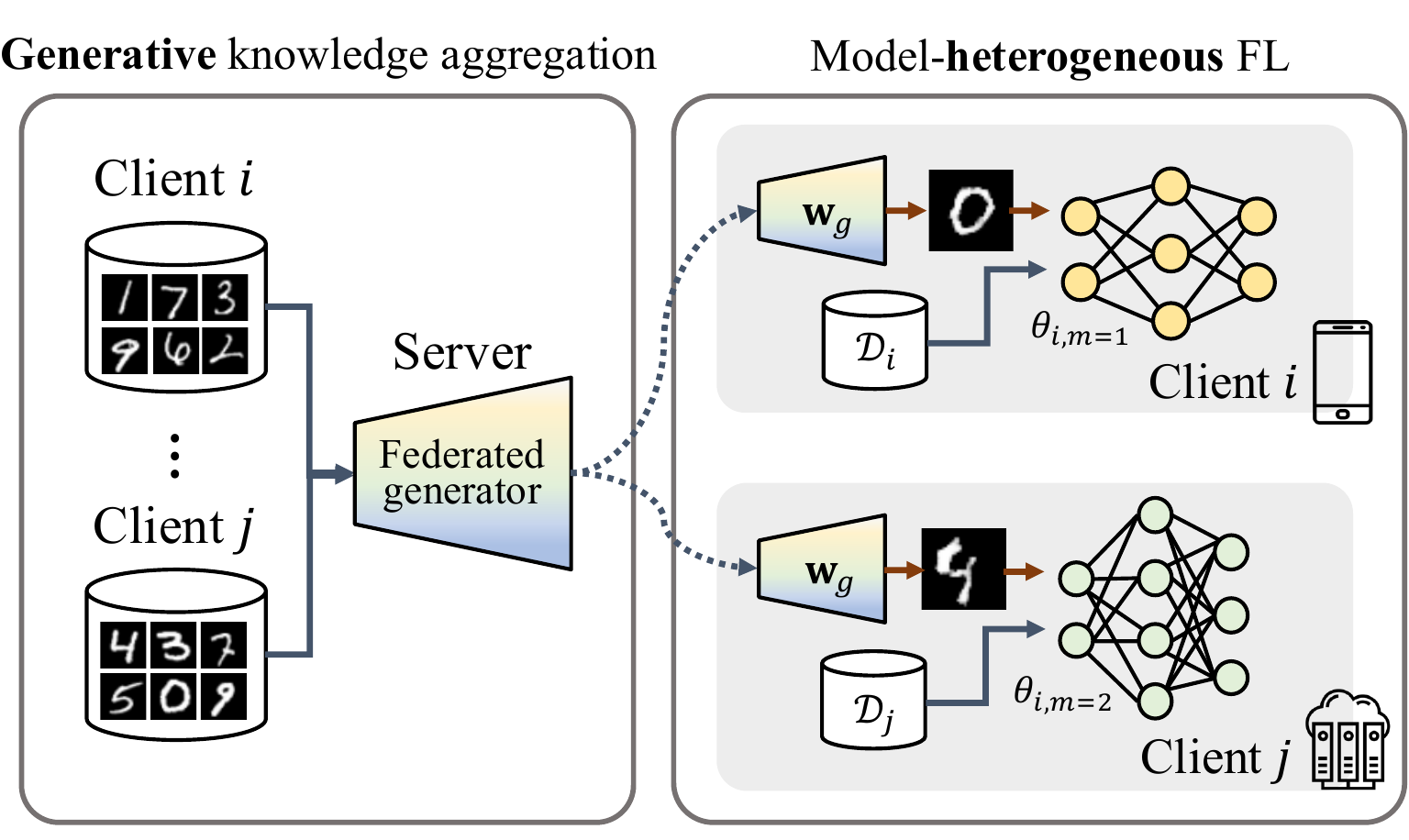}
       \caption{\blue{\textbf{Illustration of \textsc{GeFL}}. We propose a model-agnostic FL framework under model-heterogeneity, \textsc{GeFL}, consisting of (i) \textit{generative knowledge aggregation} which trains a generative model in a federated manner, and (ii) heterogeneous \textit{target network training} augmented by trained generative models.}}
    \label{fig:framework}
    \vspace{-0.4cm}
\end{figure}



To address the model heterogeneity in FL, \textsc{GeFL} incorporates a conditional generative model trained in a federated manner using local clients' data, as illustrated in Figure~\ref{fig:framework}. \textsc{GeFL} consists of two main processes: (i) federated generative model (FedGen) training for \textit{global knowledge aggregation} and (ii) \textit{target network training} augmented by generative models (see Algorithm \ref{alg:gefl} for details).

In the generative knowledge aggregation stage, generative models are trained to capture the representation of real samples. Each client $k \in \mathcal{C}$ trains a generative model $G$ with parameters $\mathbf{w}_k$ using its local data. The server then aggregates the generative model parameters as
\blue{$\mathbf{w}_g \leftarrow \text{Agg}(\{\mathbf{w}_k\}_{k\in\mathcal{C}})=\frac{1}{|\mathcal{C}|}\sum_{k\in\mathcal{C}}\mathbf{w}_{k}$} and sends them back to each client for the next round. 

\begin{algorithm}[!t]
   \caption{\textbf{GeFL} framework enabling model heterogeneous FL using federated generative models.}
   \label{alg:gefl}
\begin{algorithmic}
    \State (i) \textit{\textbf{generative knowledge aggregation}}
       \For{each round $t=1,2,\dots,T_{KA}$}
        \State GeFL server broadcasts $\mathbf{w}_{g}$ to clients.
            \For{all client $k\in\mathcal{C}$ in parallel}
            \State Initialize local parameters
            $\mathbf{w}_{k}\leftarrow \mathbf{w}_{g}$
            \For{$t=1,...,T_g$}
                \State $\{(x_i, y_i)\}_{i=1}^{B}\sim \mathcal{D}_k$
                \State $\mathbf{w}_{k}\leftarrow \mathbf{w}_{k}-\beta \nabla_{\mathbf{w}_{k}} J_G(\mathbf{w}_{k})$
            \EndFor
            \State Client $k$ sends $\mathbf{w}_{k}$ to the server
            \EndFor
    \State $\mathbf{w}_g \leftarrow \text{Agg}(\{\mathbf{w}_k\}_{k\in\mathcal{C}})$     \Comment{Algorithm~\ref{alg:agg} (Aggregate)}
    \EndFor
    \State (ii) \textbf{\textit{target network training}}
   \For{each round $t=1,2,\dots,T_{TN}$}
    \State GeFL server multicasts $\{\bm{\theta}_{g,m}\}_{m=1}^M$ to clients.
        \For{all client $k\in\mathcal{C}$ in parallel}
            \State Initialize local parameters
            $\bm{\theta}_{k,m}\leftarrow \bm{\theta}_{g,m}$
            \For{$t=1,...,T_s$}
                \State $\{(x_i, y_i)\}_{i=1}^{B}\sim G(z_i|y_i,\mathbf{w}_{g})$
                \State $\bm{\theta}_{k,m}\leftarrow \bm{\theta}_{k,m}-\alpha \nabla_{\bm{\theta}_{k,m}} J(\bm{\theta}_{k,m})$   
            \EndFor   
            \For{$t=1,...,T_r$}
                \State $\{(x_i, y_i)\}_{i=1}^{B}\sim \mathcal{D}_k$
                \State $\bm{\theta}_{k,m}\leftarrow \bm{\theta}_{k,m}-\alpha \nabla_{\bm{\theta}_{k,m}} J(\bm{\theta}_{k,m})$
            \EndFor
        \EndFor
        \State $\bm{\theta}_{g,m} \leftarrow \text{Agg}(\{\bm{\theta}_{k,m_k}\}_{k\in\mathcal{C},m_k=m}), \forall m\in\{1,\dots,M\}$ 
    \EndFor
\end{algorithmic}
\end{algorithm}
\begin{algorithm}[!t]
   \caption{\textbf{Aggregate}}
   \label{alg:agg}
\begin{algorithmic}
    \State \textbf{Input:} set of model parameters $\{\bm{\theta}_k\}_{k \in \mathcal{C}_{\text{agg}}}$ with the same architecture
    \State $\bm{\theta}_g \leftarrow \frac{1}{|\mathcal{C}_{\text{agg}}|} \sum_{k \in \mathcal{C}_{\text{agg}}} \bm{\theta}_k$ \vspace{0.05in}
    \State \textbf{Return} $\bm{\theta}_g$ 
\end{algorithmic}
\end{algorithm}
The trained generative model effectively gathers global knowledge from every client, enabling the training of target networks despite the different model architectures among clients. Specifically, we assume that each client has a model architecture with an index $m\in\{1,\cdots, M\}$ for their target network, where there exist $M$ candidate heterogeneous architectures (e.g., different CNNs or the combination of ResNet \cite{he2016deep}, EfficientNet \cite{tan2019efficientnet}, and MobileNet \cite{howard2017mobilenets}). 

Then, during target network training and refinement, target networks are updated by forwarding synthetic samples and real samples separately.
In each round of target network training, each client $k$ trains its target network $T_{\bm{\theta}_{k,m}}$ using samples generated by FedGen $G(\cdot|y_i, \mathbf{w}_g)$ conditioned on random label $y_i \sim p(y)$, treating them as augmented training samples for multiple local epochs as    
$$
   \min_{\bm{\theta}_{k,m}} \mathbb{E}_{z_i, y_i\sim\mathcal{N}(\mathbf{0}, \mathbf{I}), p(y)}[\,\text{CE}(\rho(T_{\bm{\theta}_{k,m}}(G(z_i|y_i,\mathbf{w}_g))), y_i)\,], 
   \label{eqn:gefl_upd}
$$
where $\rho$ is the softmax function and $\text{CE}$ is the cross-entropy function. After training target networks on generated samples from $G$, the target networks are trained on real data samples subsequently. In our algorithm, trained FedGens offer diverse training synthetic samples, contributing to overcoming model heterogeneity in FL. 
The learning rates are $\beta$ for the generative model $G$ and $\alpha$ for the target networks.
\blue{Then, $\bm{\theta}_{g,m} \leftarrow \text{Agg}(\{\bm{\theta}_{k,m_k}\}_{k\in\mathcal{C}, m_k=m}), \forall m \in \{1,\dots,M\}$ performs model-wise aggregation by averaging local parameters only among clients that share the same model architecture index $m_k = m$, thereby enabling integration of compatible parameters in a heterogeneous setting.}

\subsection{Evaluation of GeFL}
\paragraph{Experimental settings}
We assessed the performance of \textsc{GeFL} on three public datasets: MNIST \cite{lecun1998gradient}, Fashion-MNIST \cite{xiao2017fashion} (FMNIST in short over the paper), and CIFAR10 \cite{krizhevsky2009learning}.
For MNIST, we evaluated our framework using a diverse set of ten heterogeneous convolutional neural networks (CNNs) as target networks. For the other datasets (FMNIST and CIFAR10), distinct sets of CNNs\footnote{\blue{We also conducted experiments with different set of model architectures and the results (Table~\ref{tab:GeFL_large} in Appendix~\ref{appx:commun_cost}) showed the similar tendency.}} were utilized, differing from those employed for MNIST. The model architectures of ten networks are detailed in Appendix. The other hyperparameters used in training target networks and generative models are presented in Appendix (Table~\ref{tab:target-hyperparam} and Table~\ref{tab:gen-hyperparam}) \cite{Kang}. 
Our baseline methods consist of six relevant FL approaches: three from model-homogeneous FL, FedAvg \cite{mcmahan2017communication}, FedProx \cite{Li2020fedprox}, \blue{FedALA \cite{zhang2023fedala}}, and three from model-heterogeneous FL, AvgKD \cite{afonin2022towards}, FedDF \cite{lin2020ensemble} and LG-FedAvg \cite{liang2020think}. 
\blue{Note that FedAvg is applied in grouped settings: clients are partitioned by model architecture (e.g., ResNet18, VGG16), and FL is run within each homogeneous group. For 50 clients, 10 model types are each assigned to 5 clients. In the 10-client case, each client uses a unique model and trains independently without FL. Throughout the paper, FedAvg, LG-FedAvg, FedProx, and FedALA follow these settings.}

\paragraph{Comparison to FL baselines}
The performance of \textsc{GeFL} under model heterogeneity with different types of generative models is detailed in Table \ref{tab:GeFL_perf}. We use the prefix \textit{Fed} to indicate the type of generative model trained in a federated manner and employed for augmenting target network training. The performance was evaluated based on the mean classification accuracy (\%) across 10 heterogeneous target networks throughout the paper.

\textit{\textsc{GeFL} with all types of generative models outperformed other baselines\footnote{\blue{The improvement of \textsc{GeFL} over FedAvg is statistically significant ($p < 0.05$, paired t-test).}} in scenarios where all clients' models have different architectures.} \blue{Although LG-FedAvg relies on a shared feature extractor and cannot be applied in fully heterogeneous settings, it achieved performance comparable to FedAvg. FedProx and FedALA also performed on par with FedAvg.} While AvgKD\footnote{We present the highest performance of AvgKD within a few initial rounds (approximately 30 rounds), as the accuracy rapidly decreases thereafter due the significant heterogeneity in our setting.} allows flexibility in client model architectures, it resulted in lower accuracy than FedAvg on FMNIST and CIFAR10, attributed to its use of averaged logits from clients' models as pseudo-labels, even when some of the models were not adequately trained. {The training process of FedDF involved aggregating logits on a public dataset obtained from trained target networks. However, the performance of FedDF was notably sensitive to the choice of public dataset. We used the public dataset SVHN for MNIST, CIFAR10 for FMNIST, and CIFAR100 for the CIFAR10 dataset. When the training data and the chosen public dataset exhibit considerable dissimilarities in distribution (e.g., FMNIST and CIFAR10), FedDF failed to achieve comparable performance.} In contrast, \textsc{GeFL} consistently showed improvements, benefiting from global knowledge obtained from federated generative models. 

\paragraph{Different generative models}
Generative models learn to capture the representation of training data, aiming to generate samples that adhere to the training data distribution. 
The essential attributes expected from generative models include high-quality sampling, high sample diversity, and fast (computationally inexpensive) sampling. Notable generative models we use have different strengths and weaknesses as follows. Generative adversarial networks (GANs) \cite{radford2016unsupervised, goodfellow2014generative} provide high-quality and fast sampling but are prone to mode collapse. Variational autoencoders (VAEs) \cite{kingma2022autoencoding, pu2016variational} offer fast and diverse sampling, albeit with lower sample quality. Denoising diffusion probabilistic models (DDPMs) \cite{sohl2015deep, ho2020ddpm} deliver diverse and high-quality sampling at the cost of being computationally expensive. Note that $w$ denotes the guidance score for diffusion models \cite{ho2021classifierfree}.

\footnotetext[4]{To evaluate the performance, we conducted training of both FedGens and target networks using varying proportions of the complete dataset (detailed in Table \ref{tab:target-hyperparam}). The observed performance improvements differs by the amount of data utilized in the training process.}
Referring to Table~\ref{tab:sample_qual}, the performance gain (over FedAvg) of \textsc{GeFL} varies across generative models, with no clear correlation between Fr\'{e}chet Inception Distance (FID) score \cite{FID2017} and Inception Score (IS).
This aligns with recent findings suggesting that FID and IS metrics do not reliably predict downstream performance \cite{ravuri2019cas}. The rationale is that while generated samples introducing diversity to the local real data can enhance performance, excessively degraded samples, which deviate significantly from the real data distribution, can negatively impact the results.
This highlights the importance of evaluating \textsc{GeFL} holistically, rather than focusing solely on generative model performance. In general, GANs and diffusion models demonstrate strong \textsc{GeFL} performance. However, GANs tend to be sensitive to hyperparameter selection, whereas diffusion models incur significantly higher computational costs during sample generation.  

\paragraph{Comparison to data augmentation}
\begin{table}[!t]
    \centering
    \begin{tabular}{lccc}
        \toprule
        {Method} & {MNIST} & FMNIST & {CIFAR10} \\ \midrule 
        FedAvg \cite{mcmahan2017communication}         & $92.62_{\pm 0.10}$ & $80.56_{\pm 0.11} $ & $55.65_{\pm 1.06}$ \\
        FedProx \cite{Li2020fedprox}         & $92.45_{\pm 0.64}$ & $80.50_{\pm 0.26}$ & $55.10_{\pm 1.50}$ \\
        FedALA \cite{zhang2023fedala}         & $92.70_{\pm 0.61}$ & $80.03_{\pm 0.}$ & $51.63_{\pm 7.05}$ \\
        LG-FedAvg \cite{liang2020think}       &$ 92.71_{\pm 0.42}$ & $80.61_{\pm 0.09}$ & $54.49_{\pm 2.66}$ \\
        AvgKD \cite{afonin2022towards}  & $92.81_{\pm 0.91}$ &$ 77.12_{\pm 1.58}$ & $26.36_{\pm 4.24}$ \\
        FedDF \cite{lin2020ensemble} & $91.39_{\pm 3.12}$ & $58.81_{\pm 1.60}$ & $57.17_{\pm 1.29} $\\ \midrule
        \textsc{{GeFL}} (FedDCGAN) & $\textbf{95.32}_{\pm 0.38}$ & $\underline{\textbf{83.11}}_{\pm 1.29}$ &$\textbf{58.45}_{\pm 1.21}$ \\
        \textsc{{GeFL}} (FedCVAE)  &$ \textbf{94.46}_{\pm 0.43}$ & $\textbf{82.33}_{\pm 0.20}$ & $\textbf{57.70}_{\pm 1.09}$\\
        \textsc{{GeFL}} (FedDDPM$_{w=0}$) & $\underline{\textbf{96.44}}_{\pm 0.15}$ & $\textbf{82.43}_{\pm 0.39}$ & $\underline{\textbf{{59.36}}}_{\pm 1.58}$\\
        \textsc{{GeFL}} (FedDDPM$_{w=2}$) & $\textbf{95.17}_{\pm 0.42}$ & $\textbf{81.60}_{\pm 0.67}$ &$ {\textbf{58.47}}_{\pm 1.20}$\\
        \bottomrule    \end{tabular}
    \caption[Mean classification accuracy]{\blue{Mean classification accuracy\footnotemark (\%) with 95\% confidence intervals (denoted by $\pm$) for \textsc{GeFL} on MNIST, FMNIST, and CIFAR10 dataset}}
    \label{tab:GeFL_perf}
    \vspace{-0.1in}
\end{table}
\begin{table}[!t]
    \centering
    \begin{tabular}{llcccc}
    \toprule
    Dataset & {Model} & {FID$\downarrow$} & {IS$\uparrow$} &{Perf. gain$\uparrow$} & {MND$\downarrow$} \\ \midrule
    \multirow{4}{*}{MNIST}  & FedDCGAN        & \textbf{7.83} & 1.56 &2.7 & 1.035 \\
    & FedCVAE & 41.41 & 1.55 &1.84 & 0.714 \\
    & FedDDPM$_{w=0}$ & 77.71 & \textbf{1.77} &\textbf{3.82} & \textbf{0.640}\\
    & FedDDPM$_{w=2}$ & 73.40 & 1.60 &2.55 & 0.753 \\ \midrule
    \multirow{4}{*}{FMNIST} & FedDCGAN & \textbf{31.55} & 2.13 &\textbf{2.53} & 0.994 \\ & FedCVAE   &79.87 & 1.86 &1.75 & 0.553 \\
    & FedDDPM$_{w=0}$ & 138.41 & \textbf{2.41} &1.85 & \textbf{0.119}\\
    & FedDDPM$_{w=2}$ &82.00 & 2.18 &0.93 & 0.135\\ \midrule
    \multirow{4}{*}{CIFAR10} & FedDCGAN        & \textbf{23.44} & \textbf{4.13} &2.8 & 0.979\\
    & FedCVAE  & 125.60 & 2.92 &0.15 & \textbf{0.502}\\
    & FedDDPM$_{w=0}$ & 66.99 & 3.79 &\textbf{3.71} & 0.842\\
    & FedDDPM$_{w=2}$ & 53.38 & 3.25 &2.82 & 0.808\\    
    \bottomrule
    \end{tabular}
    \caption{Evaluation of federated generative models where neither FID nor IS exhibits a clear connection to the performance of \textsc{GeFL}. MND here represents the mean nearest neighbor distance ratio, with the value indicating severe memorization if they are larger than 1 on average, as explained in Section~\ref{Sec2.3}.}
    \label{tab:sample_qual}
    \vspace{-0.1in}
\end{table}

\begin{table}[!t]
    \centering
    \resizebox{.45\textwidth}{!}{
    \begin{tabular}{lcc}
        \toprule
        {Method} & {FedAvg} & {\textsc{GeFL} (FedDCGAN)} \\ \midrule
        None        &  ${55.65}_{\pm 0.68}$ & ${58.45}_{\pm 0.49}$ \\ \midrule
        MixUp \cite{zhang2018mixup}       &  ${60.07}_{\pm 1.13}$ & $\mathbf{62.67}_{\pm 0.24}$ \\
        CutMix \cite{yun2019cutmix} &  ${58.95}_{\pm 0.61}$ & ${61.66}_{\pm 0.41}$ \\
        AugMix \cite{hendrycks2019augmix}     &  ${53.96}_{\pm 0.37}$ & ${56.47}_{\pm 0.25}$\\
        AutoAugment \cite{cubuk2018autoaugment} &  ${56.99}_{\pm 0.43}$ & ${59.97}_{\pm 0.38}$ \\
        \bottomrule
    \end{tabular}}
    \caption{{Mean classification accuracy (\%) of \textsc{GeFL} combined with data augmentation on the CIFAR10 dataset}}
    \vspace{-0.2in}
    \label{tab:data_aug}
\end{table}


We conducted a performance comparison of \textsc{GeFL} to existing data augmentation (DA) methods. DA is a widely used technique to enhance the generalization performance, particularly in scenarios with limited training data \cite{zhang2018mixup,yun2019cutmix,hendrycks2019augmix,cubuk2018autoaugment}. Note that DA and \textsc{GeFL} are orthogonal, meaning they can be combined to achieve even greater performance gains. 

To show the effectiveness of \textsc{GeFL} in such scenarios, we provide comparison to MixUp \cite{zhang2018mixup}, CutMix \cite{yun2019cutmix}, AugMix \cite{hendrycks2019augmix}, and AutoAugment \cite{cubuk2018autoaugment} in Table \ref{tab:data_aug}. Here, we used \textsc{GeFL} with FedDCGAN which shows improved accuracy with reasonable computational complexity. Notably, \textsc{GeFL} combined with DA yielded further performance improvements. While DA augments local samples independently at each client, given the inability to share raw data with other clients, \textsc{GeFL} utilizes a generative model with global knowledge across all clients to generate synthetic samples, utilizing more diverse samples than local samples.
Throughout the paper except Table \ref{tab:data_aug}, to provide a focused evaluation of \textsc{GeFL}, we assessed its performance without incorporating DA.

\subsection{Discussions and limitations}\label{Sec2.3}
\blue{In this section, we analyze the limitations of \textsc{GeFL}, which motivate the design of our enhanced framework, \textsc{GeFL-F}. The limitations include potential privacy leakage from synthetic sample generation, performance degradation in large-scale federated setups, and resource overhead of generative models. Unlike \textsc{GeFL}, which generates raw data samples, \textsc{GeFL-F} synthesizes intermediate feature representations, aiming to reduce privacy risks, improve scalability with increasing client counts, and lower computational overhead.}
\paragraph{Privacy concerns}

While generative models offer a promising approach to address model heterogeneity, sharing them between clients and the server introduces notable privacy concerns. Although not sharing target networks alleviates conventional model inversion attacks \cite{zeiler2014visualizing, geiping2020inverting}, privacy risks still arise due to the distribution of a generative model trained across clients' local datasets. \blue{Recent studies have highlighted the phenomenon of \textit{memorization} \cite{van2021on, somepalli2023diffusion, webster2019detecting, sun2023privacy}, where generative models inadvertently reproduce training samples, posing membership inference threats.}

\blue{To evaluate privacy risks, we measure the \textit{mean nearest neighbor distance (MND) ratio}, as shown in Table~\ref{tab:sample_qual}. The MND ratio quantifies how closely synthetic samples resemble training data relative to validation data, thereby assessing potential memorization in the generative model.
This approach closely mirrors Monte Carlo (MC)-based membership inference attacks (MIAs) \cite{hilprecht2019monte}, where membership likelihood is estimated through proximity analysis. Unlike reconstruction-based attacks, which apply only to explicit likelihood models (e.g., VAEs, DDPMs), MND is model-agnostic, providing a consistent privacy metric across GANs, VAEs, and diffusion models used in \textsc{GeFL} and \textsc{GeFL-F}.}

\blue{While the exploration of advanced privacy-preserving techniques such as differential privacy (DP) \cite{abadi2016deep, xie2018differentially} is valuable, they are orthogonal to the primary contribution of this work, which focuses on enabling model-heterogeneous FL via generative models. Exploring these defenses remains an important future direction.}

In this work, the memorization behavior of federated generative models is quantified by the averaged \textit{MND ratio}, formally defined as:
\begin{equation} \mathbb{E}\left[\rho_i \right], \quad \text{where } \rho_i=\frac{\min_{\mathbf{x} \in \mathcal{V}} d(\mathbf{x}i,\mathbf{x})}{\min{\mathbf{x}\in \mathcal{S}} d(\mathbf{x}_i,\mathbf{x})}.\label{eqn:mnd}
\end{equation}
Here, $\rho_i$ represents the ratio between the minimum distance from a training sample $\mathbf{x}_i$ to the nearest sample in the synthetic dataset $\mathcal{S}$, and the minimum distance \blue{to the nearest} sample in the validation set $\mathcal{V}$. Both $\mathcal{S}$ and $\mathcal{V}$ are of equal size ($|\mathcal{S}|=|\mathcal{V}|=600$ in our experiments).

To compute distances $d(\cdot, \cdot)$, we employ the Learned Perceptual Image Patch Similarity (LPIPS) metric \cite{zhang2018the}, which correlates well with human perceptual similarity, providing a more meaningful measurement than traditional pixel-wise metrics like Euclidean distance. The MND values are averaged over 1000 sampled training points.
A $\rho_i$ value greater than $1$ suggests that a real training sample $\mathbf{x}_i$ is closer to synthetic samples than to validation samples, implying a potential memorization risk. Conversely, a $\rho_i$ less than $1$ indicates better generalization, with generated samples being less similar to individual training points.

\paragraph{Scalability across the number of clients}
\begin{figure}[!t]
  \begin{subfigure}[b]{0.325\columnwidth}
    \includegraphics[width=1.0\linewidth]{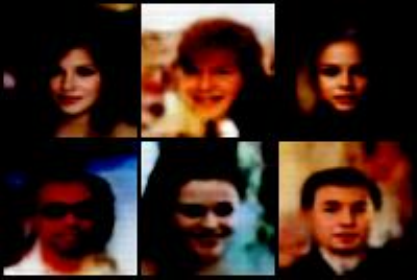}
  \end{subfigure}
  \hfill 
  \begin{subfigure}[b]{0.325\columnwidth}
    \includegraphics[width=1.0\linewidth]{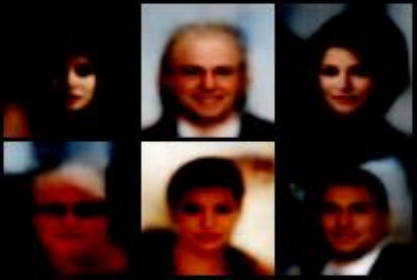}
  \end{subfigure}
  \begin{subfigure}[b]{0.325\columnwidth}
    \includegraphics[width=1.0\linewidth]{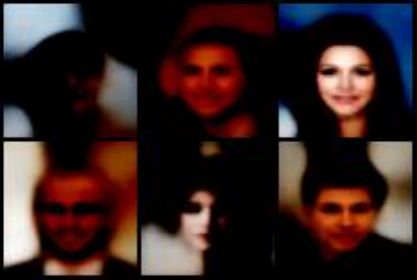} 
  \end{subfigure}
  \\ 
  \begin{subfigure}[b]{0.325\columnwidth}
    \includegraphics[width=1.0\linewidth]{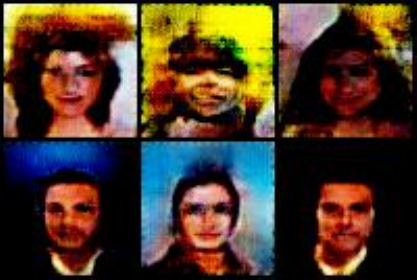}
    \caption{10 clients}
  \end{subfigure}
  \hfill 
  \begin{subfigure}[b]{0.325\columnwidth}
    \includegraphics[width=1.0\linewidth]{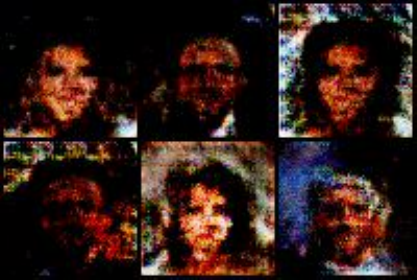}
    \caption{50 clients}
  \end{subfigure}
  \begin{subfigure}[b]{0.325\columnwidth}
    \includegraphics[width=1.0\linewidth]{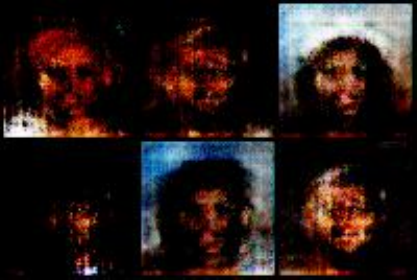}
    \caption{100 clients}
  \end{subfigure}
  \caption[Generated images]{Generated images from FedCVAE (top two rows) and FedDCGAN (bottom two rows) which are trained across the different number of clients\footnotemark}
   \label{fig:gefl_scale}
   \vspace{-0.1in}
\end{figure}

FL often struggles with distributed data, as an increase in distributed data causes greater divergence in gradients. \textsc{GeFL} also shows the vulnerabilities to the distributed data. Figure~\ref{fig:scalability} demonstrates that as the number of clients increases, both performance and the gain of \textsc{GeFL} compared to baseline decreases. 
It is due to the reduced quality of generated samples from federated generative models as Figure~\ref{fig:gefl_scale}.
\blue{As the number of clients increases, each client holds fewer samples, making it more difficult to train local models effectively. This data sparsity degrades the quality of generated images, as shown in Figure~\ref{fig:gefl_scale}, due to the increased challenge of achieving stable convergence in the federated generative model, consistent with prior observations \cite{zhao2018federated}.}
\footnotetext[5]{\blue{Throughout the paper, the total dataset size is fixed and evenly divided across clients. For example, with 100 clients, each holds 60 samples; with 10 clients, each holds 600 samples.}}


\begin{table}[t]
    \centering
    \resizebox{0.49\textwidth}{!}{
    \begin{tabular}{lccc} 
    \toprule
    {} & {\textbf{FedDCGAN}} & {\textbf{FedCVAE}} & {\textbf{FedDDPM}}  \\ \midrule
       \# Total Params.  & 5.72M & 22.35M & 7.569M\\ 
       \# Generator Params.  & 3.077M & 11.17M & 7.569M\\ \midrule
       Training Cost (MACs) & 347.74M & 169.41M & 2.21G \\
       Sampling Cost (MACs) & 277.74M & 135.29M & 220.87G\\
       \bottomrule 
    \end{tabular}
    }
    \caption{Cost comparison of generative models in \textsc{GeFL}}
    \label{tab:cost_gefl} 
    \vspace{-0.15in}
\end{table}

\paragraph{Resource comparison} \blue{To quantify the resource of our proposed FedGens, we analyze both communication and computational costs. The communication and memory cost associated with training generative models is measured by the total number of parameters that each client must transmit to the central server during federated updates.
The computational costs are broken down into two metrics: training cost and sampling cost. These are measured using multiply-accumulate operations (MACs) required during a single forward pass\footnote{\blue{The reported MACs represent the dominant computational operations involved in training and sampling generative models.}}. Training cost captures the resources consumed during the optimization of all components of the generative model, while sampling cost reflects the computational load required solely for producing synthetic data.}
Table \ref{tab:cost_gefl} provides a detailed comparison of the parameter sizes and computational costs associated with various generative models in \textsc{GeFL} on MNIST and FMNIST, with the goal of understanding their efficiency and resource demands. Note that generative models have different hyperparameters on CIFAR10, but have same tendency. The total parameters include all the components of each model that are essential for the operation of the models, such as the generator and discriminator in FedDCGAN and the encoder and decoder in FedCVAE. It is also important to note that the number of parameters directly impacts the communication cost, as participating clients must transmit these parameters to the server. On the other hand, the generator parameters refer specifically to the components directly involved in the image generation process—for example, the generator in FedDCGAN or the decoder in FedCVAE. 
\blue{It is worth noting that FedDDPM incurs a significantly higher sampling cost than other models, due to its iterative denoising process. This computational overhead may limit its practicality on resource-constrained edge devices.}

These analyses underscore the importance of evaluating model gain, privacy considerations, computational cost, and communication cost when selecting an appropriate generative model for \textsc{GeFL}.

\begin{figure*}[!t]
    \centering
    \includegraphics[width=0.93\textwidth]{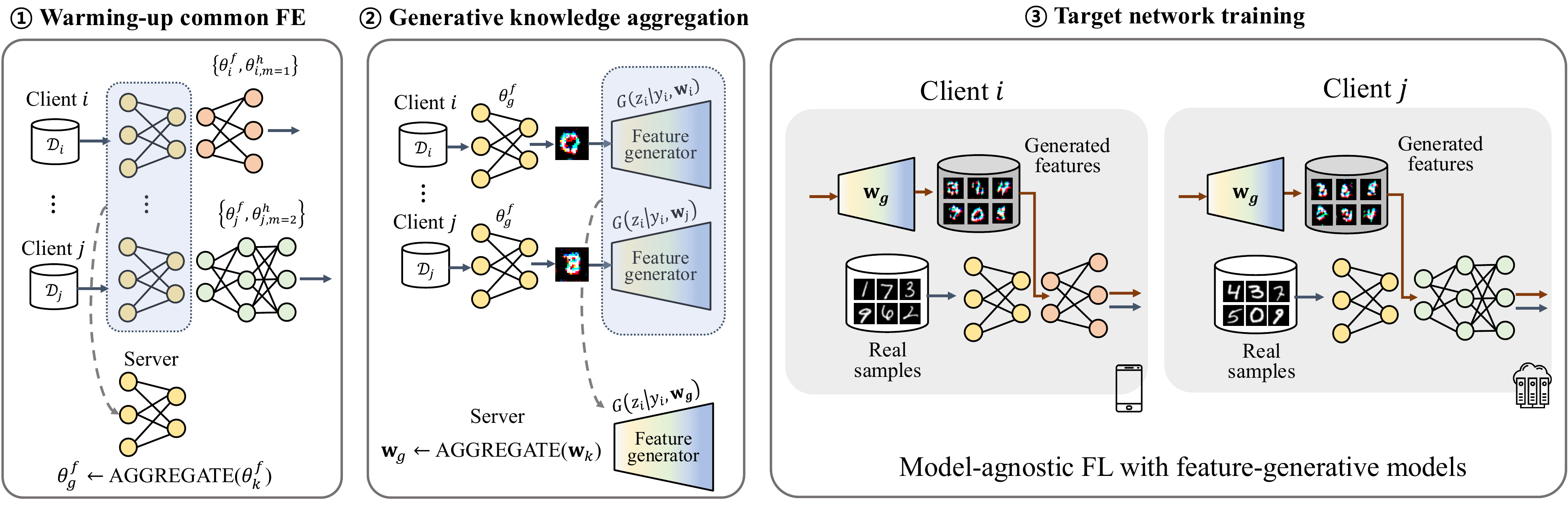}
    \caption{\blue{\textbf{Illustration of \textsc{GeFL-F} algorithm}. Each target network is decoupled into a common feature extractor and a heterogeneous header. There are three main stages in \textsc{GeFL-F} including (i) \textit{warming-up phase} for training a common feature extractor, and (ii) \textit{generative knowledge aggregation} which trains feature-generative models using warmed-up feature extractor, and (iii) \textit{target network training}, especially the heterogeneous headers, using real samples and generated features.}}
    \label{fig:gefl-f}
    \vspace{-0.1in}
\end{figure*}

\section{Feature Generative Model-Aided FL\label{sec:geflf}}

In this section, we introduce \textsc{GeFL-F}, an extension of \textsc{GeFL}, aimed at addressing key challenges such as privacy preservation, scalability, and communication efficiency \blue{stated in Section \ref{Sec2.3}}. This leverages feature-generative models to tackle these issues effectively.
We use the suffix \textit{-F} to indicate the feature-generative models (e.g., FedDCGAN-F, FedCVAE-F).

Our primary motivation is to design a lightweight federated generative model that preserves privacy on training data of individual clients while simultaneously aggregating global knowledge to benefit the training of heterogeneous networks. To achieve this, we incorporate feature-generative models that are trained on features, which represent the output of a common feature extractor comprising a few layers of the target network. Given the lower resolution of features compared to the original images, sharing feature-generative models offers a potential to mitigate privacy concerns \cite{zeiler2014visualizing, mahendran2015understanding, dosovitskiy2016inverting}, and further reduces the required size of generative models to learn a global knowledge \cite{brock2018large}. As information passes through the successive layers of a convolutional layers, the focus shifts from fine-grained pixel-level details to more abstract, high-level features. This process results in a trade-off: while high-level semantic content is preserved and enhanced, detailed pixel information is progressively discarded or generalized \cite{zeiler2014visualizing, gatys2016image}.
These findings motivate the inherent design of feature extract to prioritize semantic understanding, which is advantageous for preserving privacy.


\begin{algorithm}[t]
    \small
   \caption{\textbf{\textsc{GeFL-F}} framework enabling model heterogeneous FL using federated feature-generative models.}
   \label{alg:gefl-f}
\begin{algorithmic}
   \State (i) \textbf{{\textit{warming-up common feature extractor}}}
   \For{each round $t=1,2,\dots,T_{FE}$}
   \State The server multicasts $\{\bm{\theta}_{g}^{f},\bm{\theta}_{g,m}^{h}\}_{m=1}^M$ to clients.
   \For{each client $k\in\mathcal{C}$ in parallel}
   \State Initialize local parameters
   $\bm{\theta}_{k,m}\leftarrow \bm{\theta}_{g,m}$
   \For{$ \text{each local epoch } t=1,...,T_w$}
   \vspace{0.01in}
   \State $\{(x_i, y_i)\}_{i=1}^{B}\sim \mathcal{D}_k$
   \State $\bm{\theta}_{k,m}\leftarrow \bm{\theta}_{k,m}-\alpha \nabla_{\bm{\theta}_{k,m}} J(\bm{\theta}_{k,m})$
   \EndFor
   \EndFor
   \State $\bm{\theta}_{g}^{f}\leftarrow \text{Agg}(\{\bm{\theta}_{k}^{f}\}_{k\in\mathcal{C}})$ \Comment{Algorithm~\ref{alg:agg}}
   \State $\bm{\theta}_{g,m}^{h}
\hspace{-0.04in}\leftarrow\hspace{-0.04in}
\text{Agg}(\hspace{-0.01in}\{\hspace{-0.01in}\bm{\theta}_{k,m_k}^{h}\hspace{-0.02in}\}_{k\in\mathcal{C},m_k=m}), \forall m\in\hspace{-0.02in}\{\hspace{-0.01in} 1,...,M \hspace{-0.01in}\}$
   \State \vspace*{-0.5\baselineskip}
   \EndFor
   \State (ii) \textbf{\textit{generative knowledge aggregation}}
   \For{each round $t=1,2,\dots,T_{KA}$}
   \State The server broadcasts $\mathbf{w}_{g}$ to clients.
   \For{each client $k\in\mathcal{C}$ in parallel}
   \State Initialize local parameters
   $\mathbf{w}_{k}\leftarrow \mathbf{w}_{g}$
   \For{$\text{each local epoch } t=1,...,T_g$}
   \vspace{0.02in}
   \State $\{(x_i, y_i)\}_{i=1}^{B}\sim \mathcal{D}_k$
   \State $\{x_i\}_{i=1}^{B}\leftarrow \{F(x_i)\}_{i=1}^{B}$ 
   \State $\mathbf{w}_{k}\leftarrow \mathbf{w}_{k}-\beta \nabla_{\mathbf{w}_{k}} J_G(\mathbf{w}_{k})$
   \EndFor
   \EndFor
   \State $\mathbf{w}_{g}\leftarrow \text{Agg}(\{\mathbf{w}_{k}\}_{k\in\mathcal{C}})$
   \EndFor
   \State (iii) \textbf{\textit{target network training}}
   \For{each round $t=1,2,\dots,T_{TN}$}
   \State The server multicasts $\{\bm{\theta}_{g,m}^{h}\}_{m=1}^M$ to clients.
   \For{each client $k\in\mathcal{C}$ in parallel}
   \State Initialize local parameters
   $\bm{\theta}_{k,m}^{h}\leftarrow \bm{\theta}_{g,m}^{h}$
   \For{$\text{each local epoch } t=1,...,T_s$} 
   \vspace{0.02in}
   \State $\{(x_i^\text{feat}, y_i)\}_{i=1}^{B}\hspace{-0.01in}\sim\hspace{-0.013in}G(z_i|y_i,\mathbf{w}_{g}) $ \Comment{synthetic feature} 
   \vspace{0.02in}
   \State where $z_i \sim\mathcal{N}(\mathbf{0}, \mathbf{I}), \, y_i \sim p(y)$
   \vspace{0.02in}
   \State $\bm{\theta}_{k,m}^{h}\leftarrow \bm{\theta}_{k,m}^{h}-\alpha \nabla_{\bm{\theta}_{k,m}^{h}} J(\bm{\theta}_{k,m})$   
   \EndFor   
   \For{$ \text{each local epoch } t=1,...,T_r$}
   \vspace{0.02in}
   \State $\{(x_i^\text{real}, y_i)\}_{i=1}^{B}\sim \mathcal{D}_k$ \Comment{real sample}
   \vspace{0.02in}
   \State $\bm{\theta}_{k,m}^{h}\leftarrow \bm{\theta}_{k,m}^{h}-\alpha \nabla_{\bm{\theta}_{k,m}^{h}} J(\bm{\theta}_{k,m})$
   \EndFor
   \EndFor
   \State $\bm{\theta}_{g,m}^{h}\hspace{-0.04in}\leftarrow\hspace{-0.04in}\text{Agg}(\hspace{-0.01in}\{\hspace{-0.01in}\bm{\theta}_{k,m_k}^{h}\hspace{-0.02in}\}_{k\in\mathcal{C},m_k=m}), \forall m\in\{ 1,...,M \}$
    \EndFor
\end{algorithmic}
\end{algorithm}
\subsection{Framework}

An overview of \textsc{GeFL-F} is provided in Figure~\ref{fig:gefl-f}. Each client has its target network from $M$ candidate models, each consisting of a common feature extractor and a unique heterogeneous header. The parameters are denoted as $\bm{\theta}_m = \{\bm{\theta}^f, \bm{\theta}^h_m\}$, where $\bm{\theta}^f$ and $\bm{\theta}_m^h$ are the parameters of common feature extractor and heterogeneous header of $m$-th candidate target network, respectively. 

Several FL studies have explored the use of common feature extractors \cite{liang2020think, zhu2021datafree, wu2021fedcg, jang2022fedclassavg, mori2022continual}. They facilitate learning of global knowledge while sharing only a limited number of parameters. Our framework aligns with these practices and extends them by incorporating the use of feature-generative models. The feature-generative models learn from the feature representations (extracted by shared feature extractors), thereby capturing the collective knowledge distributed across all participating clients.

Specifically in Algorithm \ref{alg:gefl-f}, \textsc{GeFL-F} is structured into four main stages: 
(i) \textit{warming-up phase} for training the common feature extractor $F$,  
(ii) \textit{generative knowledge aggregation} by training feature-generative model $G_F$ using the warmed-up $F$, 
(iii) \textit{target network training} on synthetic features with global knowledge from $G_F$, and
(iv) \textit{target network refinement} using real data. 
Here, $J$ represents the cross entropy loss function, and $J_G$ corresponds to the loss function associated with the respective generative models. The learning rates are $\beta$ for the feature-generative model $G_F$ and $\alpha$ for the target networks. 

During the warming-up phase, common feature extractors are aggregated across all clients, while heterogeneous headers are aggregated only across the clients with the same header. In the generative knowledge aggregation stage, feature-generative models are trained to capture the representation of intermediate features obtained by forwarding real samples through the trained common feature extractor. Then, during target network training and refinement, headers of target networks are updated by forwarding synthetic features and real samples separately.

\captionsetup[subfigure]{font=small,skip=0pt}
\begin{figure*}[t]
\centering
  \begin{subfigure}{0.86\columnwidth}
    \includegraphics[width=\linewidth]{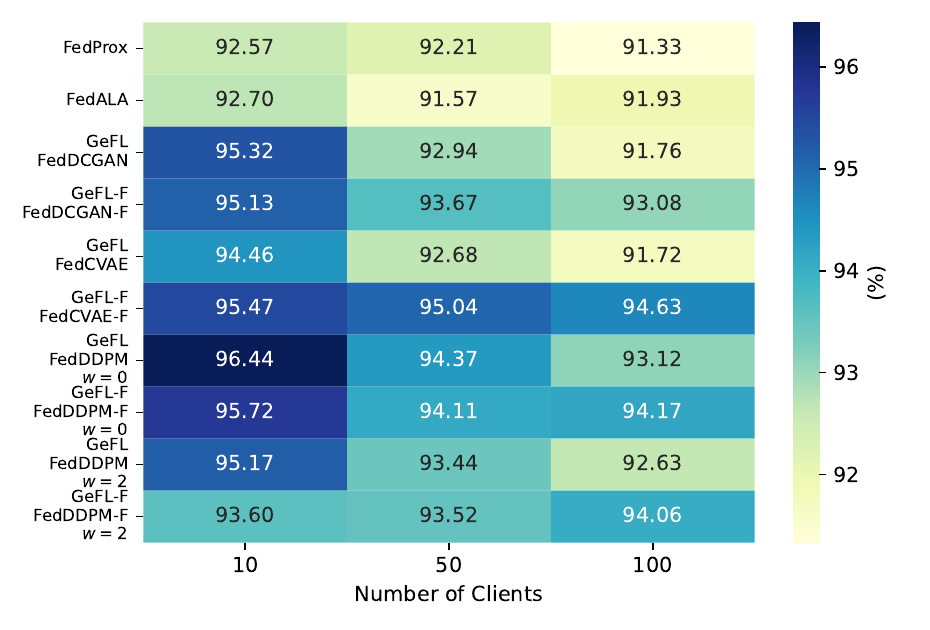}
    \caption{MNIST}
    \label{fig:scalability-mnist}
  \end{subfigure}
  \begin{subfigure}{0.86\columnwidth}
    \includegraphics[width=\linewidth]{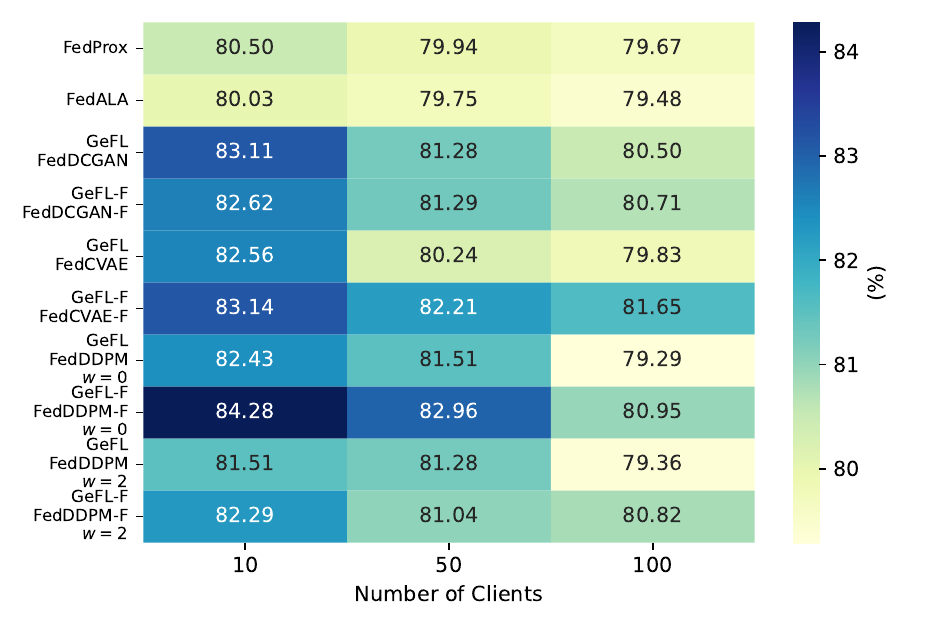}
    \caption{FMNIST}
    \label{fig:scalability-fmnist}
  \end{subfigure}
  \begin{subfigure}{0.86\columnwidth}
    \includegraphics[width=\linewidth]{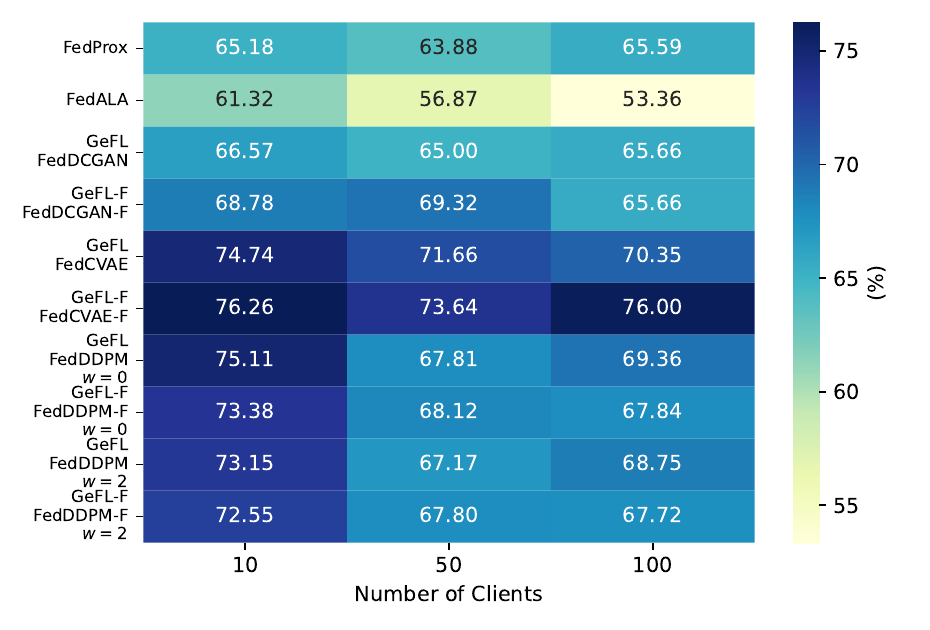}
    \caption{SVHN}
    \label{fig:scalability-svhn}
  \end{subfigure}
  \begin{subfigure}{0.86\columnwidth}
    \includegraphics[width=\linewidth]{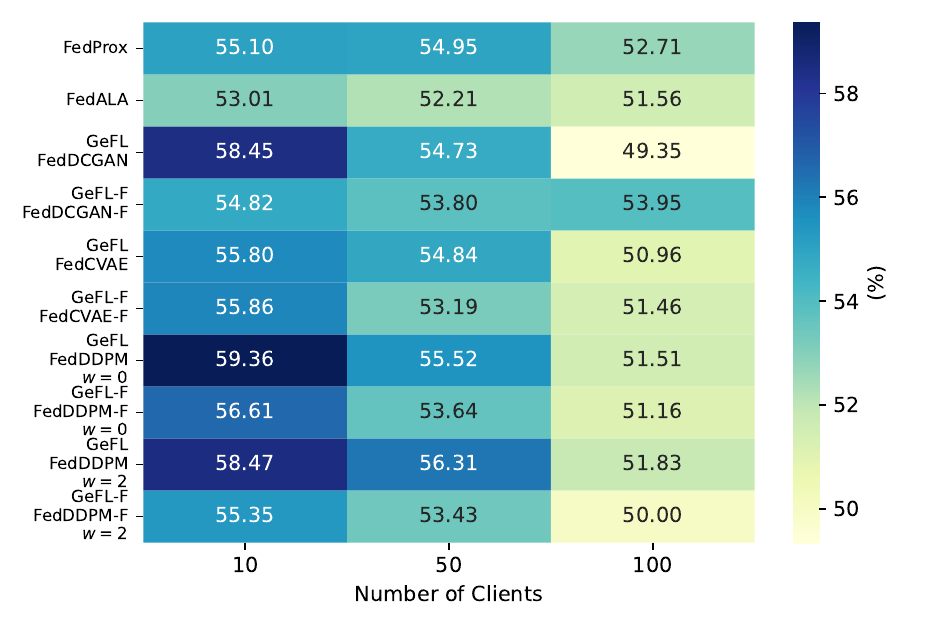}
    \caption{CIFAR10}
    \label{fig:scalability-cifar}
  \end{subfigure}
  \caption{\blue{\textbf{Scalability in client numbers} of \textsc{GeFL} and \textsc{GeFL-F}. \textsc{GeFL-F} exhibits less performance degradation in a large number of clients compared to \textsc{GeFL}.\label{fig:scalability}}}
  \vspace{-0.15in}
\end{figure*}

\begin{figure}[t]
\centering
\begin{subfigure}{0.2\textwidth}
    \centering
    \includegraphics[width=\textwidth]{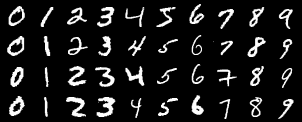}
    \caption{Real samples}
    \label{fig:real_mnist}
\end{subfigure}\\
\begin{subfigure}{0.2\textwidth}
    \centering
    \includegraphics[width=\textwidth]{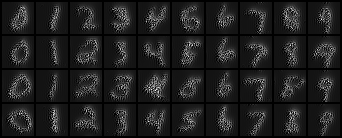}
    \caption{FedDCGAN-F}
    \label{fig:ifeddcganf}
\end{subfigure}
\begin{subfigure}{0.2\textwidth}
    \centering
    \includegraphics[width=\textwidth]{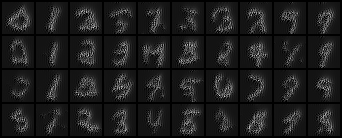}
    \caption{FedCVAE-F}
    \label{fig:ifedcvaef}
\end{subfigure}
\begin{subfigure}{0.2\textwidth}
    \centering
    \includegraphics[width=\textwidth]{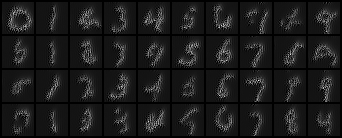}
    \caption{FedDDPM-F ($w=0$)}
    \label{fig:ifedddpmfw0}
\end{subfigure}
\begin{subfigure}{0.2\textwidth}
    \centering
    \includegraphics[width=\textwidth]{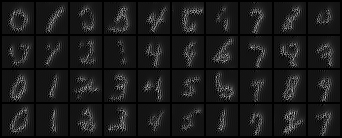}
    \caption{FedDDPM-F ($w=2$)}
    \label{fig:ifedddpmfw2}
\end{subfigure}
\caption{\textbf{Reconstructed samples generated by feature-generative models by model-inversion}}
\label{fig:mnist-model-inv}
\vspace{-0.2in}
\end{figure}

\begin{figure*}[t!]
    \centering
    \includegraphics[width=1.35\columnwidth]{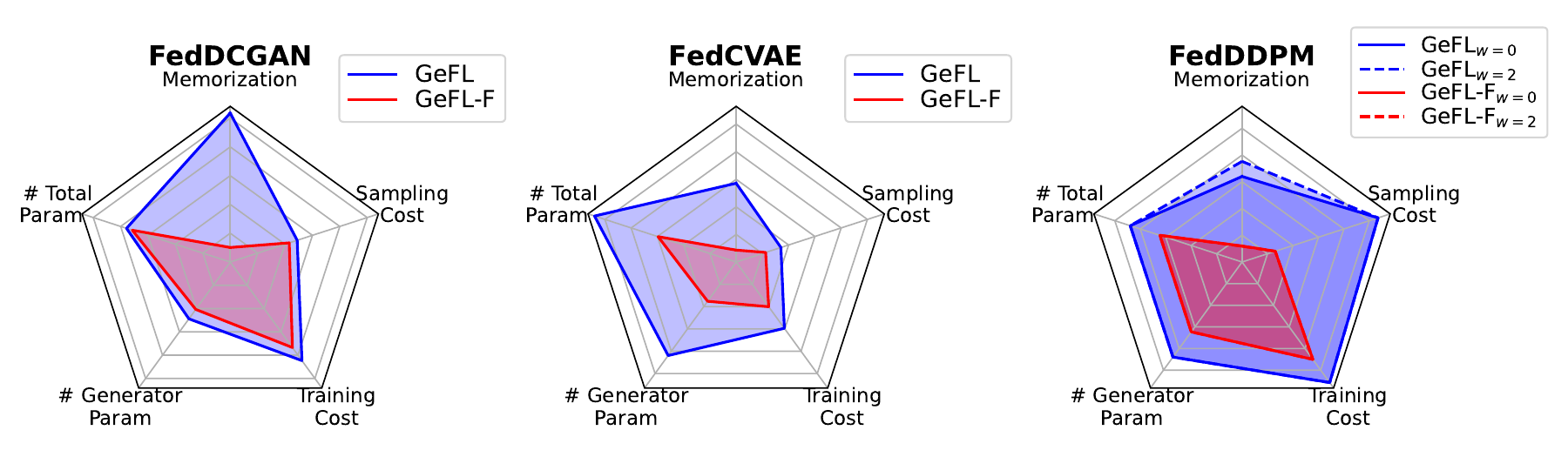}
    \caption{\textbf{Comparison of privacy, communication and computational costs} in \textsc{GeFL} and \textsc{GeFL-F}. Lower values indicate better conditions for each component. Exact values are provided in Table~\ref{tab:sample_qual}, Table~\ref{tab:cost_gefl}, and Table~\ref{tab:app-gen-cost}.}
    \label{fig:priv_scalab}
    \vspace{-0.2in}
\end{figure*}

\begin{figure*}[!t]
\centering
  \begin{subfigure}[t]{0.3\textwidth}
    \includegraphics[width=1\textwidth]{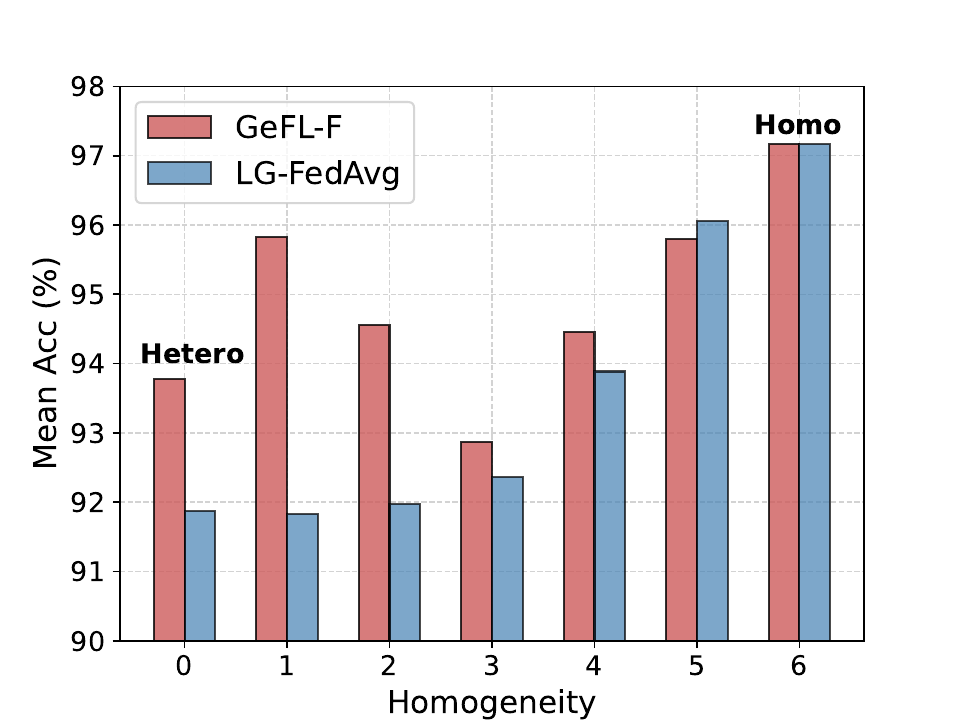}
    \caption{10 clients}
    \label{fig:perf_gefl-f-a}
  \end{subfigure}
    \begin{subfigure}[t]{0.3\textwidth}
    \includegraphics[width=1\textwidth]{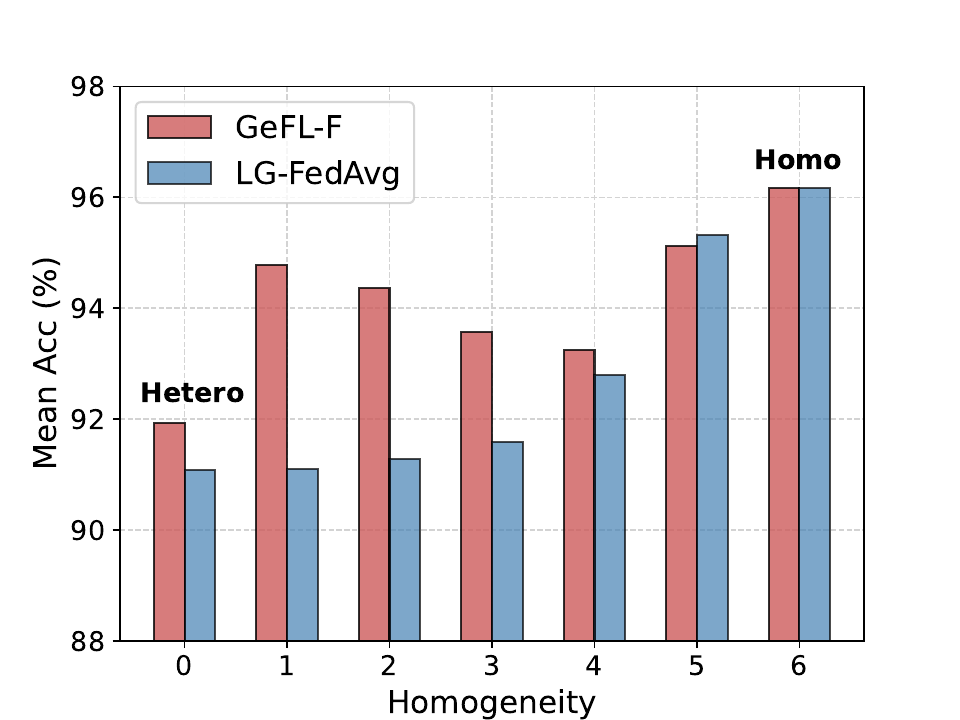}
    \caption{50 clients}
  \end{subfigure}
  \begin{subfigure}[t]{0.3\textwidth}
    \includegraphics[width=\textwidth]{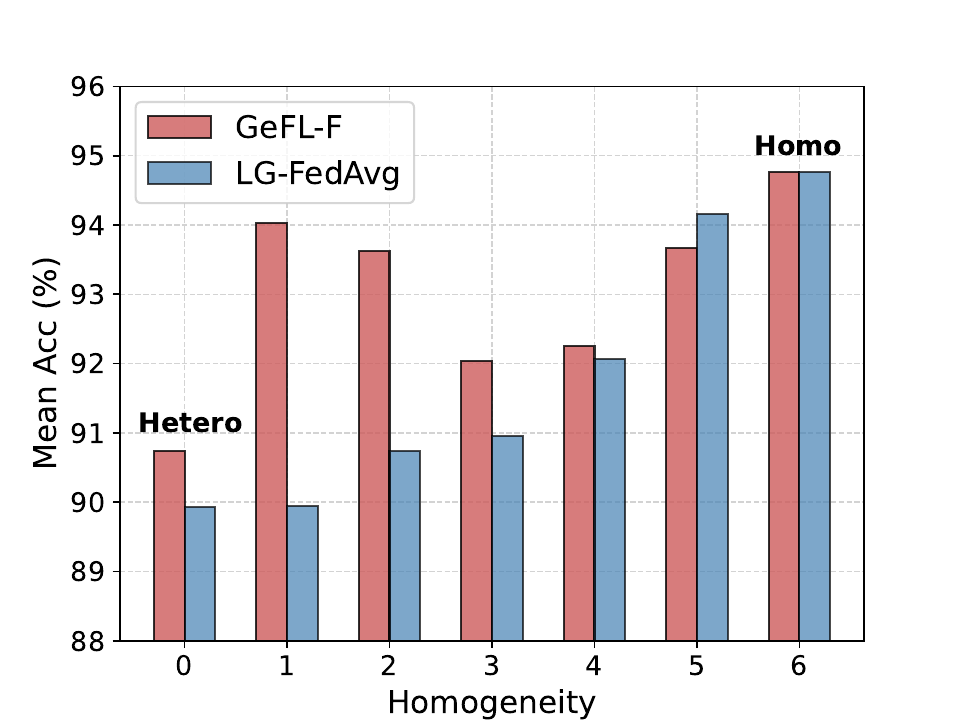}
    \caption{100 clients}
  \end{subfigure}
  \caption{\textbf{Mean classification accuracy of \textsc{GeFL-F} depending on the homogeneity} of model architectures on MNIST dataset. \textsc{GeFL-F} shows the same trend across the homogeneity regardless of the number of clients.}
  \label{fig:perf_gefl-f}
  \vspace{-0.3cm}
\end{figure*}

\subsection{Evaluation}
\paragraph{Experimental settings}

Our evaluation of \textsc{GeFL-F} was conducted on MNIST \cite{lecun1998gradient}, FMNIST \cite{xiao2017fashion}, SVHN \cite{svhn} and CIFAR10 \cite{cifar10} datasets.
We employed ten different CNNs as our target networks, each consisting of a common feature extractor and heterogeneous header. The feature extractor is designed with a single convolutional layer followed by batch normalization and pooling layers, while the headers are ten distinct CNNs except the feature extractor component.
The feature extractor results in features with a lower dimensionality and the reduced size of features leads to a reduction in the size of feature-generative models. 

The communication rounds for warming up the feature extractor (\(T_{FE}\)) are set to 20 for MNIST and FMNIST, and 50 for SVHN and CIFAR10 to address their increased complexity. The feature-generative model is trained for \(T_{KA} = 100\) rounds across MNIST, FMNIST, and SVHN and \(T_{KA} =200\) rounds for CIFAR10. Target networks are trained for \(T_{TN} = 50\) rounds for MNIST, FMNIST, and SVHN, while \(T_{TN} = 100\) rounds are used for CIFAR10. Additional experimental details are provided in the Appendix \cite{Kang}.

\paragraph{Performance, scalability, and privacy}

\blue{Figure~\ref{fig:scalability} illustrates the scalability performance of \textsc{GeFL} and \textsc{GeFL-F}. Compared to \textsc{GeFL}, FedProx, and FedALA, \textsc{GeFL-F} demonstrates improved robustness against increasing numbers of clients, effectively mitigating the performance degradation observed in large-scale federated settings.}
Among the generative models, \textsc{GeFL-F} with FedCVAE-F demonstrates superior effectiveness, particularly due to the reduced impact of image blurriness—an inherent limitation of VAE—when generating small-resolution features. 
However, for the CIFAR10 dataset, \textsc{GeFL-F} does not perform as effectively as it does for other datasets. This can be attributed to the lower performance of the feature extractor and target networks on CIFAR10. The degraded feature outputs from the feature extractor result in feature-generative models aggregating less effective global knowledge, which ultimately hinders the training of target networks. These observations highlight the critical role of both feature extractor quality and dataset characteristics in determining the efficacy of \textsc{GeFL-F}.

\begin{table}[!t]
    \centering
    \resizebox{0.49\textwidth}{!}{
    \begin{tabular}{lccc}
   \toprule
     {} & {\textbf{FedDCGAN-F}} & {\textbf{FedCVAE-F}} & {\textbf{FedDDPM-F}}  \\ \midrule
    \# Total Params.  & 5.191M & 5.543M & 4.428M \\ 
    \# Generator Params & 2.554M & 2.754M & 4.428M  \\ \midrule
    Training Cost (MACs) & 159.98M & 31.88M & 503.16M  \\ 
    Sampling Cost (MACs) & 142.25M & 25.428M & 20.124G \\ 
    \midrule
    MND ratio & 0.101 & 0.108 & \makecell{0.118 $(w=0)$ \\ 0.120 $(w=2)$} \\ 
    \bottomrule 
    \end{tabular}
    }
    \caption{Comparison of generative models in \textsc{GeFL-F}}
    \label{tab:app-gen-cost}
    \vspace{-0.2in}
\end{table}

Furthermore, \textsc{GeFL-F} alleviates the privacy problem by not sharing a federated generative model, but sharing the feature extractor and feature-generative model. We measured MND in our framework, \textsc{GeFL} and \textsc{GeFL-F}, in terms of \textit{memorization} to address the privacy.
While privacy leakage is solely due to the memorization of shared generative models in \textsc{GeFL}, where the target networks are not shared, in \textsc{GeFL-F}, a common feature extractor and feature-generative model are shared between the server and clients in the FL pipeline. A promising way that attacker can take using trained feature extractor and feature-generative model is to generate synthetic features from generative model and apply model inversion using the shared feature extractor.

We evaluated the vulnerability of \textsc{GeFL-F} in such scenario given a white-box feature extractor. We reconstructed the images from generated features by model inversion \cite{zeiler2014visualizing,dosovitskiy2016inverting}. The reconstructed images on MNIST are presented in Figure~\ref{fig:mnist-model-inv}.
Subsequently, we compare the original real images with the reconstructed ones by measuring the MND ratio, as we did in \textsc{GeFL}. Referring to MND ratio in Table \ref{tab:app-gen-cost} and Table \ref{tab:sample_qual}, it is observed that incorporating feature-generative models mitigates privacy preservation compared to \textsc{GeFL}, even though it shares a common feature extractor. 
Figure~\ref{fig:priv_scalab} illustrates that incorporating feature-generative models in \textsc{GeFL-F} led to reduced memorization (i.e., improved privacy preservation) and decreased costs such as reduced number of parameters, training cost and sampling cost compared to \textsc{GeFL}.

\paragraph{Trade-off across homogeneity levels}
\begin{figure}
    \centering
    \includegraphics[width=0.6\linewidth]{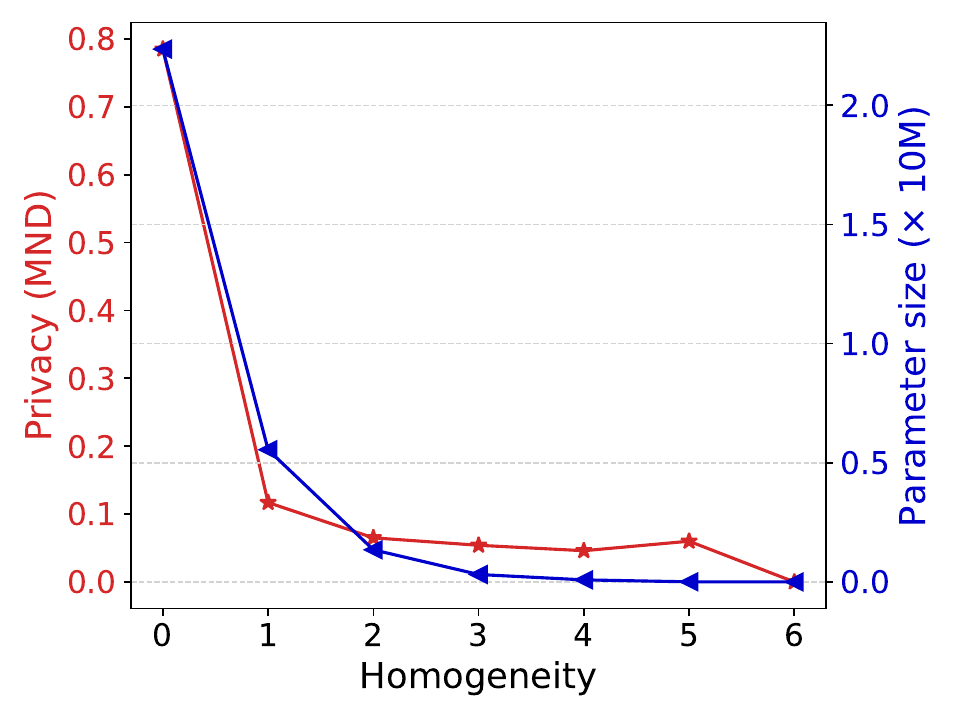}
    \caption{\textbf{Privacy and the number of parameters of generative model of \textsc{GeFL-F}} across the homogeneity level of model architectures on MNIST dataset}
    \label{fig:perf_gefl-f-b}
    \vspace{-0.25in}
\end{figure}

We assessed the performance of \textsc{GeFL-F} (FedCVAE-F) by varying the feature extractor size and the dimensionality of the generated features, transitioning from heterogeneous to homogeneous models. In Figure~\ref{fig:perf_gefl-f}, homogeneity level (HL) 0 corresponds to \textsc{GeFL}, where all client models differ in architecture, while HL 6 represents complete homogeneity, with all models sharing the same architecture. As the homogeneity level increases, the dimensionality of the generator's output decreases. The resolution of generated features for each HL is detailed in Table~\ref{tab:feat-size-homo}. At HL 0, no common feature extractor exists, and only model headers are trained. In contrast, at HL 6, the feature extractor serves as the entire model, requiring only the feature extractor to be trained.

In Figure~\ref{fig:perf_gefl-f}a, while the absolute accuracy of LG-FedAvg increased due to the larger shared feature extractor among clients, \textsc{GeFL-F} exhibited performance gains under highly heterogeneous settings (HL 0-4) due to the augmented generated features used in training. 
To elaborate further on \textsc{GeFL-F} performance, HL 1 exhibited superior performance compared to HL 0, followed by a gradual decline until HL 3, and a subsequent increase up to HL 6.
(i) GeFL exhibits gains on high levels of model heterogeneity (HL 0), but large output dimension of the generator is needed. 
(ii) Lowering level of heterogeneity to HL 1 enables more well-trained generator, providing both a reduced output dimension and considerably high-quality synthetic samples.
(iii) However, HL 2 and 3 suffer from reduced feature diversity in synthetic samples due to its excessively low dimension. It leads to performance degradation due to insufficient information being shared among the clients.
(iv) Increasing homogeneity to HL 6 expands the size of shared feature extractor, enhancing knowledge collection by sharing model parameters.


We also present the results of \textsc{GeFL-F} across various HLs and client numbers on MNIST, depicted in Figure~\ref{fig:perf_gefl-f}. \textsc{GeFL-F} showed the performance gain and has the same trend regardless of the number of clients. Notably, \textsc{GeFL-F} consistently achieves the highest performance gain at a homogeneity level of 1 compared to the baseline. 

Additionally, Figure~\ref{fig:perf_gefl-f-b} demonstrates that \textsc{GeFL-F} provides increased privacy benefits and reduced generative model parameter size as the homogeneity increases (i.e., as the common feature extractor gets larger). This privacy enhancement aligns with the finding that features passing through a larger feature extractor contain less information about the original samples \cite{zeiler2014visualizing, gatys2016image}. 


\section{Insights and Findings\label{sec:insights}}
\begin{figure}[!t]
  \begin{subfigure}[b]{0.495\columnwidth}
    \includegraphics[width=1.13\linewidth]{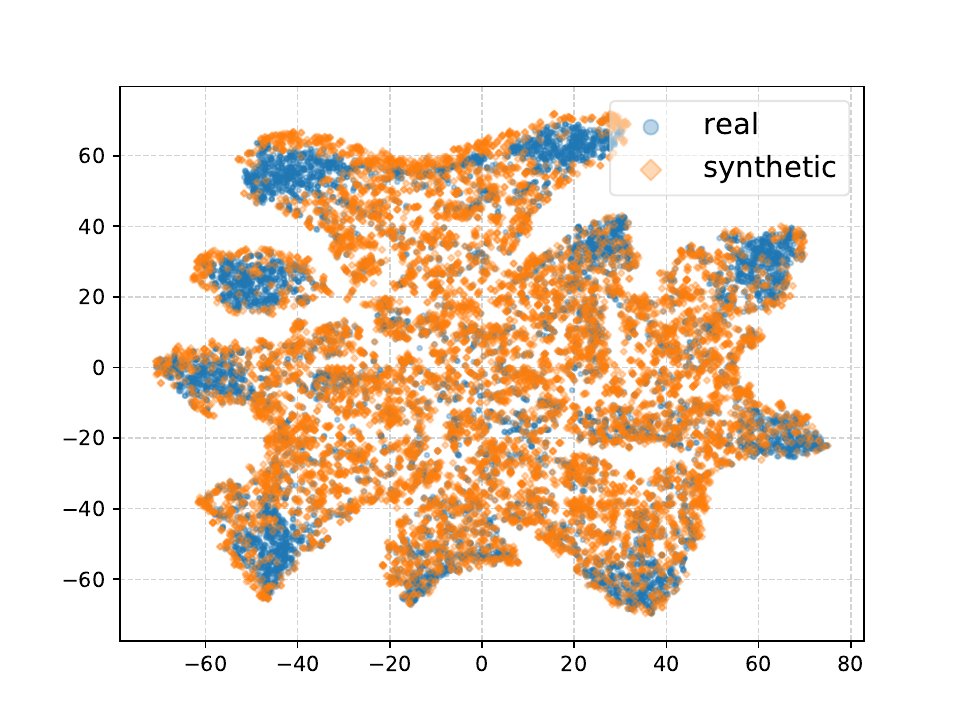}
    \caption{FedDDPM ($w=0$)}
    \label{fig:ddpm_guide-a}
  \end{subfigure}
  \begin{subfigure}[b]{0.495\columnwidth}
    \includegraphics[width=1.13\linewidth]{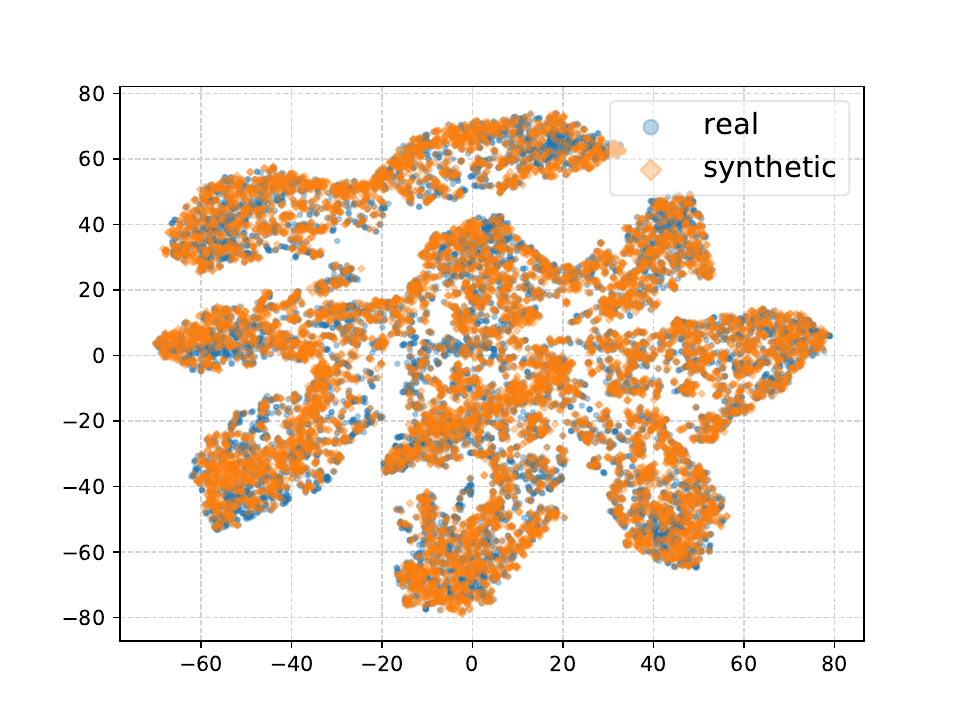}
    \caption{FedDDPM ($w=2$)}
    \label{fig:ddpm_guide-b}
  \end{subfigure}
  \caption{\textbf{t-SNE visualization of learned representation} of CIFAR10 synthetic images generated by FedDDPM with guidance scores $w=0$ and $w=2$}
  \label{fig:ddpm_guide}
\end{figure}

\begin{table}[!t]
    \centering
    \begin{tabular}{llcc} 
    \toprule
    {Dataset} & Train & \textbf{Update} & \textbf{Freeze} \\ \midrule
       \multirow{2.5}{*}{MNIST} & Syn & $\textbf{83.49}_{\pm 0.38}$ & ${81.70}_{\pm 3.79}$\\ \cmidrule(l){2-4} 
                              & Real+Syn & $\textbf{95.32}_{\pm 0.15}$ & $94.98_{\pm 0.13}$ \\ \midrule
       \multirow{2.5}{*}{FMNIST} & Syn & $\textbf{81.83}_{\pm 0.47}$ & $68.27_{\pm 7.11}$ \\ \cmidrule(l){2-4} 
                              & Real+Syn & $\textbf{83.18}_{\pm 0.54}$ & $81.16_{\pm 0.13}$ \\
       \bottomrule 
    \end{tabular}
    \caption{\textbf{Comparison between two settings of updating FedDCGAN} when training target networks on synthetic samples (Syn) and through \textsc{GeFL} (Real+Syn): (i) updating GAN during target network training, and (ii) freezing it after the knowledge aggregation stage}
    \label{tab:gan-upd} 
\end{table}
\subsection{Guidance of diffusion model and performance \label{sec:guidance}}


Recent studies suggest that a lower guidance score in diffusion models often enhances downstream task performance by improving sample diversity, albeit at the expense of reduced sample quality. In Table~\ref{tab:GeFL_perf}, our experminets on MNIST, FMNIST and CIFAR10 exhibited the consistent results where \textsc{GeFL} with DDPM of lower guidance score outperforms \textsc{GeFL} with DDPM of higher guidance score. DDPM with low guidance ($w=0$) does not achieve considerable image quality, despite demonstrating better sample diversity \cite{fan2023scaling}. We also provide t-SNE visualization results in Figure~\ref{fig:ddpm_guide}. We used pre-trained ResNet18 for CIFAR10 to see the ability of well-trained model learning features from real and synthetic images.
The orange dots represent synthetic samples while blue dots represent real samples.
The results demonstrate that synthetic samples of lower guidance score are more diverse than samples of higher guidance score. Compared to the samples of lower guidance score, the samples of higher guidance score are concentrated near the real samples.. 



\subsection{Effects of GAN update}
\label{Sec4.2}

We investigated two settings for training and sampling from FedDCGAN in \textsc{GeFL}: (i) \textit{freezing the FedDCGAN} after training it for $2T$ communication rounds and (ii) \textit{updating the FedDCGAN} for $T$ rounds during target network training following  global knowledge aggregation for $T$ rounds. Table~\ref{tab:gan-upd} demonstrates the effect of updating FedDCGAN during target network training. The table presents results for two scenarios: training exclusively on synthetic data (Syn) and training through \textsc{GeFL}, which involves training target networks on both real and synthetic data (Real+Syn).
Overall, updating FedDCGAN during the training of target networks achieved better performance for both datasets. 
This improvement is attributed to the diverse sampling from various stages of the FedDCGAN, resulting in increased mode coverage and enhanced downstream performance.


\subsection{Evaluation of synthetic data}
In this section, we highlight the differences between Syn and Real+Syn.
Referring to the Table~\ref{tab:gan-upd}, in the Syn setting, compared to the Real+Syn setting, a high standard deviation of accuracies across different runs as well as degraded performance is observed. It demonstrates the effectiveness of synthetic data combined with the real data.
Our proposed method, \textsc{GeFL}, employs both real and synthetic data and consistently showed the superior performance as well as a small standard deviation compared to the Syn, regardless of FedDCGAN freezing or updating.

Recent advances in text-to-image models and diffusion models have sparked interest in evaluating generative models based on downstream task performance, particularly when generated samples are used exclusively for training or as part of data augmentation \cite{tian2023stablerep, azizi2023synthetic}. A primary work \cite{ravuri2019cas} proposed relevant metrics, such as classification accuracy score (CAS) and naive augmentation score (NAS), where CAS measures the performance of target networks trained solely on synthetic data, while NAS assess the performance of target networks trained using a combination of synthetic and real data. A related area of research known as generative data augmentation (GDA) focuses on effectively leveraging synthetic datasets for augmentation alongside real data \cite{shmelkov2018good, yamaguchi2023regularizing}.


In the same context, we assessed the performance of training target networks in Syn and Real+Syn (GeFL) settings within a model-heterogeneous FL setup, as presented in Table~\ref{tab:cas_nas}. Within the GeFL framework, leveraging datasets generated by federated generative models for augmentation yielded superior performance compared to using them exclusively for training.

\captionsetup[subfigure]{font=small,skip=5pt}
\begin{figure}[!t]
\centering
  \begin{subfigure}{0.24\textwidth}
  \centering
    \includegraphics[width=1.1\textwidth]{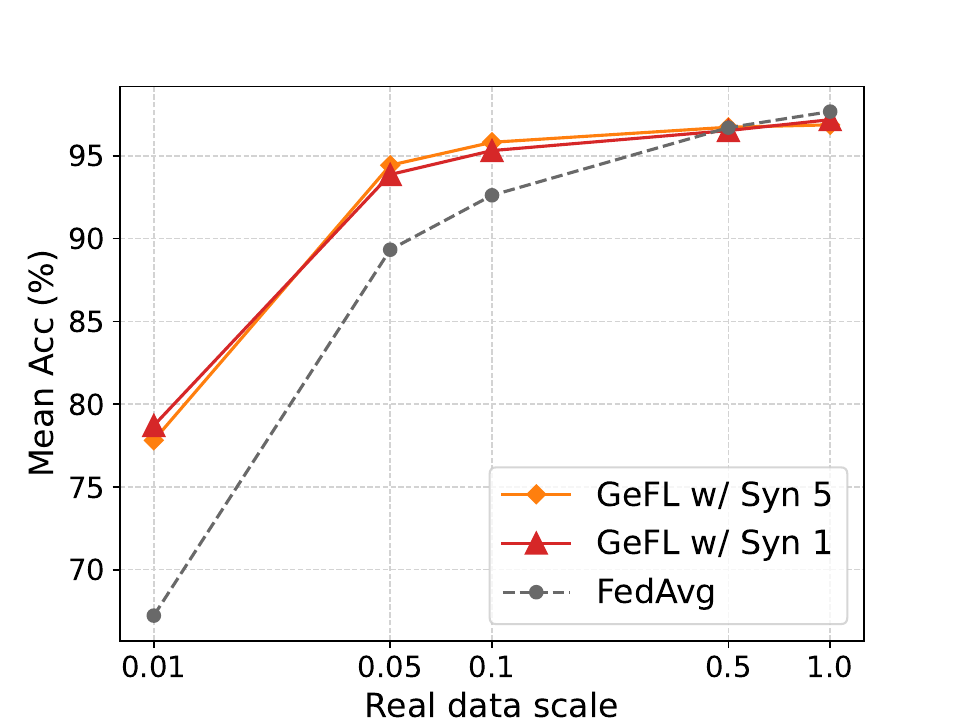}
    \caption{MNIST}
    \label{fig:mnist-scale}
  \end{subfigure}
  \begin{subfigure}{0.24\textwidth}
  \centering
    \includegraphics[width=1.1\textwidth]{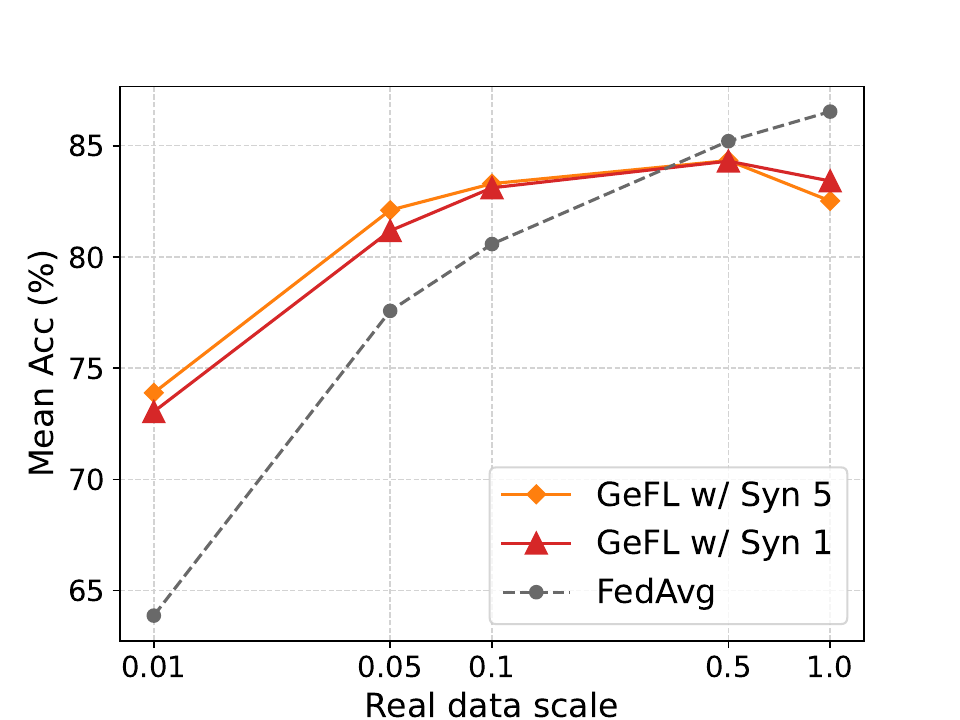}
    \caption{FMNIST}
    \label{fig:fmnist-scale}
  \end{subfigure}
  \caption{\textbf{Performance of \textsc{GeFL} with FedDCGAN across varying dataset scales} demonstrates its effectiveness in scenarios where the amount of real dataset is extremely limited. }
  \label{fig:dataset-scale}
\end{figure}

\subsection{Dataset scale in \textsc{GeFL}\label{sec:scale}}
In Figure~\ref{fig:dataset-scale}, we present the \textsc{GeFL} performance depending on dataset scales, where x-axis denotes the fraction of real dataset used. For example, MNIST has 60000 training samples and fraction rate of 0.1 employs 6000 samples in total. The real dataset is distributed over 10 clients where each client has 600 samples for the fraction rate of 0.1. Compared to FedAvg, gains of GeFL increases as the quantity of real data (possessed by whole clients) decreases.
This definitely shows the effectiveness of \textsc{GeFL} in data-limited scenarios.

Additionally, our findings suggest that simply adding more synthetic data does not always lead to improved performance.
The orange lines with circle markers denote the results of GeFL that generates 320 synthetic data samples ($T_s=5$) every round, while the red lines  with triangle markers denote the results of GeFL that generates 64 synthetic data samples ($T_s=1$) every round.
Notably, across different scales of real data, a fivefold increase in the quantity of synthetic samples showed the comparable performance to generating only onefold.

\begin{table}[!t]
    \centering
    \resizebox{0.495\textwidth}{!}{
    \begin{tabular}{llcccc} 
    \toprule
    {Dataset} & Train & {\textbf{FedDCGAN}} & {\textbf{FedCVAE}} & \makecell{\textbf{FedDDPM}\\$_{w=0}$} & \makecell{\textbf{FedDDPM}\\$_{w=2}$} \\ \midrule
       \multirow{2.5}{*}{MNIST} & Syn  & 85.18 & 91.13 & 91.76 & 92.56 \\ \cmidrule(l){2-6}
       & Real+Syn & 95.32 & 94.46 & 96.44 & 95.17 \\ \midrule
       \multirow{2.5}{*}{FMNIST} & Syn  & 81.83 & 47.96 & 31.08 & 17.20 \\ \cmidrule(l){2-6}
       & Real+Syn & 83.11 & 82.33 & 82.43 & 81.51 \\
       \bottomrule 
    \end{tabular}
    }
    \caption{\textbf{Effectiveness of generated samples} from federated generative models in \textsc{GeFL}}
    \label{tab:cas_nas}
    \vspace{-0.2in}
\end{table}

\section{Conclusion\label{sec:conclusion}}
In this paper, we proposed \textsc{GeFL}, a simple yet effective framework to tackle the challenge of model heterogeneity in FL. Our method utilizes federated generative models to aggregate knowledge across clients with diverse architectures, achieving improved performance over existing baselines.
To further enhance scalability, reduce communication and computation costs, and address privacy concerns, we introduced \textsc{GeFL-F}, an extension that uses feature-generative models trained on lower-dimensional representations. 
Experimental results validate the advantages of \textsc{GeFL-F}, including robustness to increasing client numbers, lower parameter and sampling costs, and enhanced privacy protection. \blue{These findings serve as practical guidelines for deploying FL systems in real-world scenarios, particularly where hardware capabilities vary (e.g., ASICs, mobile SoCs) and where protecting client model architecture and data privacy is critical.
Future directions include exploring decentralized FL frameworks, such as blockchain-integrated solutions like BLADE-FL \cite{li2022bladefl}, to improve robustness, auditability, and eliminate reliance on a central server.}

\bibliographystyle{IEEEtran}
\bibliography{HGref}

\begin{IEEEbiography}[{\includegraphics[width=1in,height=1.25in,clip,keepaspectratio]{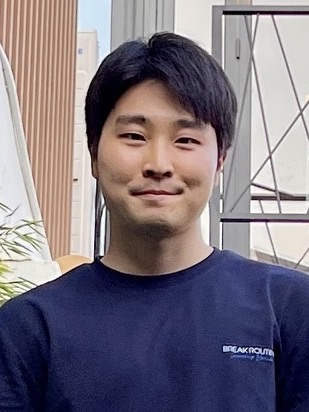}}]{Honggu Kang} received the B.Sc. degree (\textit{summa cum laude}) in electronic engineering from Hanyang University,
Seoul, South Korea, in 2017, and the M.Sc. and Ph.D. degrees from the School of Electrical Engineering, Korea Advanced Institute of Science and Technology (KAIST), Daejeon, South Korea, in 2019 and 2024, respectively.

He is currently a staff engineer in Samsung Electronics, Suwon, South Korea since 2024. His research interests include edge AI, federated learning, signal processing for wireless communications, and unmanned aerial vehicle communications. He was a recipient of the Korean Institute of Communications and Information Sciences (KICS) Fall Symposium Best Paper Award, in 2019.
\end{IEEEbiography}

\begin{IEEEbiography}[{\includegraphics[width=1in,height=1.25in,clip,keepaspectratio]{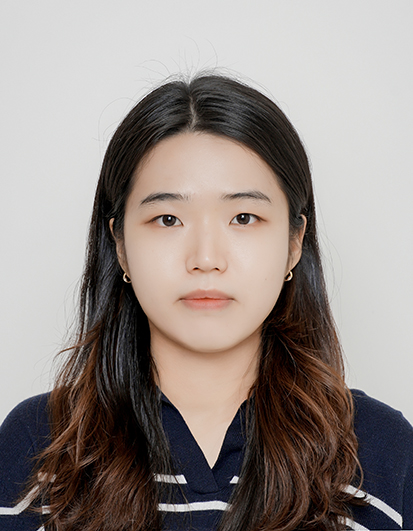}}]{Seohyeon Cha} received the B.S. (\textit{summa cum laude}) and M.S. degrees from the School of Electrical Engineering, Korea Advanced Institute of Science and Technology (KAIST), Daejeon, South Korea, in 2022 and 2024, respectively.

She is currently pursuing the Ph.D. degree in Electrical and Computer Engineering with The University of Texas at Austin, Austin, TX, USA. Her primary research interests include collaborative learning systems, federated learning, and trustworthy machine learning. She was a recipient of the National Science and Engineering Scholarship.
\end{IEEEbiography}

\begin{IEEEbiography}[{\includegraphics[width=1in,height=1.25in,clip,keepaspectratio]{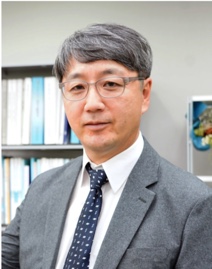}}]{Joonhyuk Kang}
received the B.S.E. and M.S.E. degrees from Seoul National University, Seoul, South Korea, in 1991 and 1993, respectively, and the Ph.D. degree in electrical and computer engineering from The University of Texas at Austin, Austin, in 2002. From 1993 to 1998, he was a Research Staff Member at Samsung Electronics, Suwon, South Korea, where he was involved in the development of DSP-based real-time control systems. In 2000, he was with Cwill Telecommunications, Austin, TX, USA, where he participated in the project for multicarrier CDMA systems with antenna array. He was a Visiting Scholar with the School of Engineering and Applied Sciences, Harvard University, Cambridge, MA, USA, from 2008 to 2009.
He served as Head of the School of Electrical Engineering (EE), KAIST, Daejeon, South Korea. His research interests include signal processing and machine learning for wireless communication systems. He is a life-member of the Korea Information and Communication Society and the Tau Beta Pi (the Engineering Honor Society). He was a recipient of IEEE VTS Jack Neubauer Memorial Award in 2021 for his paper titled ``Mobile Edge Computing via a UAV-Mounted Cloudlet: Optimization of Bit Allocation and Path Planning.''
\end{IEEEbiography}

\newpage
\appendices
\onecolumn
\section{Resource Overhead\label{appx:cost}}
\subsection{Communication cost\label{appx:commun_cost}}
In this section, we present a detailed comparison of communication and memory overhead across methods. Referring to Table~\ref{tab:comm_homo}, we analyze the communication cost for each FL framework. This includes the upload (UL) and download (DL) costs for different model components: feature extractors (FE), knowledge aggregation modules (KA), and target networks (TN). Importantly, communication overhead is conditional on whether clients share model architectures. For example, in both \textsc{GeFL} and \textsc{GeFL-F}, when a client's architecture is unique (i.e., not shared with others), it does not exchange TN parameters rendering UL/DL costs for TN negligible. In contrast, shared architectures involve full parameter exchanges.

As described in Algorithm~\ref{alg:gefl} and Algorithm~\ref{alg:gefl-f}, both \textsc{GeFL} and \textsc{GeFL-F} involve generative knowledge aggregation, which inherently introduces upload and download communication costs between the clients and the server. LG-FedAvg is designed to share only a subset of the target network — specifically, the parameters of the initial layers (e.g., the feature extractor). FedAvg, FedProx, and FedALA shares only the target network parameters.

In the case of FedDF, each client independently trains its local target network using its private dataset. After local training, clients send their target network parameters to the server. The server then utilizes a shared public dataset—which significantly influences the final performance—to perform inference on each received client target network and collect their logit outputs for each sample in the public dataset. These logits are then averaged across all clients to form a soft consensus (i.e., pseudo-labels), which serves as the target for distillation. The server uses this averaged logit information to further update a selected client target network using the public dataset, effectively distilling the collective knowledge.

In contrast, AvgKD operates in a more decentralized fashion. After local training, each client receives the heterogeneous target network parameters from all other clients. Using these target networks, the client performs forward passes on its own local data to obtain logits from all other models. These logits are then averaged across target networks to form soft labels for each local sample. The client uses these aggregated logits as pseudo-labels to train or fine-tune its own target network, after which it broadcasts the updated target network back to the other clients. This process is repeated iteratively to refine the knowledge aggregation across heterogeneous target networks. This procedure incurs a substantial communication overhead due to the need to exchange the complete set of heterogeneous model parameters among clients. As each client possesses a distinct model architecture, the communication burden scales with the number and diversity of participating models, making the process inefficient—particularly in large-scale or highly heterogeneous federated learning environments.

\begin{table}[!h]
\centering
\footnotesize
\caption{Communication overhead in FL methods. UL/DL = upload/download, FE = feature extractor, KA = knowledge aggregation, TN = target network. Parentheses indicate communication occurs only when clients share a homogeneous model with other clients. $|\cdot|$ denotes the model size (i.e., the number of model parameters)}
\begin{tabularx}{0.95\linewidth}{l 
    >{\centering\arraybackslash}X 
    >{\centering\arraybackslash}X 
    >{\centering\arraybackslash}X 
    >{\centering\arraybackslash}X 
    >{\centering\arraybackslash}X 
    >{\centering\arraybackslash}X}
\toprule
\textbf{Method} 
& \makecell{UL to\\Server\\(FE)} 
& \makecell{DL from\\Server\\(FE)} 
& \makecell{UL to\\Server\\(KA)} 
& \makecell{DL from\\Server\\(KA)} 
& \makecell{UL to\\Server\\(TN)} 
& \makecell{DL from\\Server\\(TN)} \\
\midrule
GeFL & -- & -- & $T_{\mathrm{KA}} |\mathbf{w}_g|$ & $T_{\mathrm{KA}} |\mathbf{w}_g|$ & ($T_{\mathrm{TN}} |\bm{\theta}_{g,m}|$) & ($T_{\mathrm{TN}} |\bm{\theta}_{g,m}|$) \\
GeFL-F & $T_{\mathrm{FE}} |\bm{\theta}_g^f|$ & $T_{\mathrm{FE}} |\bm{\theta}_g^f|$ & $T_{\mathrm{KA}} |\mathbf{w}_g|$ & $T_{\mathrm{KA}} |\mathbf{w}_g|$ & ($T_{\mathrm{TN}} |\bm{\theta}_{g,m}|$) & ($T_{\mathrm{TN}} |\bm{\theta}_{g,m}|$) \\ \midrule
FedDF & -- & -- & -- & -- & $T_{\mathrm{TN}} |\bm{\theta}_{g,m}|$ & $T_{\mathrm{TN}} |\bm{\theta}_g|$ \\ \midrule
AvgKD & -- & -- & -- & -- & $T_{\mathrm{TN}} |\bm{\theta}_{g,m}|$ & $\sum_{m \ne i} T_{\mathrm{TN}} |\bm{\theta}_{g,m}|$ \\ \midrule
FedAvg & -- & -- & -- & -- & ($T_{\mathrm{TN}} |\bm{\theta}_{g,m}|$) & ($T_{\mathrm{TN}} |\bm{\theta}_{g,m}|$) \\
FedProx & -- & -- & -- & -- & ($T_{\mathrm{TN}} |\bm{\theta}_{g,m}|$) & ($T_{\mathrm{TN}} |\bm{\theta}_{g,m}|$) \\
FedALA & -- & -- & -- & -- & ($T_{\mathrm{TN}} |\bm{\theta}_{g,m}|$) & ($T_{\mathrm{TN}} |\bm{\theta}_{g,m}|$) \\ \midrule
LG-FedAvg & -- & -- & -- & -- & \makecell{$T_{\mathrm{FE}} |\bm{\theta}_g^f|$\\ ($T_{\mathrm{TN}} |\bm{\theta}_{g,m}|$)} & \makecell{$T_{\mathrm{FE}} |\bm{\theta}_g^f|$\\ ($T_{\mathrm{TN}} |\bm{\theta}_{g,m}|$)} \\
\bottomrule
\end{tabularx}
\label{tab:comm_homo}
\end{table}

\begin{table}[!h]
    \centering
    \caption{Cumulative communication costs (download only) across datasets} \begin{tabular}{lccccc}
        \toprule
        \textbf{Dataset} & FedDF & AvgKD & GeFL (DCGAN) & GeFL (CVAE) & GeFL (DDPM) \\
        \midrule
        MNIST & 69.68 MB & 6.12 GB & 2.13 GB & 8.32 GB & 2.82 GB \\
        FMNIST & 82.50 MB & 7.25 GB & 2.13 GB & 8.32 GB & 2.82 GB \\
        CIFAR10 & 144.42 MB & 12.69 GB & 8.99 GB & 16.74 GB & 5.64 GB \\ \midrule
        CIFAR10 (Table~\ref{tab:model_parameters}) & 395.63 GB & 34.77 TB & 8.99 GB & 16.74 GB & 5.64 GB \\
        \bottomrule
        \end{tabular} \label{tab:commun_cost}
\end{table}
Specifically, to quantify the cumulative communication costs, we calculate the total download volume corresponding to the methods in Table~\ref{tab:GeFL_perf}. The results are summarized in the upper three rows of Table~\ref{tab:commun_cost}. Note that we employed relatively lightweight target models consisting of only a few convolutional layers in our main experiments. These smaller models inherently require fewer parameters and typically converge in fewer communication rounds.
Additionally, we provide communication estimates for models given in Table~\ref{tab:model_parameters} under $T_{KA} = 200$ and $T_{TN} = 300$. These are shown in the last row of Table~\ref{tab:commun_cost}. The result supports the validity and extensibility of our approach.

To further demonstrate the effectiveness of the \textsc{GeFL}, we provide the comparative accuracy results on CIFAR10 dataset with heterogeneous models in Table~\ref{tab:model_parameters}. Referring to Table~\ref{tab:GeFL_large}, despite using a compact generator, our method outperforms baseline FedAvg. Even with smaller generative models, the generated samples provide sufficient diversity to enhance downstream performance across heterogeneous models.

\begin{table}[!h]
\centering
\begin{tabular}{lr}
\toprule
\textbf{Model} & \textbf{Number of parameters} \\
\midrule
ResNet18 & 11,181,642 \\
VGG16 & 134,301,514 \\
Wide\_ResNet50\_2 & 66,854,730 \\
MobileNet\_v3\_Large & 4,214,842 \\
ShuffleNet\_v2\_x1.0 & 1,263,854 \\
ResNeXt50\_32x4d & 23,000,394 \\
MobileNet\_v3\_Small & 1,528,106 \\
RegNet\_Y\_400MF & 3,907,554 \\
EfficientNet\_v2\_S & 20,190,298 \\
ConvNeXt\_Base & 87,558,666 \\
\midrule
DCGAN generator & 11,396,353 \\
DCGAN discriminator & 667,329 \\
\bottomrule
\end{tabular}
\caption{Comparison of neural network models \cite{PytorchModels} and their parameter counts}
\label{tab:model_parameters}
\end{table}

\begin{table}[!h]
    \centering
    \begin{tabular}{lc}
        \toprule
        FedAvg & 49.68 $\pm$ 0.47 \\
        \midrule
        \textsc{{GeFL}} (FedDCGAN) & \textbf{55.85} $\pm$ 0.32\\
        \bottomrule \end{tabular}
    \caption[Mean classification accuracy]{Mean classification accuracy (\%) evaluation of \textsc{GeFL} on the CIFAR10 dataset}
    \label{tab:GeFL_large}
\end{table}

Regarding memory usage, beyond the memory required to store and train the target network (which is a shared requirement across all methods), our proposed frameworks—\textsc{GeFL} and \textsc{GeFL-F}—incur additional memory overhead due to training FedGens and the storage and manipulation of synthetic samples generated by the federated generative models. These synthetic samples are used for target network training and must be retained temporarily in memory during training.
In comparison, FedDF introduces memory overhead at the server-side, as it requires feeding a shared public dataset through all received heterogeneous target models to compute ensemble logits. This necessitates keeping both the dataset and multiple client models in memory during the aggregation process.
Similarly, AvgKD imposes memory burdens on the client-side, as each client must load and perform inference using all other clients' models on its local dataset to compute averaged pseudo-labels. This results in significant memory usage, especially as the number of clients or model sizes increase, since clients must store and process multiple foreign models simultaneously.

\subsection{Computational cost}
In this section, we present a comparison of computational overhead across methods in Table~\ref{tab:computing}.
In FL, the computational costs heavily depend on multiple factors such as the model size ($|\bm{\theta}|$), the number of communication rounds ($T_{TN}$), local training steps ($T_s$, $T_r$), batch size ($B$), and the heterogeneity across clients. These parameters can vary significantly across deployments, making a generalized Big-O complexity analysis less meaningful in practice.
Instead, to provide a fair and practical comparison, we summarized the computing overhead at a conceptual level in Table~\ref{tab:computing}, categorizing the main components of computation (e.g., generative model training, model aggregation, target network training) for \textsc{GeFL}, \textsc{GeFL-F}, and the baseline methods (FedAvg, FedProx, FedDF, AvgKD, LG-FedAvg).
This abstraction helps clearly highlight where additional resource burdens arise in different methods without being tied to specific hyperparameter values. It also avoids misleading interpretations caused by variable experimental settings.
We believe that our current abstract analysis provides useful insights into the resource considerations relevant to each method.

As shown, FedDF imposes a significant server-side computational burden, primarily due to the requirement of performing forward passes over a public dataset using each of the received heterogeneous client models and subsequently updating a selected model using aggregated logits.
In contrast, AvgKD shifts the computational load to the client side. Each client must receive and perform inference using all other clients' models on its own local dataset, followed by fine-tuning its model using the averaged pseudo-labels. This leads to considerable per-client computation, especially in systems with many participants or large models.
On the other hand, our proposed methods, \textsc{GeFL} and \textsc{GeFL-F}, introduce computational overhead associated with training a generative model.
Importantly, \textsc{GeFL} and \textsc{GeFL-F} completely eliminate the need to average logits across all client models in every communication round, a process which can be both computationally intensive and communication-heavy, especially when dealing with large-scale model heterogeneity or high client counts.
Furthermore, our results confirm that the generative models employed in \textsc{GeFL} and \textsc{GeFL-F} can be kept relatively lightweight, yet still achieve strong performance gains, validating their practicality in resource-constrained environments. This balance between computational efficiency and performance makes them attractive alternatives for real-world FL applications where client capabilities vary and scalability is a concern.

\begin{table}[!h]
\centering
\footnotesize
\caption{Computation overhead in FL methods. Parentheses indicate communication occurs only when clients share a homogeneous model with other clients.}
\begin{tabularx}{\linewidth}{l 
    >{\centering\arraybackslash}X 
    >{\centering\arraybackslash}X 
    >{\centering\arraybackslash}X}
\toprule
\multicolumn{4}{c}{\textbf{Server-side Computing}}\\ \midrule
\textbf{Method} & \textbf{FE Model Processing} & \textbf{Addressing Hetero Model} & \textbf{Target Net Processing} \\
\midrule
GeFL       & --  & Generative model aggregation & (Target network aggregation) \\
GeFL-F     & FE model aggregation               & Generative model aggregation                       & (Target network aggregation) \\ \midrule
FedDF      & --     & \makecell{Public data inference\\on client models} & \makecell{Update target model via\\averaged logits (distillation)} \\ \midrule
AvgKD      & --     & --  & -- \\ \midrule
FedAvg     & --     & --  & (Target network aggregation) \\
FedProx    & --     & --  & (Target network aggregation) \\
FedALA    & --     & --  & (Target network aggregation) \\ \midrule
LG-FedAvg  & --     & --  & \makecell{Common parameter aggregation\\(Target network aggregation)} \\
\bottomrule
\end{tabularx}

\vspace{2em}

\begin{tabularx}{\linewidth}{l 
    >{\centering\arraybackslash}X 
    >{\centering\arraybackslash}X 
    >{\centering\arraybackslash}X}
\toprule
\multicolumn{4}{c}{\textbf{Client-side Computing}}\\ \midrule
\textbf{Method} & \textbf{FE Model Processing} & \textbf{Addressing Hetero Model} & \textbf{Target Net Processing} \\
\midrule
GeFL       & --  & Generative model training  & Target network training \\
GeFL-F     & Feature extractor training  & Generative model training                        & Target network training \\ \midrule
FedDF      & -- & -- & Target network training \\ \midrule
AvgKD      & --   & \makecell{Inference on local dataset\\using peer models}    & \makecell{Train target model using\\averaged logits (distillation)} \\ \midrule
FedAvg     & -- & -- & Target network training \\
FedProx    & -- & -- & Target network training \\
FedALA    & -- & -- & Target network training \\ \midrule
LG-FedAvg  & -- & -- & Target network training \\
\bottomrule
\end{tabularx}
\label{tab:computing}
\end{table}

\section{Experimental Details}
All hyperparameter settings for \textsc{GeFL} and \textsc{GeFL-F} are summarized in Table~\ref{tab:target-hyperparam} and Table~\ref{tab:gen-hyperparam}. Table~\ref{tab:target-hyperparam} outlines the hyperparameter configurations for training target networks in \textsc{GeFL} and \textsc{GeFL-F}. Notably, $T_{FE}$ is exclusive to training the feature extractor in \textsc{GeFL-F}.

Table~\ref{tab:gen-hyperparam} presents the hyperparameter settings for training federated generative models (FedGens) in \textsc{GeFL} and \textsc{GeFL-F}. The latent dimensions, denoted as $d_g$ and $d_d$, correspond to parameters of FedDCGAN and FedDCGAN-F, respectively. Similarly, the latent size $l$ is used for FedCVAE and FedCVAE-F, while $n_{feat}$ pertains to FedDDPM and FedDDPM-F. Details on these parameters are provided in Appendix~\ref{appx:gen_model}.

The number of FL communication rounds for training generative models, $T_{KA}$, in Table~\ref{tab:gen-hyperparam} indicates the total rounds used to update the FedGens (e.g., FedDCGANs) during the training process in both \textsc{GeFL} and \textsc{GeFL-F}. For instance, under the \textit{update} setting discussed in Section~\ref{Sec4.2}, FedDCGAN is updated for half of the rounds ($T_{KA}/2$) during global knowledge aggregation stages and subsequently updated for the remaining half during target network training. Conversely, in the \textit{freeze} setting, FedDCGAN is updated for the full $T_{KA}$ rounds exclusively during the global knowledge aggregation stages.

The results presented in Table~\ref{fig:perf_gefl-f-b} and Table~\ref{fig:perf_gefl-f} were obtained based on the configurations outlined in Table~\ref{tab:feat-size-homo}. Detailed model architectures used to generate feature outputs are provided in Appendix~\ref{appx:model_arch}.

For baseline evaluation, we applied a proximal term with a scaling factor of $1\times10^{-2}$  for FedProx \cite{Li2020fedprox}. For AvgKD \cite{afonin2022towards}, pseudo labels were aggregated from the outputs of 10 heterogeneous models. In the case of FedDF \cite{lin2020ensemble}, we used SVHN as the public dataset for MNIST, and CIFAR10 as the public dataset for FMNIST. For LG-FedAvg \cite{liang2020think}, the first convolutional layer was averaged across all heterogeneous models, while subsequent layers were averaged within their respective submodels.
  
For evaluating baselines, we used 1e-2 multiplied to proximal term for FedProx \cite{Li2020fedprox}. Pseudo labels are aggregated from the outputs of 10 heterogeneous models for AvgKD \cite{afonin2022towards}. For FedDF \cite{lin2020ensemble}, we used SVHN as the public dataset for MNIST, CIFAR10 as the public dataset for FMNIST, and CIFAR100 as the public dataset for CIFAR10. For LG-FedAvg \cite{liang2020think} the first conv layer was employed as averaging over all the heterogeneous models while the other layers are averaged across submodels.

To evaluate the FID and IS scores, we generated 1,000 conditional images uniformly distributed across classes, with 100 images per class.

\begin{table}[!h]
    \centering
    \begin{tabular}{lcccc}
        \toprule
        {} & MNIST & FMNIST & CIFAR10 & SVHN \\ \midrule 
        $T_{TN}$ & {50} & {50} & {100} & {100} \\ 
        Optimizer & SGD & SGD & SGD & SGD \\
        Learning rate $\alpha$ & 1e-1 & 1e-1 & 1e-1 & 1e-1\\
        Batch size $B$ & 64 & 64 & 128 & 64 \\
        Data fraction & 0.1 & 0.1 & 0.5 & 0.1 \\
        $T_w/T_g/T_s/T_r$ & \multicolumn{4}{c}{5/5/1/5} \\ \midrule
        $T_{FE}$ &20 &20 &50 &50 \\
        \bottomrule 
        \end{tabular}
    \caption{Hyperparameters for training \textbf{target networks} in GeFL and GeFL-F and \textbf{feature extractor} in GeFL-F.}
    \label{tab:target-hyperparam}
\end{table}


\begin{table}[!h]
    \centering
    \begin{tabular}{clcccccc}
        \toprule
        & \textbf{} & \textbf{FedDCGAN} & \textbf{FedCVAE} & \textbf{FedDDPM} & \textbf{FedDCGAN-F} & \textbf{FedCVAE-F} & \textbf{FedDDPM-F}\\ \midrule
        &Optimizer & Adam & Adam & Adam & Adam & Adam & Adam\\
        &Learning rate $\beta$     & $2 \times 10^{-4}$ & $1 \times 10^{-3}$ & $1 \times 10^{-4}$ & $2 \times 10^{-4}$ & $1 \times 10^{-3}$ & $1 \times 10^{-4}$\\ 
        &Weight decay & -  & $1 \times 10^{-3}$ & - & -  & $1 \times 10^{-3}$ & -\\ 
        &$b_1, b_2$ (Adam) & 0.5, 0.999 & - & - & 0.5, 0.999 & - & - \\ \midrule
        \multirowcell{6}{CIFAR10} &Latent dimension ($d_g$, $d_d$) &256, 64 & - & - &256, 64 & - & - \\
        &Latent size $l$ & - & 50 &- & - & 50 &-\\
        & $n_{feat}$ (of U-Net in DDPM) & - & - & 128 & - & - & 128\\
        &Time step & - & - & 400 & - & - & 500\\
        &Batch size $B$ &64 &64 &64 &64 &64 &64 \\
        &$T_{KA}$ & 200 & 200 & 200 & 200 & 200 & 200\\ \midrule
        \multirowcell{6}{MNIST\\FMNIST\\SVHN} &Latent dimension ($d_g$, $d_d$) &128, 128 & - & - &128, 128 & - & - \\
        &Latent size $l$& - & 16 &- & - & 16 &-\\
        & $n_{feat}$ (of U-Net in DDPM) & - & - & 128 & - & - & 128\\
        &Time step & - & - & 100 & - & - & 100\\
        &Batch size $B$ &64 &64 &64 &64 &64 &64 \\
        &$T_{KA}$ & 100 & 100 & 100 & 100 & 100 & 100\\
        \bottomrule
    \end{tabular}
    \caption{
    Hyperparameters for training both \textbf{generative models and feature-generative models} in a federated scenario.}
    \label{tab:gen-hyperparam}
\end{table}

\begin{table}[!h]
    \centering
    \begin{tabular}{lccccccc}
        \toprule
        {Homogeneity level} & 0 & 1 & 2 & 3 & 4 & 5 & 6\\ \midrule
        Channel & 1 & 3 & 10 & 20 & 40 & 80 & -\\
        Image size & 32\texttimes32 & 16\texttimes16 & 8\texttimes8 & 4\texttimes4 & 2\texttimes2 & 1\texttimes1 & - \\
        $T_{FE}$ & 0 & 20 & 20 & 20 & 50 & 60 & 70 \\
        $T_{TN}$ & 70 & 50 & 50 & 50 & 20 & 10 & 0 \\
        $T_{KA}$ & 100 & 100 & 100 & 100 & 100 & 100 & - \\
        \bottomrule    
        \end{tabular}
    \caption{The size of features according to the homogeneity level}
    \label{tab:feat-size-homo}
\end{table}



\section{Model Architecture\label{appx:model_arch}}
\subsection{Generative Models\label{appx:gen_model}}
In this section, we provide detailed descriptions of the architectures used in our generative models, namely FedGen and its lightweight variant, FedGen-F. These models are developed based on three types of generative frameworks: GAN, VAE, and DDPM. Each pair—FedGAN/FedGAN-F, FedVAE/FedVAE-F, and FedDDPM/FedDDPM-F—shares a common backbone architecture but differs in terms of depth, width, and parameter count.

FedGen is designed with higher model capacity to maximize generative fidelity and expressiveness. In contrast, FedGen-F is carefully crafted to operate under stricter computational and communication constraints typically found in edge devices participating in federated learning. To this end, FedGen-F reduces the number of layers, the size of hidden channels, and overall parameter count while preserving the key architectural principles required for learning meaningful data distributions.
This architectural contrast allows us to evaluate trade-offs between performance and efficiency.

\subsubsection{GAN}
Generative Adversarial Networks (GANs) consist of two components: a generator $G(\boldsymbol{z}; \theta_g)$ and a discriminator $D(\boldsymbol{x}; \theta_d)$. The generator learns to map input noise $\boldsymbol{z}$, sampled from a prior distribution, to the data space, thereby approximating the data distribution. The discriminator outputs a scalar indicating the probability that a given input $\boldsymbol{x}$ is from the real data rather than generated by $G$.

The training objective is a two-player minimax game defined as:
\[
\min _G \max _D \mathbb{E}_{\boldsymbol{x} \sim p(\boldsymbol{x})}[\log D(\boldsymbol{x})] + \mathbb{E}_{\boldsymbol{z} \sim p_{\boldsymbol{z}}(\boldsymbol{z})}[\log (1 - D(G(\boldsymbol{z})))].
\]

\paragraph*{Implementation Details}
We implement both FedDCGAN and FedDCGAN-F using generator and discriminator modules. Referring to Tables~\ref{table:generator} and~\ref{table:generatorf}, the generator takes a 100-dimensional latent vector sampled from a uniform distribution, which is projected in steps 1 and 2. A conditional label is projected in steps 3 and 4.\footnote{Descriptions of \texttt{ConvTranspose2d}, \texttt{BatchNorm2d}, \texttt{relu}, \texttt{leaky-relu}, and \texttt{Conv2d} are based on the documentation provided by the PyTorch library (\url{https://pytorch.org/docs/stable/nn.html}).} These two projections are concatenated and further processed in subsequent layers.

Similarly, referring to Tables~\ref{table:discriminator} and~\ref{table:discriminatorf}, in the discriminator, the input image is projected in step 1, and the label is projected in step 2. These outputs are concatenated and passed through additional convolutional layers.

We set the architecture parameters $(d_g, d_d, c)$ as follows:
\begin{itemize}
    \item MNIST and FMNIST: $(128, 128, 1)$
    \item CIFAR-10: $(256, 64, 3)$
    \item CelebA and SVHN: $(128, 128, 3)$
\end{itemize}

\begin{table}[!h]
\centering
\begin{subtable}[t]{0.48\textwidth}
\centering
\begin{tabular}{|c|l|l|c|}
\hline
\textbf{Step} & \textbf{Layer} & \textbf{Parameters} & \textbf{Activation} \\ \hline
1 & \texttt{ConvTranspose2d} & $(100, 2d_g, 4, 1, 0)$ & \texttt{relu} \\ \hline
2 & \texttt{BatchNorm2d} & $2d_g$ & \\ \hline
3 & \texttt{ConvTranspose2d} & $(10, 2d_g, 4, 1, 0)$ & \texttt{relu} \\ \hline
4 & \texttt{BatchNorm2d} & $2d_g$ & \\ \hline
5 & Concatenate & & \\ \hline
6 & \texttt{ConvTranspose2d} & $(4d_g, 2d_g, 4, 2, 1)$ & \texttt{relu} \\ \hline
7 & \texttt{BatchNorm2d} & $2d_g$ & \\ \hline
8 & \texttt{ConvTranspose2d} & $(2d_g, 2d_g, 4, 2, 1)$ & \texttt{relu} \\ \hline
9 & \texttt{BatchNorm2d} & $d_g$ & \\ \hline
10 & \texttt{ConvTranspose2d} & $(d_g, c, 4, 2, 1)$ & \texttt{tanh} \\ \hline
\end{tabular}
\caption{FedDCGAN Generator Architecture}
\label{table:generator}
\end{subtable}
\hfill
\begin{subtable}[t]{0.48\textwidth}
\centering
\begin{tabular}{|c|l|l|c|}
\hline
\textbf{Step} & \textbf{Layer} & \textbf{Parameters} & \textbf{Activation} \\ \hline
1 & \texttt{Conv2d} & $(c, d_d/2, 4, 2, 1)$ & \texttt{leaky-relu}(0.2) \\ \hline
2 & \texttt{Conv2d} & $(10, d_d/2, 4, 2, 1)$ & \texttt{leaky-relu}(0.2) \\ \hline
3 & Concatenate & & \\ \hline
4 & \texttt{Conv2d} & $(d_d, 2d_d, 4, 2, 1)$ & \texttt{leaky-relu}(0.2) \\ \hline
5 & \texttt{BatchNorm2d} & $2d_d$ & \\ \hline
6 & \texttt{Conv2d} & $(2d_d, 4d_d, 4, 2, 1)$ & \texttt{leaky-relu}(0.2) \\ \hline
7 & \texttt{BatchNorm2d} & $4d_d$ & \\ \hline
8 & \texttt{Conv2d} & $(4d_d, 1, 4, 1, 0)$ & \texttt{sigmoid} \\ \hline
\end{tabular}
\caption{FedDCGAN Discriminator Architecture}
\label{table:discriminator}
\end{subtable}
\caption{Architectures of FedDCGAN Generator and Discriminator.}
\label{table:feddcgan_arch}
\end{table}

\begin{table}[!h]
\centering
\begin{subtable}[t]{0.48\textwidth}
\centering
\begin{tabular}{|c|l|l|c|}
\hline
\textbf{Step} & \textbf{Layer} & \textbf{Parameters} & \textbf{Activation} \\ \hline
1 & \texttt{ConvTranspose2d} & $(100, 2d_g, 4, 1, 0)$ & \texttt{leaky-relu}(0.2) \\ \hline
2 & \texttt{BatchNorm2d} & $2d_g$ & \\ \hline
3 & \texttt{ConvTranspose2d} & $(10, 2d_g, 4, 1, 0)$ & \texttt{leaky-relu}(0.2) \\ \hline
4 & \texttt{BatchNorm2d} & $2d_g$ & \\ \hline
5 & Concatenate & & \\ \hline
6 & \texttt{ConvTranspose2d} & $(4d_g, 2d_g, 4, 2, 1)$ & \texttt{leaky-relu}(0.2) \\ \hline
7 & \texttt{BatchNorm2d} & $2d_g$ & \\ \hline
8 & \texttt{ConvTranspose2d} & $(2d_g, c, 4, 2, 1)$ & \texttt{relu} \\ \hline
\end{tabular}
\caption{FedDCGAN-F Generator Architecture}
\label{table:generatorf}
\end{subtable}
\hfill
\begin{subtable}[t]{0.48\textwidth}
\centering
\begin{tabular}{|c|l|l|c|}
\hline
\textbf{Step} & \textbf{Layer} & \textbf{Parameters} & \textbf{Activation} \\ \hline
1 & \texttt{Conv2d} & $(c, d_d/2, 4, 2, 1)$ & \texttt{leaky-relu}(0.2) \\ \hline
2 & \texttt{Conv2d} & $(10, d_d/2, 4, 2, 1)$ & \texttt{leaky-relu}(0.2) \\ \hline
3 & Concatenate & & \\ \hline
4 & \texttt{Conv2d} & $(d_d, 2d_d, 4, 2, 1)$ & \texttt{leaky-relu}(0.2) \\ \hline
5 & \texttt{BatchNorm2d} & $2d_d$ & \\ \hline
6 & \texttt{Conv2d} & $(2d_d, 4d_d, 4, 2, 1)$ & \texttt{leaky-relu}(0.2) \\ \hline
7 & \texttt{BatchNorm2d} & $4d_d$ & \\ \hline
8 & \texttt{Conv2d} & $(4d_d, 1, 2, 1, 0)$ & \texttt{sigmoid} \\ \hline
\end{tabular}
\caption{FedDCGAN-F Discriminator Architecture}
\label{table:discriminatorf}
\end{subtable}
\caption{Architectures of FedDCGAN-F Generator and Discriminator.}
\label{table:feddcganf_arch}
\end{table}
\subsubsection{VAE}
Variational Autoencoders (VAEs) are trained via maximum likelihood estimation to produce samples similar to the training set. Specifically, VAEs aim to maximize the probability of each training example $\boldsymbol{x}$ as follows:
\begin{equation*}
\max P(\boldsymbol{x}) = \int P(\boldsymbol{x}|\boldsymbol{z};\theta) P(\boldsymbol{z}) \, d\boldsymbol{z}.
\end{equation*}
The model consists of two components: an encoder that maps the input $\boldsymbol{x}$ to a latent variable $\boldsymbol{z}$, and a decoder that reconstructs $\boldsymbol{x}$ from $\boldsymbol{z}$. The latent variable $\boldsymbol{z}$ is sampled from a standard Gaussian prior, $\mathcal{N}(0, I)$.

However, for most $\boldsymbol{z}$ values, the conditional likelihood $P(\boldsymbol{x}|\boldsymbol{z})$ is close to zero. To address this, a new function $Q(\boldsymbol{z}|\boldsymbol{x})$ is introduced as an approximate posterior that provides a distribution over latent variables $\boldsymbol{z}$ likely to generate the given $\boldsymbol{x}$. This approximate posterior is typically modeled as $\boldsymbol{z} = \mu + \sigma \odot \epsilon$ where $\epsilon \sim \mathcal{N}(0, I)$.

The training objective is derived from the evidence lower bound (ELBO) and is given by:
\begin{equation*}
\log P(\boldsymbol{x}) - \mathcal{D}_{\mathrm{KL}}\left[ Q(\boldsymbol{z}|\boldsymbol{x}) \,\|\, P(\boldsymbol{z}|\boldsymbol{x}) \right] = \mathbb{E}_{\boldsymbol{z} \sim Q} \left[\log P(\boldsymbol{x}|\boldsymbol{z})\right] - \mathcal{D}_{\mathrm{KL}} \left[ Q(\boldsymbol{z}|\boldsymbol{x}) \,\|\, P(\boldsymbol{z}) \right],
\end{equation*}
where $\mathcal{D}_{\mathrm{KL}}$ denotes the Kullback–Leibler divergence. At test time, new samples are generated by decoding $\boldsymbol{z} \sim \mathcal{N}(0, I)$, without requiring the encoder.

\paragraph*{Implementation Details}
We implement both FedCVAE and FedCVAE-F using the encoder and decoder.
As summarized in Table~\ref{table:encoder} and Table~\ref{table:encoderf}, the encoder takes an input image and its associated label, reshaped to the same spatial size (e.g., $32 \times 32$ or $16 \times 16$), and concatenates them along the channel dimension (Step 1). This concatenated input is passed through a series of convolutional and fully connected layers up to Step 11. The output is then split into two separate linear layers that produce the parameters of the approximate posterior: $\mu$ via \texttt{Linear$(1024, l)$} (or \texttt{Linear$(512, l)$}), and $\sigma$ via a separate \texttt{Linear$(1024, l)$} (or \texttt{Linear$(512, l)$}) layer.
As shown in Table~\ref{table:decoder} and Table~\ref{table:decoderf}, the decoder concatenates the latent vector $\boldsymbol{z} \in \mathbb{R}^l$ with the one-hot encoded class label (dimension equal to the number of classes), and feeds the resulting vector through a sequence of transposed convolutional and fully connected layers to reconstruct the image.
We used the following configurations of input channels $c$ and latent dimensions $l$, $(c, l)$ as follows:
\begin{itemize}
    \item {MNIST and FMNIST}: $(1, 16)$    
    \item {CIFAR-10}: $(3, 50)$
    \item {CelebA}: $(3, 32)$
    \item {SVHN}: $(3, 16)$
\end{itemize}

\begin{table}[!h]
\centering
\begin{subtable}[t]{0.48\textwidth}
\centering
\begin{tabular}{|c|l|l|c|}
\hline
\textbf{Step} & \textbf{Layer}        & \textbf{Parameters}      & \textbf{Activation} \\ \hline
1 & Concatenate & Input image + label (1 channel) & \\ \hline
2 & \texttt{Conv2d} & $(c+1, 64, 4, 2, 1)$    & \texttt{relu} \\ \hline
3 & \texttt{BatchNorm2d}  & $64$ & \\ \hline
4 & \texttt{Conv2d} & $(64, 128, 4, 2, 1)$    & \texttt{relu}      \\ \hline
5 & \texttt{BatchNorm2d}  & $128$ & \\ \hline
6 & \texttt{Conv2d}       & $(128, 256, 4, 2, 1)$   & \texttt{relu}      \\ \hline
7 & \texttt{BatchNorm2d}  & $256$ & \\ \hline
8 & \texttt{Conv2d}       & $(256, 512, 4, 2, 1)$   & \texttt{relu}      \\ \hline
9 & \texttt{BatchNorm2d}  & $512$ & \\ \hline
10 & \texttt{Conv2d}       & $(512, 1024, 4, 2, 1)$  & \texttt{relu}      \\ \hline
11 & \texttt{BatchNorm2d}  & $1024$ & \\ \hline
12 & \texttt{Linear} (for $\mu$) & $1024, l$ & \\ \hline
13 & \texttt{Linear} (for $\sigma$) & $1024, l$ &  \\ \hline
\end{tabular}
\caption{FedCVAE Encoder Architecture}
\label{table:encoder}
\end{subtable}
\hfill
\begin{subtable}[t]{0.48\textwidth}
\centering
\begin{tabular}{|c|l|l|c|}
\hline
\textbf{Step} & \textbf{Layer}        & \textbf{Parameters}      & \textbf{Activation} \\ \hline
1             & Concatenate           & $z$ + label              &                     \\ \hline
2             & \texttt{Linear}       & $l + 10$                 &                     \\ \hline
3             & \texttt{ConvTranspose2d} & $(1024, 512, 4, 2, 1)$ & \texttt{relu}      \\ \hline
4             & \texttt{BatchNorm2d}  & $512$                    &                     \\ \hline
5             & \texttt{ConvTranspose2d} & $(512, 256, 4, 2, 1)$  & \texttt{relu}      \\ \hline
6             & \texttt{BatchNorm2d}  & $256$                    &                     \\ \hline
7             & \texttt{ConvTranspose2d} & $(256, 128, 4, 2, 1)$  & \texttt{relu}      \\ \hline
8             & \texttt{BatchNorm2d}  & $128$                    &                     \\ \hline
9             & \texttt{ConvTranspose2d} & $(128, 64, 4, 2, 1)$   & \texttt{relu}      \\ \hline
10            & \texttt{BatchNorm2d}  & $64$                     &                     \\ \hline
11            & \texttt{ConvTranspose2d} & $(64, c, 4, 2, 1)$     & \texttt{relu}      \\ \hline
\end{tabular}
\caption{FedCVAE Decoder Architecture}
\label{table:decoder}
\end{subtable}
\caption{Architectures of FedCVAE Encoder and Decoder.}
\label{table:fedcvae_arch}
\end{table}

\begin{table}[!h]
\centering
\begin{subtable}[t]{0.48\textwidth}
\centering
\begin{tabular}{|c|l|l|c|}
\hline
\textbf{Step} & \textbf{Layer}        & \textbf{Parameters}      & \textbf{Activation} \\ \hline
1   & Concatenate           & Input image + label (1 channel) & \\ \hline
2   & \texttt{Conv2d}       & $(c+1, 64, 4, 2, 1)$    & \texttt{relu}      \\ \hline
3   & \texttt{BatchNorm2d}  & $64$   & \\ \hline
4   & \texttt{Conv2d}       & $(64, 128, 4, 2, 1)$    & \texttt{relu}      \\ \hline
5   & \texttt{BatchNorm2d}  & $128$  & \\ \hline
6   & \texttt{Conv2d}       & $(128, 256, 4, 2, 1)$   & \texttt{relu}      \\ \hline
7   & \texttt{BatchNorm2d}  & $256$  & \\ \hline
8   & \texttt{Conv2d}       & $(256, 512, 4, 2, 1)$   & \texttt{relu}      \\ \hline
9   & \texttt{BatchNorm2d}  & $512$  & \\ \hline
10  & \texttt{Linear} (for $\mu$) & $512, l$ & \\ \hline
11  & \texttt{Linear} (for $\sigma$) & $512, l$ & \\ \hline
\end{tabular}
\caption{FedCVAE-F Encoder Architecture}
\label{table:encoderf}
\end{subtable}
\hfill
\begin{subtable}[t]{0.48\textwidth}
\centering
\begin{tabular}{|c|l|l|c|}
\hline
\textbf{Step} & \textbf{Layer}        & \textbf{Parameters}      & \textbf{Activation} \\ \hline
1  & Concatenate           & $z$ + label & \\ \hline
2  & \texttt{Linear}       & $l + 10$ & \\ \hline
3  & \texttt{ConvTranspose2d} & $(512, 256, 4, 2, 1)$  & \texttt{relu}      \\ \hline
4 & \texttt{BatchNorm2d}  & $256$ & \\ \hline
5 & \texttt{ConvTranspose2d} & $(256, 128, 4, 2, 1)$  & \texttt{relu} \\ \hline
6 & \texttt{BatchNorm2d}  & $128$ & \\ \hline
7 & \texttt{ConvTranspose2d} & $(128, 64, 4, 2, 1)$   & \texttt{relu}      \\ \hline
8 & \texttt{BatchNorm2d}  & $64$ & \\ \hline
9 & \texttt{ConvTranspose2d} & $(64, c, 4, 2, 1)$     & \texttt{relu}      \\ \hline
\end{tabular}
\caption{FedCVAE-F Decoder Architecture}
\label{table:decoderf}
\end{subtable}
\caption{Architectures of FedCVAE-F Encoder and Decoder.}
\label{table:fedcvaef_arch}
\end{table}

\subsubsection{DDPM}
Denoising Diffusion Probabilistic Models (DDPMs) are a class of generative models that define a latent variable model for the data distribution as 
\[
p_\theta\left(\mathbf{x}_0\right):=\int p_\theta\left(\mathbf{x}_{0: T}\right) d \mathbf{x}_{1: T},
\]
where $\mathbf{x}_0$ denotes the observed data, and $\mathbf{x}_1, \ldots, \mathbf{x}_T$ are latent variables with the same dimensionality as $\mathbf{x}_0$. DDPMs model the generative process as a Markov chain that starts from pure noise and progressively denoises through learned reverse transitions.

The forward process, also known as the diffusion process, gradually adds Gaussian noise to $\mathbf{x}_0$ over $T$ time steps, following a predefined variance schedule. This process is also Markovian and allows sampling of any intermediate state $\mathbf{x}_t$ in closed form, which makes training efficient and analytically tractable.

The reverse process is also modeled as a Markov chain, where the goal is to learn the conditional distributions $p_\theta(\mathbf{x}_{t-1} \mid \mathbf{x}_t)$ that reverse the noising process. The training objective is derived from variational inference and is given by:
\[
\mathbb{E}_q\left[
\mathcal{D}\left(q\left(\mathbf{x}_T \mid \mathbf{x}_0\right) \| p\left(\mathbf{x}_T\right)\right) 
+ \sum_{t>1} \mathcal{D}\left(q\left(\mathbf{x}_{t-1} \mid \mathbf{x}_t, \mathbf{x}_0\right) \| p_\theta\left(\mathbf{x}_{t-1} \mid \mathbf{x}_t\right)\right) 
- \log p_\theta\left(\mathbf{x}_0 \mid \mathbf{x}_1\right)
\right],
\]
where $\mathcal{D}$ denotes KL divergence and $q$ represents the known forward process distributions.

\paragraph*{Implementation Details.}
Our implementations, FedDDPM and its lightweight counterpart FedDDPM-F, adopt the U-Net backbone architecture \cite{ronneberger2015unet}, a widely used structure for DDPMs due to its multi-scale representation capability. We followed architectural adjustments \cite{github_diffusion}, and refer to our code \cite{Kang} for further implementation specifics.

To reflect different resource settings and performance trade-offs, we used:
\begin{itemize}
    \item FedDDPM with $T=100$ for MNIST, FMNIST, and SVHN, and $T=400$ for CIFAR10.
    \item FedDDPM-F with $T=100$ for MNIST, FMNIST, and SVHN, and $T=500$ for CIFAR10.
\end{itemize}
The variant FedDDPM-F is designed to reduce the number of parameters and computational cost by simplifying the network architecture, making it more suitable for federated settings with constrained clients.

\subsection{Target network architectures}
For MNIST, FMNIST, SVHN, and CIFAR10 datasets, we employed 10 heterogeneous CNN architectures, detailed in Table~\ref{tab:hetero_arch}, Table~\ref{tab:hetero_arch_bn}, and Table~\ref{tab:hetero_arch_bn_cifar}. In all tables, conv($c,k,p$) denotes a 2d convolutional layer, where $c$ is the output channel size, $k$ is the kernel size, and $p$ is the padding size. bn($c$) represents a batch normalization layer \cite{ioffe2015batch} with $c$ denoting the channel size. The term relu denotes the rectified linear layer \cite{agarap2018deep}, and maxpool($k,s,p$) denotes a max-pooling layer. where $k$ is the kernel size, $s$ is the stride, and $p$ is the padding size. Finally, fc($in/out$) indicates a fully connected layer, where $in$ is the number of input nodes and $out$ is the number of output nodes.

In the experiments depicted in Figure~\ref{fig:perf_gefl-f} and Figure~\ref{fig:perf_gefl-f-b}, we utilized ten CNN-10 models as target networks to ensure fair comparison. For evaluating ten heterogeneous CNN-10 models, the target networks are not aggregated even though the target networks are equivalent in architecture. Despite utilizing equivalent models, as heterogeneity increases, the common feature extractor ($\bm{\theta}_{g}^{f}$) decreases in size. Note that $\bm{\theta}_{g}^{f}$ is aggregated across the heterogeneous models.
For instance, CNN-10 in Table~\ref{tab:hetero_arch}, at homogeneity level 1, $\bm{\theta}_{g}^{f}$ consists of conv(3,3\texttimes3,1) followed by bn(3), relu, maxpool(2\texttimes 2,2,0).
On the other hand, at homogeneity level 2, conv(10,3\texttimes3,1)  followed by relu, maxpool(2\texttimes 2,2,0) are additionally included in 
$\bm{\theta}_{g}^{f}$.


\begin{table}[h!]
\resizebox{0.999\textwidth}{!}{
\begin{tabular}{|c|c|c|c|c|c|c|c|c|c|}
\toprule \textbf{CNN-1} & \textbf{CNN-2} & \textbf{CNN-3} & \textbf{CNN-4} & \textbf{CNN-5} & \textbf{CNN-6} & \textbf{CNN-7} & \textbf{CNN-8} & \textbf{CNN-9} & \textbf{CNN-10} \\
\hline\hline conv(3,3\texttimes3,1) & conv(3,3\texttimes 3,1) & conv(3,3\texttimes 3,1) & conv(3,3\texttimes 3,1) & conv(3,3\texttimes 3,1) & conv(3,3\texttimes 3,1) & conv(3,3\texttimes 3,1) & conv(3,3\texttimes 3,1) & conv(3,3\texttimes 3,1) & conv(3,3\texttimes 3,1)\\
bn(3) & bn(3) & bn(3) & bn(3) & bn(3) & bn(3) & bn(3) & bn(3) & bn(3) & bn(3)\\
relu & relu & relu & relu & relu & relu & relu & relu & relu & relu\\
maxpool(2\texttimes 2,2,0)& maxpool(2\texttimes 2,2,0)& maxpool(2\texttimes 2,2,0)& maxpool(2\texttimes 2,2,0)& maxpool(2\texttimes 2,2,0)& maxpool(2\texttimes 2,2,0)& maxpool(2\texttimes 2,2,0)& maxpool(2\texttimes 2,2,0)& maxpool(2\texttimes 2,2,0)& maxpool(2\texttimes 2,2,0)\\
conv(16,3\texttimes 3,1) & conv(16, 3\texttimes 3,1) & conv(20,3\texttimes 3,1) & conv(10,3\texttimes 3,1) & conv(16,3\texttimes 3,1) & conv(20,3\texttimes 3,1) & conv(10,3\texttimes 3,1) & conv(16,3\texttimes 3,1) & conv(20,3\texttimes 3,1) & conv(10,3\texttimes 3,1) \\
relu & relu & relu & relu & relu & relu & relu & relu & relu & relu\\
maxpool(2\texttimes 2,2,0)& maxpool(2\texttimes 2,2,0)& maxpool(2\texttimes 2,2,0)& maxpool(2\texttimes 2,2,0)& maxpool(2\texttimes 2,2,0)& maxpool(2\texttimes 2,2,0)& maxpool(2\texttimes 2,2,0)& maxpool(2\texttimes 2,2,0)& maxpool(2\texttimes 2,2,0)& maxpool(2\texttimes 2,2,0)\\
 & conv(32,3\texttimes 3,1) & conv(40,3\texttimes 3,1) & conv(20,3\texttimes 3,1) & conv(32,3\texttimes 3,1) & conv(40,3\texttimes 3,1)  & conv(20,3\texttimes 3,1)  & conv(32,3\texttimes 3,1)  & conv(40,3\texttimes 3,1)  & conv(20,3\texttimes 3,1)\\
 & relu & relu & relu & relu & relu & relu & relu & relu & relu\\
& maxpool(2\texttimes 2,2,0)& maxpool(2\texttimes 2,2,0)& maxpool(2\texttimes 2,2,0)& maxpool(2\texttimes 2,2,0)& maxpool(2\texttimes 2,2,0)& maxpool(2\texttimes 2,2,0)& maxpool(2\texttimes 2,2,0)& maxpool(2\texttimes 2,2,0)& maxpool(2\texttimes 2,2,0)\\
 & & & & conv(64,3\texttimes 3,1) & conv(80,3\texttimes 3,1) & conv(40,3\texttimes 3,1) & conv(64,3\texttimes 3,1) & conv(80,3\texttimes 3,1) & conv(40,3\texttimes 3,1)\\
 & & & & relu & relu & relu & relu & relu & relu\\
 & & & & maxpool(2\texttimes 2,2,0)& maxpool(2\texttimes 2,2,0)& maxpool(2\texttimes 2,2,0)& maxpool(2\texttimes 2,2,0)& maxpool(2\texttimes 2,2,0)& maxpool(2\texttimes 2,2,0)\\
 & & & & & & & conv(128,3\texttimes 3,1) & conv(100,3\texttimes 3,1) & conv(80,3\texttimes 3,1)\\
 & & & & & & & relu & relu & relu\\
 & & & & & & & maxpool(2\texttimes 2,2,0)& maxpool(2\texttimes 2,2,0)& maxpool(2\texttimes 2,2,0)\\
fc(1024/10) & fc(512/10) & fc(640/10) & fc(320/10) & fc(256/10) & fc(320/10) & fc(160/10) & fc(128/10) & fc(100/10) & fc(80/10)\\
\bottomrule
\end{tabular}
}
\caption{\textbf{Heterogeneous target networks (MNIST) }having a common feature extractor and heterogeneous headers.}
\label{tab:hetero_arch}
\end{table}

\begin{table}[h!]
\resizebox{ \textwidth}{!}{
\begin{tabular}{|c|c|c|c|c|c|c|c|c|c|}
\toprule \textbf{CNN-1} & \textbf{CNN-2} & \textbf{CNN-3} & \textbf{CNN-4} & \textbf{CNN-5} & \textbf{CNN-6} & \textbf{CNN-7} & \textbf{CNN-8} & \textbf{CNN-9} & \textbf{CNN-10} \\
\hline\hline conv(3,3\texttimes3,1) & conv(3,3\texttimes 3,1) & conv(3,3\texttimes 3,1) & conv(3,3\texttimes 3,1) & conv(3,3\texttimes 3,1) & conv(3,3\texttimes 3,1) & conv(3,3\texttimes 3,1) & conv(3,3\texttimes 3,1) & conv(3,3\texttimes 3,1) & conv(3,3\texttimes 3,1)\\
bn(3) & bn(3) & bn(3) & bn(3) & bn(3) & bn(3) & bn(3) & bn(3) & bn(3) & bn(3)\\
relu & relu & relu & relu & relu & relu & relu & relu & relu & relu\\
maxpool(2\texttimes 2,2,0)& maxpool(2\texttimes 2,2,0)& maxpool(2\texttimes 2,2,0)& maxpool(2\texttimes 2,2,0)& maxpool(2\texttimes 2,2,0)& maxpool(2\texttimes 2,2,0)& maxpool(2\texttimes 2,2,0)& maxpool(2\texttimes 2,2,0)& maxpool(2\texttimes 2,2,0)& maxpool(2\texttimes 2,2,0)\\
conv(16,3\texttimes 3,1) & conv(16, 3\texttimes 3,1) & conv(20,3\texttimes 3,1) & conv(10,3\texttimes 3,1) & conv(16,3\texttimes 3,1) & conv(20,3\texttimes 3,1) & conv(10,3\texttimes 3,1) & conv(16,3\texttimes 3,1) & conv(20,3\texttimes 3,1) & conv(10,3\texttimes 3,1) \\
bn(16) & bn(16) & bn(20) & bn(10) & bn(16) & bn(20) & bn(10) & bn(16) & bn(20) & bn(10)\\
relu & relu & relu & relu & relu & relu & relu & relu & relu & relu\\
maxpool(2\texttimes 2,2,0)& maxpool(2\texttimes 2,2,0)& maxpool(2\texttimes 2,2,0)& maxpool(2\texttimes 2,2,0)& maxpool(2\texttimes 2,2,0)& maxpool(2\texttimes 2,2,0)& maxpool(2\texttimes 2,2,0)& maxpool(2\texttimes 2,2,0)& maxpool(2\texttimes 2,2,0)& maxpool(2\texttimes 2,2,0)\\
 & conv(32,3\texttimes 3,1) & conv(40,3\texttimes 3,1) & conv(20,3\texttimes 3,1) & conv(32,3\texttimes 3,1) & conv(40,3\texttimes 3,1)  & conv(20,3\texttimes 3,1)  & conv(32,3\texttimes 3,1)  & conv(40,3\texttimes 3,1)  & conv(20,3\texttimes 3,1)\\
 & bn(32) & bn(40) & bn(20) & bn(32) & bn(40) & bn(20) & bn(32) & bn(40) & bn(20)\\
 & relu & relu & relu & relu & relu & relu & relu & relu & relu\\
& maxpool(2\texttimes 2,2,0)& maxpool(2\texttimes 2,2,0)& maxpool(2\texttimes 2,2,0)& maxpool(2\texttimes 2,2,0)& maxpool(2\texttimes 2,2,0)& maxpool(2\texttimes 2,2,0)& maxpool(2\texttimes 2,2,0)& maxpool(2\texttimes 2,2,0)& maxpool(2\texttimes 2,2,0)\\
 & & & & conv(64,3\texttimes 3,1) & conv(80,3\texttimes 3,1) & conv(40,3\texttimes 3,1) & conv(64,3\texttimes 3,1) & conv(80,3\texttimes 3,1) & conv(40,3\texttimes 3,1)\\
 & & & & bn(64) & bn(80) & bn(40) & bn(64) & bn(80) & bn(40)\\
 & & & & relu & relu & relu & relu & relu & relu\\
 & & & & maxpool(2\texttimes 2,2,0)& maxpool(2\texttimes 2,2,0)& maxpool(2\texttimes 2,2,0)& maxpool(2\texttimes 2,2,0)& maxpool(2\texttimes 2,2,0)& maxpool(2\texttimes 2,2,0)\\
 & & & & & & & conv(128,3\texttimes 3,1) & conv(100,3\texttimes 3,1) & conv(80,3\texttimes 3,1)\\
 & & & & & & & bn(128) & bn(100) & bn(80)\\
 & & & & & & & relu & relu & relu\\
 & & & & & & & maxpool(2\texttimes 2,2,0)& maxpool(2\texttimes 2,2,0)& maxpool(2\texttimes 2,2,0)\\
fc(1024/10) & fc(512/10) & fc(640/10) & fc(320/10) & fc(256/10) & fc(320/10) & fc(160/10) & fc(128/10) & fc(100/10) & fc(80/10)\\
\bottomrule
\end{tabular}
}
\caption{\textbf{Heterogeneous target networks (FMNIST and SVHN)} having a common feature extractor and heterogeneous headers.}
\label{tab:hetero_arch_bn}
\end{table}

\begin{table}[h!]
\resizebox{ \textwidth}{!}{
\begin{tabular}{|c|c|c|c|c|c|c|c|c|c|}
\toprule \textbf{CNN-1} & \textbf{CNN-2} & \textbf{CNN-3} & \textbf{CNN-4} & \textbf{CNN-5} & \textbf{CNN-6} & \textbf{CNN-7} & \textbf{CNN-8} & \textbf{CNN-9} & \textbf{CNN-10} \\
\hline\hline conv(3,3\texttimes3,1) & conv(3,3\texttimes 3,1) & conv(3,3\texttimes 3,1) & conv(3,3\texttimes 3,1) & conv(3,3\texttimes 3,1) & conv(3,3\texttimes 3,1) & conv(3,3\texttimes 3,1) & conv(3,3\texttimes 3,1) & conv(3,3\texttimes 3,1) & conv(3,3\texttimes 3,1)\\
bn(3) & bn(3) & bn(3) & bn(3) & bn(3) & bn(3) & bn(3) & bn(3) & bn(3) & bn(3)\\
relu & relu & relu & relu & relu & relu & relu & relu & relu &  relu\\
 conv(10,3\texttimes3,1)& conv(10,3\texttimes3,1)& conv(10,3\texttimes3,1)& conv(10,3\texttimes3,1)& conv(10,3\texttimes3,1)& conv(10,3\texttimes3,1)& conv(10,3\texttimes3,1)& conv(10,3\texttimes3,1)& conv(10,3\texttimes3,1)&conv(10,3\texttimes3,1)\\
 bn(10)& bn(10)& bn(10)& bn(10)& bn(10)& bn(10)& bn(10)& bn(10)& bn(10)&bn(10)\\
 relu & relu & relu & relu & relu & relu & relu & relu & relu &relu\\
maxpool(2\texttimes 2,2,0)& maxpool(2\texttimes 2,2,0)& maxpool(2\texttimes 2,2,0)& maxpool(2\texttimes 2,2,0)& maxpool(2\texttimes 2,2,0)& maxpool(2\texttimes 2,2,0)& maxpool(2\texttimes 2,2,0)& maxpool(2\texttimes 2,2,0)& maxpool(2\texttimes 2,2,0)& maxpool(2\texttimes 2,2,0)\\
conv(16,3\texttimes 3,1)& conv(16,3\texttimes 3,1)& conv(20,3\texttimes 3,1)& conv(10,3\texttimes 3,1)& conv(10,3\texttimes 3,1)& conv(20,3\texttimes 3,1)& conv(10,3\texttimes 3,1)& conv(16,3\texttimes 3,1)& conv(20,3\texttimes 3,1)& conv(10,3\texttimes 3,1)\\
 relu & relu & relu & relu & relu & relu & relu & relu & relu &relu\\
maxpool(2\texttimes 2,2,0)& maxpool(2\texttimes 2,2,0)& maxpool(2\texttimes 2,2,0)& maxpool(2\texttimes 2,2,0)& maxpool(2\texttimes 2,2,0)& maxpool(2\texttimes 2,2,0)& maxpool(2\texttimes 2,2,0)& maxpool(2\texttimes 2,2,0)& maxpool(2\texttimes 2,2,0)& maxpool(2\texttimes 2,2,0)\\
 & conv(32\texttimes 3,1)& conv(40,3\texttimes 3,1)& conv(20,3\texttimes 3,1)& conv(32,3\texttimes 3,1) & conv(40,3\texttimes 3,1)  & conv(20,3\texttimes 3,1)  & conv(32,3\texttimes 3,1)& conv(40,3\texttimes 3,1)  & conv(20,3\texttimes 3,1)\\
 & relu & relu & relu& relu & relu & relu& relu & relu & relu\\
& maxpool(2\texttimes 2,2,0)& maxpool(2\texttimes 2,2,0)& maxpool(2\texttimes 2,2,0)& maxpool(2\texttimes 2,2,0)& maxpool(2\texttimes 2,2,0)& maxpool(2\texttimes 2,2,0)& maxpool(2\texttimes 2,2,0)& maxpool(2\texttimes 2,2,0)& maxpool(2\texttimes 2,2,0)\\
 & & & & conv(64,3\texttimes 3,1) & conv(80,3\texttimes 3,1) & conv(40,3\texttimes 3,1) & conv(64,3\texttimes 3,1) & conv(80,3\texttimes 3,1) & conv(40,3\texttimes 3,1)\\
 & & & & relu & relu & relu & relu & relu & relu\\
 & & & & maxpool(2\texttimes 2,2,0)& maxpool(2\texttimes 2,2,0)& maxpool(2\texttimes 2,2,0)& maxpool(2\texttimes 2,2,0)& maxpool(2\texttimes 2,2,0)& maxpool(2\texttimes 2,2,0)\\
 & & & & & & & conv(128,3\texttimes 3,1) & conv(100,3\texttimes 3,1) & conv(80,3\texttimes 3,1)\\
 & & & & & & & relu & relu & relu\\
 & & & & & & & maxpool(2\texttimes 2,2,0)& maxpool(2\texttimes 2,2,0)& maxpool(2\texttimes 2,2,0)\\
fc(1024/10) & fc(512/10) & fc(640/10) & fc(320/10) & fc(256/10) & fc(320/10) & fc(160/10) & fc(128/10) & fc(100/10) & fc(80/10)\\
\bottomrule
\end{tabular}
}
\caption{\textbf{Heterogeneous target networks (CIFAR10)} having a common feature extractor and heterogeneous headers.}
\label{tab:hetero_arch_bn_cifar}
\end{table}
\newpage

\section{Additional Experiment Results}
\subsection{Different target networks}


For the CIFAR10 dataset, we employed a diverse set of 10 heterogeneous target networks, comprising eight models with EfficientNet backbones (EfficientNet-B1 through EfficientNet-B7 \cite{tan2019efficientnet}) and two models with ResNet backbones (ResNet18 and ResNet32 \cite{he2016deep}). The hyperparameter settings for training the federated generative models are consistent with those outlined in Table~\ref{tab:gen-hyperparam}.
As observed in Table~\ref{tab:data_aug_res}, similar to the results in Table~\ref{tab:data_aug}, \textsc{GeFL} consistently outperforms the baseline methods. The performance degradation observed in both the baselines and \textsc{GeFL} can be attributed to the increased training demands, requiring more communication rounds and larger volumes of data to effectively train the target networks, particularly for larger models such as EfficientNet and ResNet.


\begin{table}[!h]
    \centering
    \begin{tabular}{lcc}
        \toprule
        {Method} & {FedAvg} & {\textsc{GeFL} (FedDCGAN)} \\ \midrule
        None        &  ${45.11}_{\pm 0.46}$ & ${50.43}_{\pm 1.01}$ \\ \midrule
        MixUp       &  ${47.99}_{\pm 3.15}$ & $\mathbf{52.95}_{\pm 0.78}$ \\
        CutMix      &  ${46.61}_{\pm 1.29}$ & ${51.21}_{\pm 1.09}$ \\
        AugMix      &  ${45.18}_{\pm 0.92}$ & ${51.76}_{\pm 0.92}$\\
        AutoAugment &  ${47.89}_{\pm 0.21}$ & ${51.77}_{\pm 2.03}$ \\
        \bottomrule
    \end{tabular}
    \caption{\textbf{Mean classification accuracy (\%) comparison to data augmentation on Res+Eff target networks.} \textsc{GeFL} outperforms other baselines and is effective combined with data augmentation.}
    \label{tab:data_aug_res}
\end{table}

\subsection{Other Datasets}
We conducted experiments on the CelebA dataset, which contains higher-resolution facial images and is commonly used in privacy and fairness research. In this study, we focused on a single-attribute classification task (gender).
Specifically, we evaluated \textsc{GeFL} using FedCVAE and \textsc{GeFL-F} using FedCVAE-F, varying the number of clients to test scalability. The experimental results, summarized in Table~\ref{tab:GeFL_celeba}, demonstrate consistent trends with our original evaluations on lower-resolution datasets. Notably, \textsc{GeFL} significantly outperforms LG-FedAvg (grouped) across all client scales, confirming its effectiveness in aggregating heterogeneous client knowledge. Moreover, \textsc{GeFL-F} maintains competitive performance with fewer clients and surpasses \textsc{GeFL} as the number of clients increases —highlighting its advantage in scalability and communication efficiency.
In addition to classification accuracy, we examined privacy risks through the MND ratio (Table~\ref{tab:priv_celeba}). The results show that \textsc{GeFL-F} achieves a notably lower MND ratio (0.16) compared to \textsc{GeFL} (0.66), indicating enhanced resistance to memorization and better privacy preservation.
These findings on the CelebA dataset strengthen our claim that both \textsc{GeFL} and \textsc{GeFL-F} generalize well across different resolutions and client scales, while \textsc{GeFL-F} offers improved privacy and scalability under resource constraints. 

\begin{table}[!h]
    \centering
    \begin{tabular}{lccc}
        \toprule
        \# of clients & 10 & 50 & 100\\
        \midrule
        LG-FedAvg (grouped) & 75.02 & 72.214 & 69.747\\
        \midrule
        \textsc{{GeFL}} (FedCVAE) & 83.57 & 81.65 & 73.84\\
        \textsc{{GeFL-F}} (FedCVAE-F) & 83.63 & 80.68 & 79.19\\
        \bottomrule \end{tabular}
    \caption[Mean classification accuracy]{Mean classification accuracy (\%) evaluation of \textsc{GeFL} on the CelebA dataset}
    \label{tab:GeFL_celeba}
\end{table}

\begin{table}[!h]
    \centering
    \begin{tabular}{lcc}
        \toprule
        & \textsc{GeFL} (FedCVAE) & \textsc{GeFL-F} (FedCVAE-F)\\
        \midrule
        MND ratio & 0.66 & 0.16\\
        \bottomrule
        \end{tabular}
    \caption[MND]{Mean neighbor distance (MND) ratio evaluation of \textsc{GeFL} and \textsc{GeFL-F} on the CelebA dataset}
    \label{tab:priv_celeba}
\end{table}

\subsection{Discussion on Applicability to Non-Image Domains}
While this paper focuses on image classification tasks, the underlying concept of generative model-aided FL may extend to other data domains, such as text or tabular data. However, such extensions present several challenges. For instance, the MND metric used to evaluate privacy leakage and synthetic sample diversity is specifically suited for images and may not directly apply to textual representations.

Furthermore, generative models in the NLP domain—such as large transformer-based models—are typically resource-intensive, making them impractical for deployment on edge devices in federated settings. That said, adapting the framework to non-image domains introduces new challenges. For instance, state-of-the-art generative models such as GPT or BERT are computationally expensive and unsuitable for deployment on resource-constrained clients. Moreover, such models are often trained centrally on massive corpora, limiting their compatibility with federated learning paradigms.
Although lightweight alternatives (e.g., distilled transformers or small RNNs) exist, scaling up language models has been shown to predictably improve performance and sample efficiency on a wide range of downstream task.

In summary, while \textsc{GeFL-F} offers a promising direction for extending generative federated learning beyond vision tasks, doing so demands further research into domain-appropriate architectures, representation schemes, and evaluation metrics. Exploring such extensions and developing domain-specific metrics for privacy and utility remain valuable directions for future work.

\subsection{Memorization of generative models}
\begin{figure}[!h]
\centering
    \begin{subfigure}{0.22\textwidth}
        \centering
        \includegraphics[width=1\textwidth]{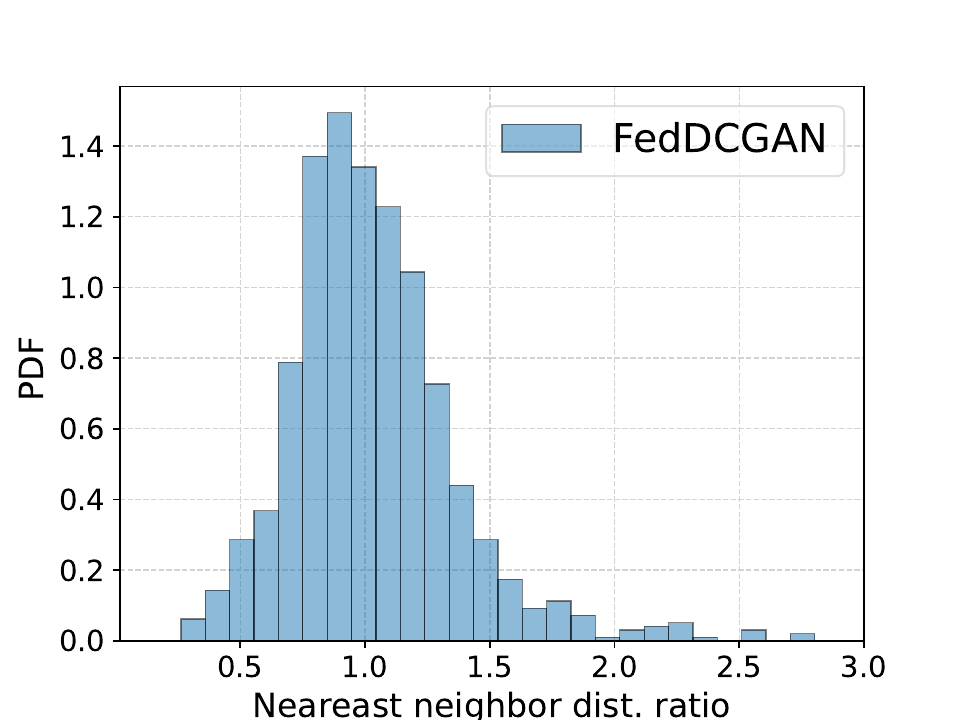}
        \caption*{$\text{MND}=1.035$}
    \end{subfigure}
    \begin{subfigure}{0.22\textwidth}
        \centering
        \includegraphics[width=1\textwidth]{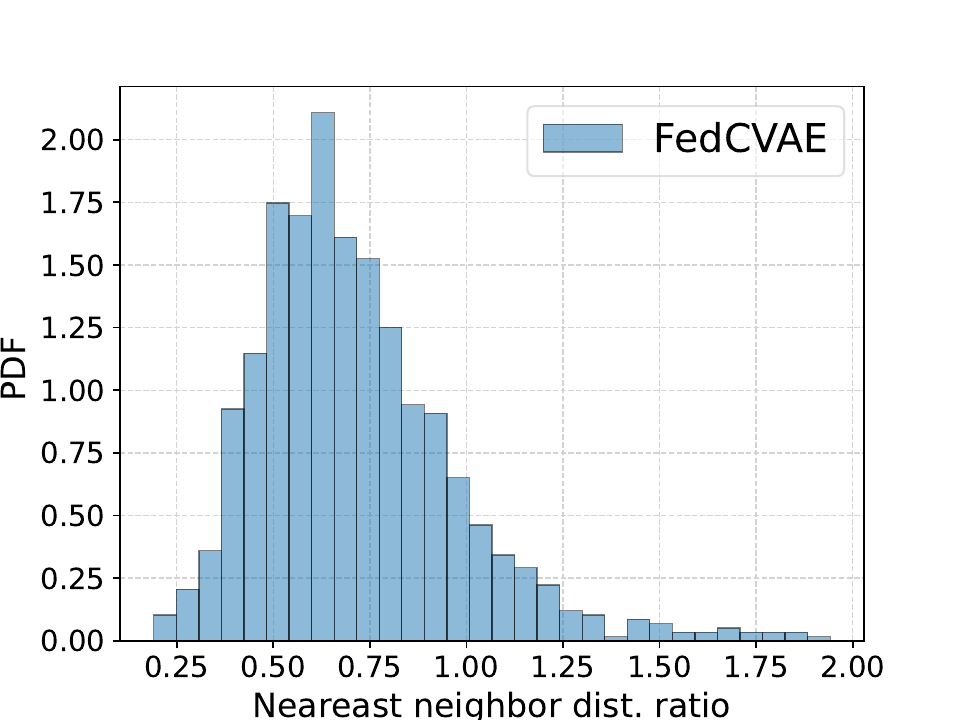}
        \caption*{$\text{MND}=0.714$}
    \end{subfigure}
    \begin{subfigure}{0.22\textwidth}
        \centering
        \includegraphics[width=1\textwidth]{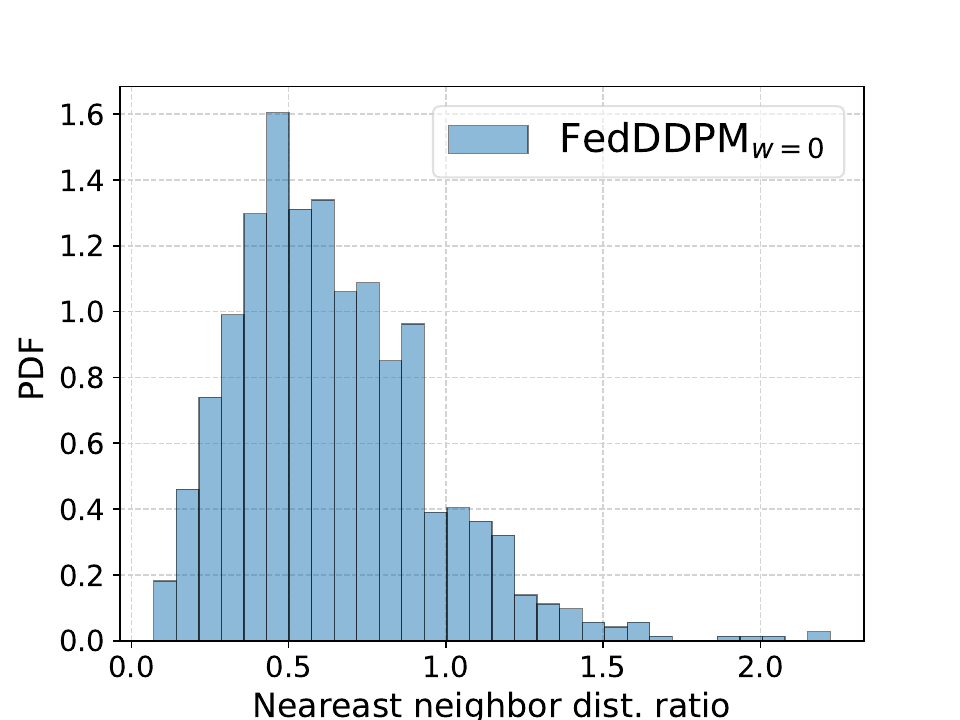}
        \caption*{$\text{MND}=0.640$}
    \end{subfigure}
    \begin{subfigure}{0.22\textwidth}
        \centering
        \includegraphics[width=1\textwidth]{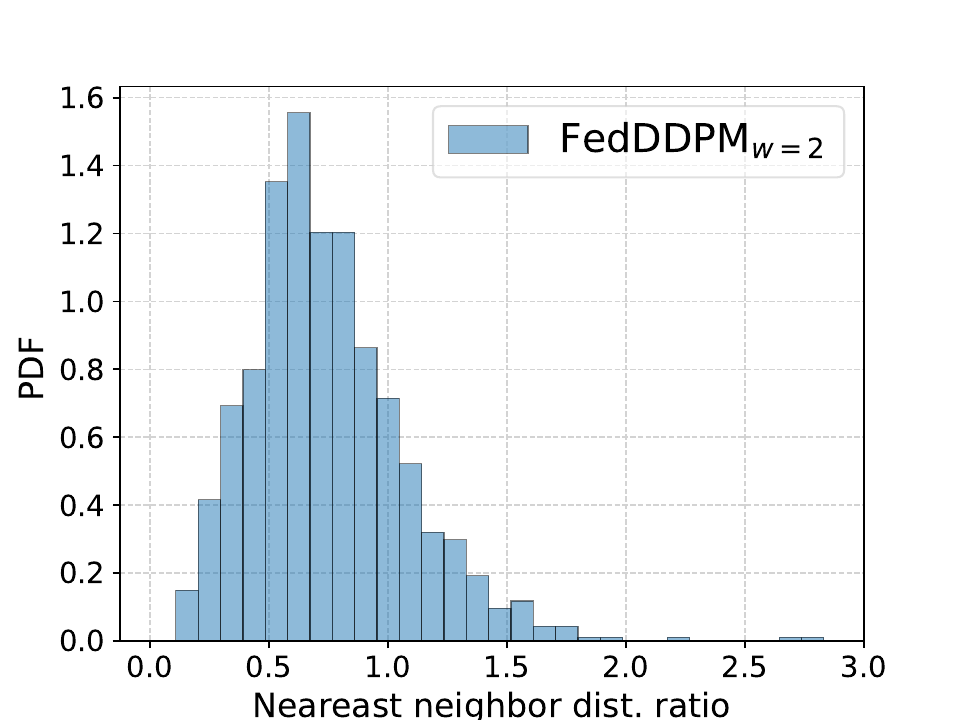}
        \caption*{$\text{MND}=0.753$}
    \end{subfigure}
    \\
    \begin{subfigure}{0.22\textwidth}
        \centering
        \includegraphics[width=1\textwidth]{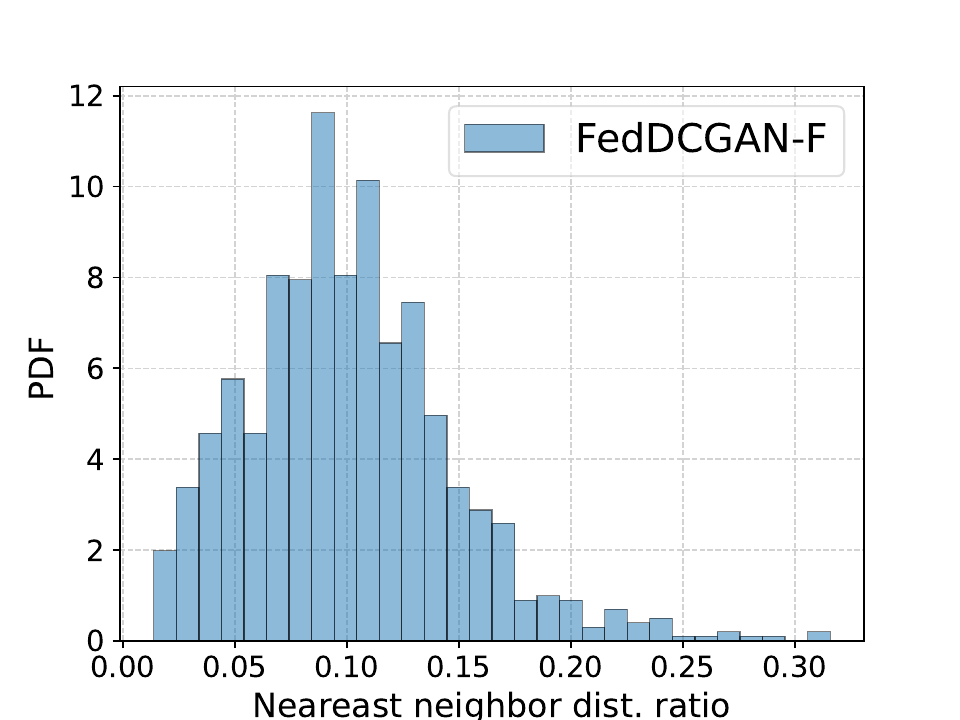}
        \caption*{$\text{MND}=0.101$}
    \end{subfigure}
    \begin{subfigure}{0.22\textwidth}
        \centering
        \includegraphics[width=1\textwidth]{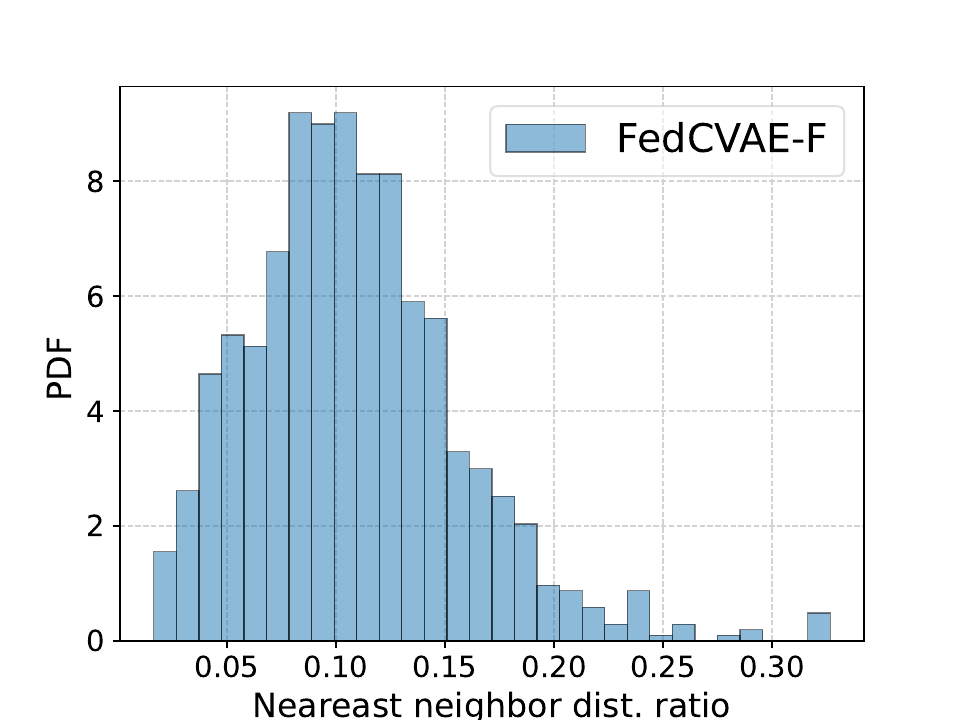}
        \caption*{$\text{MND}=0.108$}
        \label{fig:fedcvaef_memo}
    \end{subfigure}
    \begin{subfigure}{0.22\textwidth}
        \centering
        \includegraphics[width=1\textwidth]{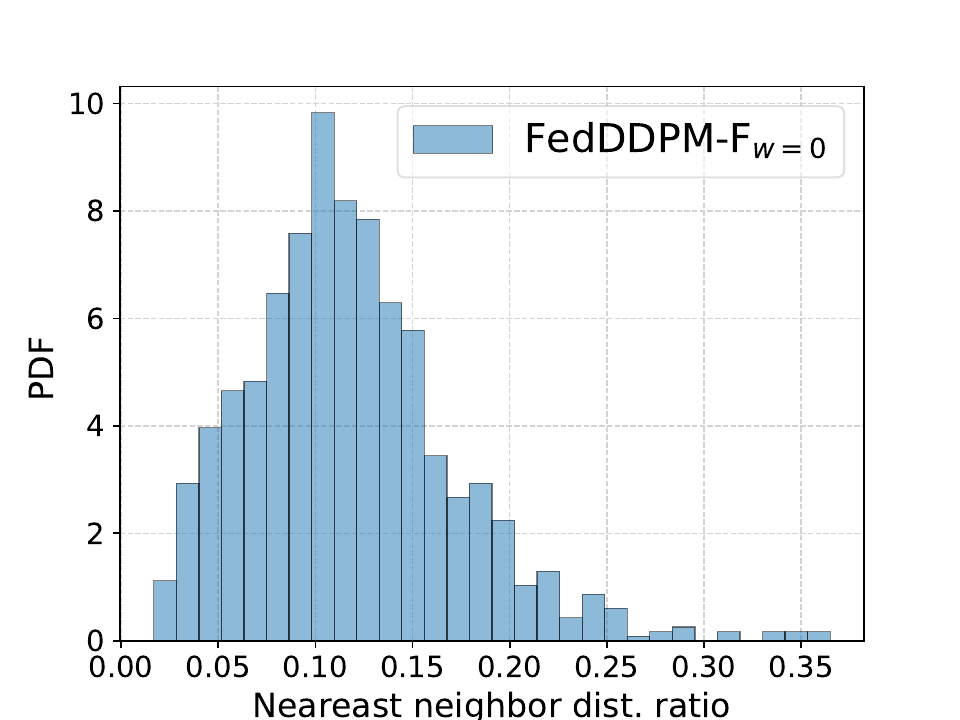}
        \caption*{$\text{MND}=0.118$}
    \end{subfigure}
    \begin{subfigure}{0.22\textwidth}
        \centering
        \includegraphics[width=1\textwidth]{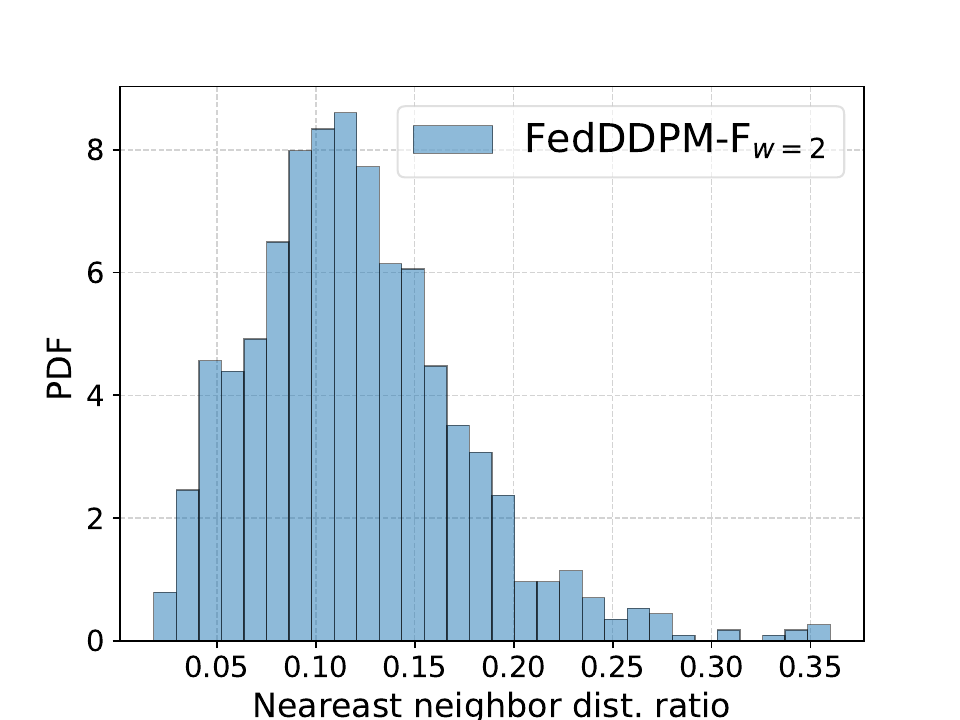}
        \caption*{$\text{MND}=0.120$}
    \end{subfigure}
\caption{\textbf{Memorization analysis of generative models and feature-generative models} in \textsc{GeFL} and \textsc{GeFL-F}, respectively, assessed through the MND ratio on MNIST dataset. Here, the x-axis and y-axis represent the nearest neighbor distance ratio and the distribution over training set, respectively.}
\label{fig:memo-fedgen}
\end{figure}

We measure privacy in our framework, \textsc{GeFL} and \textsc{GeFL-F}, in terms of \textit{memorization}. In \textsc{GeFL}, where the target networks are not shared, privacy leakage is solely due to the memorization of shared generative models. Memorization occurs when there is an increased probability of generating a sample closely resembling the training data \cite{van2021on}. This is particularly concerning in scenarios involving sensitive data like medical images or images containing private information. A typical way to evaluate memorization is to compare the generated samples to their nearest neighbors in the training set \cite{meehan2020nonparametric, somepalli2023diffusion}. 

We elaborated on how we assess memorization in \textsc{GeFL} using the \textit{mean nearest neighbor distance (MND) ratio}, as detailed in Section~\ref{Sec2.3}. The MND ratio for each training sample is computed as the ratio between the distance to the nearest synthetic sample and the nearest validation sample. While any metric can be employed as a distance \cite{van2021on}, we use the perceptual metric LPIPS, which captures image similarity as human perception, in contrast to traditional metrics like Euclidean distance. The memorization of federated generative models of \textsc{GeFL} is evaluated as Figure~\ref{fig:memo-fedgen}, illustrating the distribution of the nearest neighbor distance ratio over the training set of MNIST. 

In \textsc{GeFL-F}, a common feature extractor and feature-generative model are shared between the server and clients in the FL pipeline. A promising way that attacker can take using trained feature extractor and feature-generative model is to generated synthetic features from generative model and apply model inversion using feature extractor. To evaluate the vulnerability of \textsc{GeFL-F} in such scenario, we assume a white-box feature extractor and apply model inversion \cite{zeiler2014visualizing} using generated features to get reconstructed images as Figure~\ref{fig:mnist-model-inv}. Subsequently, we compare the original real images with the reconstructed ones by measuring the MND ratio, as we did in \textsc{GeFL}. In Figure~\ref{fig:memo-fedgen}, it is observed that incorporating feature-generative models mitigates privacy preservation compared to \textsc{GeFL}, even though it shares a common feature extractor.

\subsection{Guidance score and time steps of diffusion models}
We present additional results analyzing the impact of the guidance score on performance, with a particular focus on its correlation with time steps $T$. By varying the time steps for DDPM, we observed that under sufficient time steps, lower guidance scores ($w=0$) yielded superior performance compared to higher guidance scores ($w=2$), as shown in Table~\ref{tab:guide-ddpm}. Within our framework, while lower guidance scores enhance sample diversity, using fewer time steps results in degraded image quality, leading to better performance with higher guidance scores under such conditions.

\begin{table}[!h]
    \centering
        \begin{subtable}[c]{0.4\textwidth}
        \centering
    \begin{tabular}{lccc}
        \toprule
        {$T$} & 30 & 60 & 100 \\ \midrule \midrule
        {$w=0$} & {94.79$_{\pm 0.37}$} & {\textbf{95.68}$_{\pm 0.32}$} & {\textbf{96.44}$_{\pm 0.05}$}\\
        {$w=2$} & {\textbf{95.18}$_{\pm 0.14}$} & {95.60$_{\pm 0.16}$} & {95.17$_{\pm 0.14}$}\\\bottomrule      
    \end{tabular}
        \caption{MNIST}
    \end{subtable}
    \begin{subtable}[c]{0.4\textwidth}
        \centering
    \begin{tabular}{lccc}
        \toprule
        {$T$} & 30 & 60 & 100 \\ \midrule \midrule
        {$w=0$} & {80.49$_{\pm 1.29}$} & {81.72$_{\pm 0.09}$} & {\textbf{82.43}$_{\pm 0.22}$}\\
        {$w=2$} & {\textbf{82.08}$_{\pm 0.35}$} & {\textbf{81.87}$_{\pm 0.43}$} & {81.51$_{\pm 0.38}$}\\\bottomrule    
    \end{tabular}
        \caption{FMNIST}
    \end{subtable}
    \caption{{Mean accuracy (\%) of \textsc{GeFL} with FedDDPM across different guidance scores and time steps $T$.}}
    \label{tab:guide-ddpm}
\end{table}



\newpage
\section{Example of generated samples and features}

\begin{figure}[!h]
\centering
\begin{subfigure}{0.35\textwidth}
    \centering
    \includegraphics[width=\textwidth]{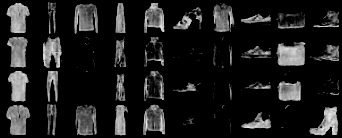}
    \caption{FedDCGAN}
    \label{fig:gan-fmnist}
\end{subfigure}
\begin{subfigure}{0.35\textwidth}
    \centering
    \includegraphics[width=\textwidth]{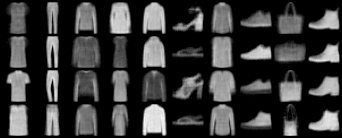}
    \caption{FedCVAE}
    \label{fig:vae-fmnist}
\end{subfigure}

\begin{subfigure}{0.35\textwidth}
    \centering
    \includegraphics[width=\textwidth]{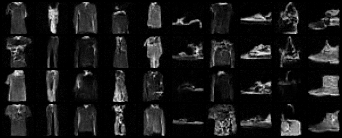}
    \caption{FedDDPM$_{w=0}$}
    \label{fig:ddpm0-fmnist}
\end{subfigure}
\begin{subfigure}{0.35\textwidth}
    \centering
    \includegraphics[width=\textwidth]{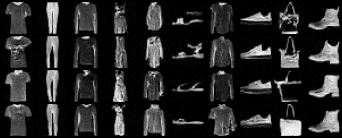}
    \caption{FedDDPM$_{w=2}$}
    \label{fig:ddpm2-fmnist}
\end{subfigure}
\caption{\textbf{Generated samples of FMNIST} by federated generative models with 10 clients in \textsc{GeFL}.}
\end{figure}

\begin{figure}[!h]
\centering
\begin{subfigure}{0.35\textwidth}
    \centering
    \includegraphics[width=\textwidth]{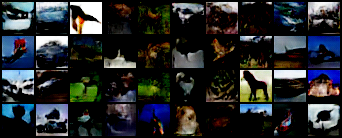}
    \caption{FedDCGAN}
    \label{fig:gan-cifar10}
\end{subfigure}
\begin{subfigure}{0.35\textwidth}
    \centering
    \includegraphics[width=\textwidth]{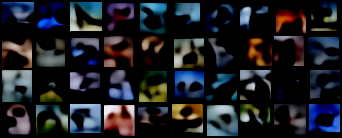}
    \caption{FedCVAE}
    \label{fig:vae-cifar10}
\end{subfigure}

\begin{subfigure}{0.35\textwidth}
    \centering
    \includegraphics[width=\textwidth]{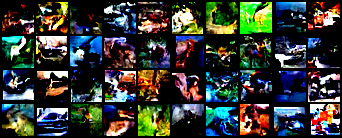}
    \caption{FedDDPM$_{w=0}$}
    \label{fig:ddpm0-cifar10}
\end{subfigure}
\begin{subfigure}{0.35\textwidth}
    \centering
    \includegraphics[width=\textwidth]{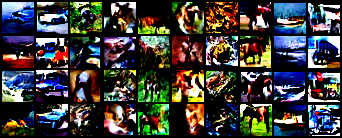}
    \caption{FedDDPM$_{w=2}$}
    \label{fig:ddpm2-cifar10}
\end{subfigure}
\caption{\textbf{Generated samples of CIFAR10} by federated generative models with 10 clients in \textsc{GeFL}.}
\end{figure}

\begin{figure}[!h]
\centering
\begin{subfigure}{0.25\textwidth}
    \centering
    \includegraphics[width=\textwidth]{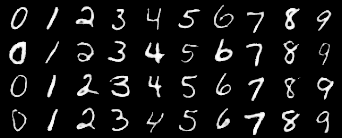}
    \caption{FedDCGAN (10 clients)}
    \label{fig:feddcganf10}
\end{subfigure}
\begin{subfigure}{0.25\textwidth}
    \centering
    \includegraphics[width=\textwidth]{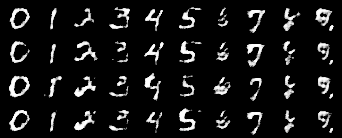}
    \caption{FedDCGAN (50 clients)}
\end{subfigure}
\begin{subfigure}{0.25\textwidth}
    \centering
    \includegraphics[width=\textwidth]{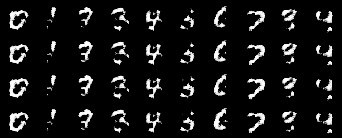}
    \caption{FedDCGAN (100 clients)}
    \label{fig:feddcgan100}
\end{subfigure}

\begin{subfigure}{0.25\textwidth}
    \centering
    \includegraphics[width=\textwidth]{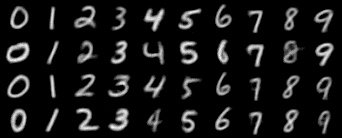}
    \caption{FedCVAE (10 clients)}
\end{subfigure}
\begin{subfigure}{0.25\textwidth}
    \centering
    \includegraphics[width=\textwidth]{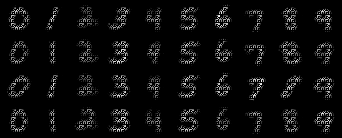}
    \caption{FedCVAE (50 clients)}
\end{subfigure}
\begin{subfigure}{0.25\textwidth}
    \centering
    \includegraphics[width=\textwidth]{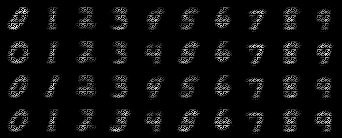}
    \caption{FedCVAE (100 clients)}
\end{subfigure}

\begin{subfigure}{0.25\textwidth} 
    \centering
    \includegraphics[width=\textwidth]{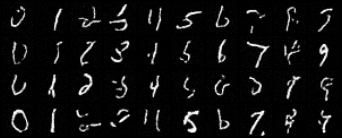}
    \caption{FedDDPM$_{w=0}$ (10 clients)}
\end{subfigure}
\begin{subfigure}{0.25\textwidth}
    \centering
    \includegraphics[width=\textwidth]{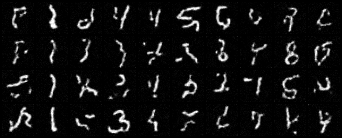}
    \caption{FedDDPM$_{w=0}$ (50 clients)}
\end{subfigure}
\begin{subfigure}{0.25\textwidth}
    \centering
    \includegraphics[width=\textwidth]{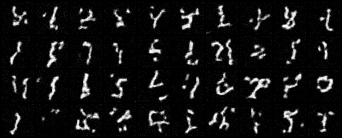}
    \caption{FedDDPM$_{w=0}$ (100 clients)}
\end{subfigure}

\begin{subfigure}{0.25\textwidth}
    \centering
    \includegraphics[width=\textwidth]{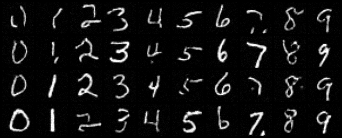}
    \caption{FedDDPM$_{w=2}$ (10 clients)}
    \label{fig:fedddpmw210}
\end{subfigure}
\begin{subfigure}{0.25\textwidth}
    \centering
    \includegraphics[width=\textwidth]{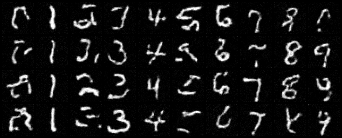}
    \caption{FedDDPM$_{w=2}$ {(50 clients)}}
    \label{fig:fedddpmw250}
\end{subfigure}
\begin{subfigure}{0.25\textwidth}
    \centering
    \includegraphics[width=\textwidth]{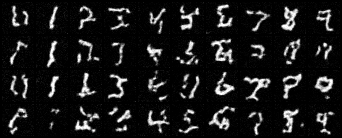}
    \caption{FedDDPM$_{w=2}$ (100 clients)}
    \label{fig:fedddpmw2100}
\end{subfigure}
\caption{\textbf{Generated samples of MNIST by federated generative models in \textsc{GeFL}} across the different number of users.}
\label{fig:fedgen-orig}
\end{figure}

\begin{figure}[!h]
\centering
\begin{subfigure}{0.25\textwidth}
    \centering
    \includegraphics[width=\textwidth]{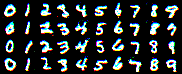}
    \caption{FedDCGAN (10 clients)}
\end{subfigure}
\begin{subfigure}{0.25\textwidth}
    \centering
    \includegraphics[width=\textwidth]{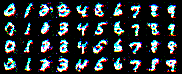}
    \caption{FedDCGAN (50 clients)}
    \label{fig:feddcgan50}
\end{subfigure}
\begin{subfigure}{0.25\textwidth}
    \centering
    \includegraphics[width=\textwidth]{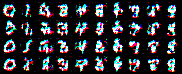}
    \caption{FedDCGAN (100 clients)}
\end{subfigure}

\begin{subfigure}{0.25\textwidth}
    \centering
    \includegraphics[width=\textwidth]{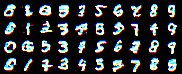}
    \caption{FedCVAE (10 clients)}
    \label{fig:fedcvaef10}
\end{subfigure}
\begin{subfigure}{0.25\textwidth}
    \centering
    \includegraphics[width=\textwidth]{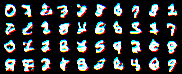}
    \caption{FedCVAE (50 clients)}
    \label{fig:fedcvae50}
\end{subfigure}
\begin{subfigure}{0.25\textwidth}
    \centering
    \includegraphics[width=\textwidth]{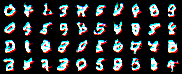}
    \caption{FedCVAE (100 clients)}
    \label{fig:fedcvae100}
\end{subfigure}

\begin{subfigure}{0.25\textwidth} 
    \centering
    \includegraphics[width=\textwidth]{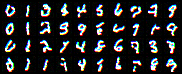}
    \caption{FedDDPM$_{w=0}$ (10 clients)}
    \label{fig:fedddpmw010}
\end{subfigure}
\begin{subfigure}{0.25\textwidth}
    \centering
    \includegraphics[width=\textwidth]{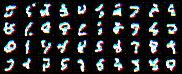}
    \caption{FedDDPM$_{w=0}$ (50 clients)}
    \label{fig:fedddpmw050}
\end{subfigure}
\begin{subfigure}{0.25\textwidth}
    \centering
    \includegraphics[width=\textwidth]{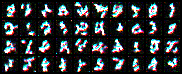}
    \caption{FedDDPM$_{w=0}$ (100 clients)}
    \label{fig:fedddpmw0100}
\end{subfigure}

\begin{subfigure}{0.25\textwidth}
    \centering
    \includegraphics[width=\textwidth]{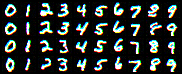}
    \caption{FedDDPM$_{w=2}$ (10 clients)}
\end{subfigure}
\begin{subfigure}{0.25\textwidth}
    \centering
    \includegraphics[width=\textwidth]{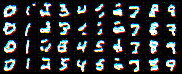}
    \caption{FedDDPM$_{w=2}$ {(50 clients)}}
\end{subfigure}
\begin{subfigure}{0.25\textwidth}
    \centering
    \includegraphics[width=\textwidth]{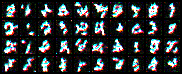}
    \caption{FedDDPM$_{w=2}$ (100 clients)}
\end{subfigure}
\caption{\textbf{Feature samples of MNIST generated by feature-generative models in \textsc{GeFL-F}} across the different number of users.}
\label{fig:fedgen-features}
\end{figure}

\end{document}